\documentclass[10pt]{article}

\PassOptionsToPackage{authoryear,round}{natbib}
\PassOptionsToPackage{colorlinks=true,
    linkcolor=blue!70!black,
    citecolor=blue!70!black,
    urlcolor=blue!70!black}{hyperref}

\usepackage[preprint]{tmlr}
\bibpunct{(}{)}{;}{a}{,}{,}
\usepackage[section]{placeins}
\usepackage{tikz}

\usepackage{amsmath,amssymb,amsfonts}

\usepackage{url}

\usepackage{booktabs}
\usepackage{multirow}
\usepackage{longtable}
\usepackage{array}

\usepackage{xcolor}
\usepackage{enumitem}
\usepackage{caption}
\usepackage{subcaption}
\usepackage{graphicx}
\usepackage{verbatim}
\usepackage{hyperref}
\hypersetup{
    colorlinks=true,
    linkcolor=blue!70!black,
    citecolor=blue!70!black,
    urlcolor=blue!70!black
}

\newenvironment{reviewchange}{\begingroup}{\endgroup}

\def\month{12}
\def\year{2026}
\usepackage[preprint]{tmlr}

\def\month{12}
\def\year{2026}
\def\openreview{\url{https://openreview.net/forum?id=rbj6jsUJtR}}
\title{Beyond Independent Optimization:
Compression, MoE Routing, and Quantization Interactions in Multimodal Edge Intelligence}

\author{
\name Jay Gor \textsuperscript{*}
\email 23bce113@nirmauni.ac.in \\
\addr Nirma University
\AND
\name Karm Dave \textsuperscript{*}
\email 23bce137@nirmauni.ac.in \\
\addr Nirma University
\AND
\name Akshita Abrol \textsuperscript{*}
\email akshita.abrol@singaporetech.edu.sg \\
\addr Singapore Institute of Technology, Singapore
\AND
\name Rajesh Gupta \email rajesh.gupta@marwadieducation.edu.in  \\
\addr Department of CE-AI and Big Data, Marwadi University
\AND
\name Sudeep Tanwar \email rajesh.gupta@marwadieducation.edu.in \\
\addr Department of CE-AI and Big Data, Marwadi University
\AND
\name Zhengkui Wang \email zhengkui.wang@singaporetech.edu.sg \\
\addr Singapore Institute of Technology, Singapore
\\[0.6em]
{\normalfont\small
\textsuperscript{*}The first three authors contributed equally to this work.}
}
\begin{document}

\maketitle

\begin{abstract}
\begin{reviewchange}
Large vision-language models, multimodal large language models, and sparse
Mixture-of-Experts architectures are increasingly constrained not only by model
quality, but by the cost of moving, storing, routing, and quantizing multimodal
representations under tight latency, memory, and energy budgets.  This paper reviews
more than one hundred recent works on efficient multimodal intelligence across five
interconnected themes: visual token redundancy and compression, token management in
multimodal LLMs and video, KV-cache optimisation for long-context inference, MoE
routing with quantization and high-throughput serving, and edge deployment with
hardware-aware benchmarking.

The central argument is that these techniques should not be evaluated in
isolation. We organise their potential dependencies through a cross-stage framework
and distinguish relationships supported by direct evidence from those supported
only by indirect evidence or remaining as open questions. Existing studies show
that low-bit quantization can alter MoE routing. Our controlled ScienceQA
evaluations on DeepSeek-VL2-Tiny and InternVL3.5-30B-A3B, extended to
Qwen3.5-35B-A3B and Qwen3.5-122B-A10B on ScienceQA, MMMU-Pro, and MathVista,
show that visual-token pruning changes expert assignments among retained tokens.
Across the tested pruning budgets, L2 selection consistently produces lower
routing change than random selection, but this ordering does not consistently
translate into higher task accuracy. Other interactions involving cache management
and hardware behaviour remain inferences or open hypotheses.

To complement the literature synthesis, we conduct two empirical studies. The
first measures routing changes under visual-token pruning. The second introduces
Temporal Routing Consistency (TRC), a diagnostic for expert-assignment overlap
across adjacent visual inputs, and validates it against sequence-order and
clip-coherence controls on DeepSeek-VL2-Tiny and InternVL3.5-30B-A3B. TRC responds
in the expected direction for both models, but natural variation in TRC does not
show a reliable association with answer correctness in either evaluated sample.

Beyond cataloguing methods, the survey synthesizes the design tensions that shape
deployable multimodal systems, including accuracy versus token budget, static
versus content-adaptive compression, sparse routing efficiency versus expert
collapse, low-bit serving versus modality-specific degradation, and benchmark
convenience versus end-to-end edge realism. The paper highlights the need to
evaluate not only individual efficiency techniques, but also how interventions
at one stage can alter downstream model behaviour.

\end{reviewchange}
\end{abstract}

\newpage

\section{Introduction}
\label{sec:intro}

The rapid scaling of vision-language models has produced systems of remarkable
capability, but capability has come at a cost that is no longer simply computational:
it is systemic.  Models that jointly process high-resolution image tiles, dense video
frame sequences, OCR spans, and long text contexts routinely generate token sequences
of tens of thousands of elements.  Inference on consumer or edge hardware therefore
may require visual token compression, KV-cache optimisation, Mixture-of-Experts (MoE)
routing, quantization, and hardware-aware serving to operate together rather than as
independent optimisations.

\begin{reviewchange}
These techniques can have coupled failure modes.  Published work shows that low-bit
quantization can change MoE routing, while links from upstream interventions to cache
traffic and hardware behaviour remain largely untested.  Edge-memory effects also
depend on the model, workload, and device
~\citep{shao2026surveytokencompressionefficient,liu2025surveyinferenceoptimizationtechniques,fu2026eaquantenhancingposttrainingquantization,zhong2026breakingmodalityheterogeneitylowbit,10906629}.
These observations motivate a proposed failure-propagation framework in which a local
intervention may become a distribution shift, routing error, quantization fault, or
evidence-retention problem at a later stage.

\end{reviewchange}

Three converging pressures make this problem urgent.  First, \emph{token density}: one
image at size $336\!\times\!336$ processed by CLIP-ViT-L/14 with a patch size of
$14\!\times\!14$ creates 576 tokens, while tiling techniques produce thousands.
Visual encodings contain less average information per token than text
targets~\citep{shao2026surveytokencompressionefficient}; however, deleting or merging those tokens changes the
representation on which downstream routing, quantization, and caching policies were
learned.  Second, \emph{sparse-model overhead}: in MoE architectures, only a fraction
of parameters is active per token, but that sparsity creates irregular memory access,
load imbalance, and collapse risk~\citep{liu2025surveyinferenceoptimizationtechniques,sha2026sparsemixtureofexpertsroutingvisual}.  When compression precedes routing, this
fragile balance is may perturbed by altered token statistics.  Third,
\emph{deployment constraints}: mobile platforms~\citep{li2025palmbenchcomprehensivebenchmarkcompressed}, edge AI
accelerators~\citep{chittyvenkata2024llminferencebenchinferencebenchmarkinglarge}, and time-sensitive energy-constrained embedded
systems impose tight constraints on latency, memory usage, and power consumption. As
inference energy usage may scale super-linearly with the level of accuracy achieved,
efficiency becomes a primary concern~\citep{yang2024doubleexponentialincreasesinferenceenergy}.

The review traces this coupled landscape from foundational redundancy principles in
Vision Transformers~\citep{he2021maskedautoencodersscalablevision,jaegle2021perceivergeneralperceptioniterative,tong2022videomaemaskedautoencodersdataefficient} through KV-cache management,
MoE routing, quantization, edge deployment, and hardware-aware evaluation
benchmarks~\citep{kwon2023efficientmemorymanagementlarge,tschand2025mlperfpowerbenchmarkingenergy,John_2024,sukthanker2024hwgptbenchhardwareawarearchitecturebenchmark}.  The review is
organized around five primary themes, summarised in Figure~\ref{fig:taxonomy-tree}.

\begin{figure}[h]
\centering
\begingroup
\centering
\resizebox{\textwidth}{!}{%
\begin{tikzpicture}[
  x=1mm,
  y=1mm,
  every node/.style={font=\sffamily},
  rootbox/.style={
    draw=black!70,
    line width=0.7pt,
    rounded corners=2pt,
    fill=violet!12,
    align=center,
    font=\sffamily\bfseries\mdseries,
    text width=50mm,
    minimum height=16mm,
    inner sep=2.4pt,
    rotate=90
  },
  family/.style args={#1}{
    draw=#1!75!black,
    line width=0.65pt,
    rounded corners=2pt,
    fill=#1!17,
    align=center,
    font=\sffamily\bfseries\small,
    text width=32mm,
    minimum height=12mm,
    inner sep=2.4pt
  },
  subfamily/.style args={#1}{
    draw=#1!75!black,
    line width=0.6pt,
    rounded corners=2pt,
    fill=#1!10,
    align=center,
    font=\sffamily\bfseries\scriptsize,
    text width=42mm,
    minimum height=12.5mm,
    inner sep=2.4pt
  },
  detail/.style args={#1}{
    draw=#1!75!black,
    line width=0.55pt,
    rounded corners=2pt,
    fill=#1!4,
    align=left,
    font=\sffamily\scriptsize,
    text width=76mm,
    minimum height=16mm,
    inner sep=2.8pt
  },
  connector/.style={draw=black!55,line width=0.45pt}
]

\node[rootbox] (root) at (0,0)
  {Efficient Multimodal\\Intelligence at\\Scale \& Edge};

\node[family=cyan]   (vision) at (34,88)  {Vision Encoder\\Sparsification};
\node[family=orange] (mllm)   at (34,44)  {MLLM Sequence\\Compression};
\node[family=yellow] (kv)     at (34,0)   {KV-Cache\\Optimisation};
\node[family=green]  (moe)    at (34,-44) {MoE Routing\\Dynamics};
\node[family=gray]   (edge)   at (34,-88) {Edge Quantization\\\& Systems};

\node[subfamily=cyan]   (s1)  at (86,99)  {Bipartite and Salience-Aware Merging};
\node[subfamily=cyan]   (s2)  at (86,77)  {Self-Supervised Redundancy};
\node[subfamily=orange] (s3)  at (86,55)  {Multi-Stage Dropping};
\node[subfamily=orange] (s4)  at (86,33)  {Search and Learned Pruning};
\node[subfamily=yellow] (s5)  at (86,11)  {Graph-Based KV Eviction};
\node[subfamily=yellow] (s6)  at (86,-11) {State-Space / Semantic Cache};
\node[subfamily=green]  (s7)  at (86,-33) {Routing Stability};
\node[subfamily=green]  (s8)  at (86,-55) {Temporal Routing Consistency};
\node[subfamily=gray]   (s9)  at (86,-77) {Routing-Consistent PTQ};
\node[subfamily=gray]   (s10) at (86,-99) {Hot/Cold Expert Storage};

\node[detail=cyan] (d1) at (154,99)
  {\textbf{Representative methods:} ToMe, ATM, AdaMerge, ToSA, MaMe/MaRe, VisionTrim\\
   \citep{bolya2023tokenmergingvitfaster,Lee_2025,lee2026adamergesalienceawareadaptivetoken,huang2025tosatokenmergingspatial,huo2026mamemarematrixbased,yu2026visiontrimunifiedvisiontoken}};
\node[detail=cyan] (d2) at (154,77)
  {\textbf{Redundancy evidence:} MAE, VideoMAE, Perceiver\\
   \citep{he2021maskedautoencodersscalablevision,tong2022videomaemaskedautoencodersdataefficient,jaegle2021perceivergeneralperceptioniterative,wu2022rapredundancyawarevideolanguagepretraining,Li_2023}};
\node[detail=orange] (d3) at (154,55)
  {\textbf{Pipeline-aware dropping:} MustDrop, HiRED, ZipVL, SparseVLM\\
   \citep{liu2024multistagevisiontokendropping,arif2024hiredattentionguidedtokendropping,he2024zipvlefficientlargevisionlanguage,zhang2025sparsevlmvisualtokensparsification}};
\node[detail=orange] (d4) at (154,33)
  {\textbf{Selection policies:} GreedyPrune, EvoComp, FastVID, CS-VLM\\
   \citep{pei2025greedypruneretentingcriticalvisual,song2026evocomplearningvisualtoken,shen2025fastviddynamicdensitypruning,kiruluta2025csvlmcompressedsensingattention}};
\node[detail=yellow] (d5) at (154,11)
  {\textbf{Dependency-aware eviction:} GraphKV, MadaKV, H$_2$O, SnapKV, PyramidKV\\
   \citep{li2025graphkvbreakingstaticselection,li2025madakvadaptivemodalityperceptionkv,zhang2023h2oheavyhitteroracleefficient,li2024snapkvllmknowslooking,cai2025pyramidkvdynamickvcache}};
\node[detail=yellow] (d6) at (154,-11)
  {\textbf{Cache compression:} RetentiveKV, PagedAttention, ChunkKV, StreamingLLM\\
   \citep{liu2026retentivekvstatespacememoryuncertaintyaware,kwon2023efficientmemorymanagementlarge,liu2025chunkkvsemanticpreservingkvcache,xiao2024efficientstreaminglanguagemodels}};
\node[detail=green] (d7) at (154,-33)
  {\textbf{Stable routing:} Auxiliary-free balancing, ReMoE, Eigen-router, Spectral regularization\\
   \citep{wang2024auxiliarylossfreeloadbalancingstrategy,wang2025remoefullydifferentiablemixtureofexperts,do2026eigenvectorsexpertstrainingfreenoncollapsing,delibasoglu2026spectralmanifoldregularizationstable}};
\node[detail=green] (d8) at (154,-55)
  {\textbf{Video routing pressure:} DynTok, V2Drop, HERMES, LVC, MMViR\\
   \citep{zhang2025dyntokdynamiccompressionvisual,chen2026variationawarevisiontokendropping,faure2025hermestemporalcoherentlongformunderstanding,wang2025lvclightweightcompressionframework,li2026mmvirmultimodalmultigranularityrepresentation}};
\node[detail=gray] (d9) at (154,-77)
  {\textbf{Quantized routing:} EAQuant, VSRAQ, MoQE, GEMQ, MoBiE, VLMQ\\
   \citep{fu2026eaquantenhancingposttrainingquantization,park2026valueandstructurealignmentroutingconsistentquantization,kim2023mixturequantizedexpertsmoqe,deng2026gemqglobalexpertlevelmixedprecision,zhao2026mobieefficientinferencemixture,xue2026vlmqtokensaliencydrivenposttraining}};
\node[detail=gray] (d10) at (154,-99)
  {\textbf{On-device serving:} EdgeMoE, D2MoE, ZipMoE, ExpertFlow, Janus\\
   \citep{yi2025edgemoeempoweringsparselarge,Wang_2025,yang2026zipmoeefficientondevicemoe,shen2025expertflowadaptiveexpertscheduling,zhang2026janusdisaggregatingattentionexperts}};

\draw[connector] (root.south) -- ++(6,0) |- (vision.west);
\draw[connector] (root.south) -- ++(6,0) |- (mllm.west);
\draw[connector] (root.south) -- ++(6,0) -- (kv.west);
\draw[connector] (root.south) -- ++(6,0) |- (moe.west);
\draw[connector] (root.south) -- ++(6,0) |- (edge.west);

\draw[connector] (vision.east) -- ++(6,0) |- (s1.west);
\draw[connector] (vision.east) -- ++(6,0) |- (s2.west);
\draw[connector] (mllm.east) -- ++(6,0) |- (s3.west);
\draw[connector] (mllm.east) -- ++(6,0) |- (s4.west);
\draw[connector] (kv.east) -- ++(6,0) |- (s5.west);
\draw[connector] (kv.east) -- ++(6,0) |- (s6.west);
\draw[connector] (moe.east) -- ++(6,0) |- (s7.west);
\draw[connector] (moe.east) -- ++(6,0) |- (s8.west);
\draw[connector] (edge.east) -- ++(6,0) |- (s9.west);
\draw[connector] (edge.east) -- ++(6,0) |- (s10.west);

\foreach \s/\d in {s1/d1,s2/d2,s3/d3,s4/d4,s5/d5,s6/d6,s7/d7,s8/d8,s9/d9,s10/d10}
  \draw[connector] (\s.east) -- (\d.west);

\end{tikzpicture}%
}
\endgroup
\begin{reviewchange}
\caption{Taxonomy of the five operational bottlenecks reviewed in this survey, with
representative mechanisms and methods.  The rows organise the literature; they do not
imply that every method was evaluated jointly with the downstream stages.}
\label{fig:taxonomy-tree}
\end{reviewchange}
\end{figure}

\begin{reviewchange}
To distinguish the strength of the available evidence, we classify cross-stage
interactions into three categories.  \emph{Direct evidence} denotes a relationship
measured in a cited study or in our original experiments.  A \emph{supported inference} is a conclusion drawn
from multiple published results but not directly evaluated as the complete interaction
claimed.  \emph{Open hypotheses} are conceptual mechanisms, diagnostics, or
directions proposed in this survey that remain to be empirically validated.  We use
this distinction throughout the paper when discussing interactions among compression,
routing, quantization, caching, and hardware execution; formal definitions and signalling
rules appear in Section~\ref{sec:evidence-classification}.

This paper differs from prior surveys by treating compression, routing, quantization,
cache management, and hardware constraints as a connected deployment pipeline rather
than as parallel technique families.  Its main contributions are:
\begin{enumerate}[leftmargin=2em, itemsep=4pt]
  \item \textbf{A proposed failure-propagation taxonomy.}  We organize
  compression-induced representation shift, routing inconsistency under low-bit
  quantization, and modality-heterogeneous quantization degradation into a single
  conceptual chain.  The framework identifies testable interfaces at which optimising
  one stage may leave downstream failure modes unaddressed; for example,
  routing-aware post-training quantization
  recovers $1.15\%$--$2.28\%$ average score in low-bit MoE settings~\citep{fu2026eaquantenhancingposttrainingquantization}, while
  modality-aware VLM quantization retains $93.5\%$ of FP16 performance at W3A3
  (69.5 vs.\ 74.3)~\citep{zhong2026breakingmodalityheterogeneitylowbit}.

  \item \textbf{Temporal routing diagnostic for video MoE models.}
  \emph{Temporal Routing Consistency} (TRC) measures overlap between the experts that
  receive the most routing mass from adjacent frames, separating routing continuity
  from downstream task quality. We validate TRC against sequence-order and clip-coherence controls on NExT-QA. TRC responds as expected to these controls, but natural variation in TRC does not predict answer correctness in the evaluated sample.

  \item \textbf{A co-design view of multimodal efficiency.}  We characterise
  compression ratio, routing behaviour, quantization precision, KV-cache residency, and
  hardware capacity as mutually constrained variables that shape the achievable
  accuracy-efficiency Pareto frontier on a given deployment target.

  \item \textbf{Systematic coverage of recent efficient multimodal inference work.}
  We review over one hundred contributions published primarily from 2020--2026, with
  earlier foundational references where needed, across visual token compression,
  multi-stage MLLM token management, video redundancy, KV-cache optimisation, MoE
  routing and serving, quantization, edge deployment, and hardware-aware benchmarking.

  \item \textbf{Two empirical studies.}  We complement the literature synthesis with two experiments that test central claims of our framework. On DeepSeek-VL2-Tiny and InternVL3.5-30B-A3B, we directly measure compression-induced expert reassignment under visual-token pruning on ScienceQA and validate the TRC diagnostic introduced above on NExT-QA. We further extend the pruning measurement to Qwen3.5-35B-A3B and Qwen3.5-122B-A10B on ScienceQA, MMMU-Pro, and MathVista (Section~\ref{sec:interface_b-larger-models}). The evaluations use model-specific preprocessing, precision, and routing spaces, so we report within-model effects rather than treating expert identities or absolute metric levels as directly comparable across architectures.
\end{enumerate}
\end{reviewchange}

Compared with surveys focused on MoE inference~\citep{liu2025surveyinferenceoptimizationtechniques}
or multimodal token compression~\citep{shao2026surveytokencompressionefficient}, our
unit of analysis is the interface between optimisations.

\begin{table}[t]
\centering
\caption{Master optimization matrix for efficient multimodal inference.  The table
maps the dominant model class to the phase where optimization is applied, the primary
system bottleneck, and the survey sections where each design axis is analysed.}
\label{tab:master-optimization-matrix}
\small
\setlength{\tabcolsep}{4pt}
\begin{tabular}{p{0.20\linewidth} p{0.20\linewidth} p{0.28\linewidth} p{0.22\linewidth}}
\toprule
\textbf{Model class} & \textbf{Optimization phase} & \textbf{Dominant bottleneck} & \textbf{Representative design axis} \\
\midrule
Vision Transformers & Encoder token processing & Quadratic attention over dense visual patches & Merge, prune, or reweight redundant spatial tokens \\
Multimodal LLMs & Vision-to-language prefilling & Visual-token overload inside the LLM context & Stage-aware dropping, query-conditioned sparsification, positional preservation \\
Video MLLMs & Temporal token construction & Redundant frames plus expert jitter across adjacent frames & Dynamic frame budgets, variation-aware dropping, temporal routing consistency \\
Long-context MLLMs & Autoregressive decoding & KV-cache memory growth and modality-heterogeneous attention & Graph-aware eviction, modality-adaptive cache policies, state-space memory \\
MoE Transformers & Routing and expert execution & Load imbalance, expert collapse, and irregular memory movement & Differentiable routing, batch-level balancing, expert scheduling \\
Quantized VLM/MoE models & Deployment calibration & Activation outliers, routing inconsistency, and modality-specific precision sensitivity & Expert-aware calibration, structure-preserving routing, mixed precision \\
Edge multimodal systems & On-device serving & SRAM limits, flash/DRAM transfers, energy per byte, and thermal constraints & Hot-cold storage, prefetching, cache-affinity scheduling, hardware-aware benchmarking \\
\bottomrule
\end{tabular}
\end{table}
Table~\ref{tab:master-optimization-matrix} maps each model class to its dominant
bottleneck and intervention point.
\begin{table}[!htbp]
\centering
\caption{Comparison of visual token compression methods. Speedup figures are as
reported in the respective papers and are not directly comparable across base models
or hardware.  The final column records an unresolved downstream evaluation issue.}
\label{tab:compression-methods}
\small
\renewcommand{\arraystretch}{1.15}
\setlength{\tabcolsep}{4pt}
\begin{tabular}{p{0.15\linewidth} p{0.22\linewidth} p{0.13\linewidth} p{0.25\linewidth} p{0.17\linewidth}}
\toprule
\textbf{Method} & \textbf{Approach and stage} & \textbf{Reported speedup} & \textbf{Key limitation} & \textbf{Downstream evaluation gap} \\
\midrule
ToMe \citep{bolya2023tokenmergingvitfaster} & Bipartite merging in ViT encoder & $2\times$ throughput & Fixed threshold; no spatial awareness & Routing after merged representations was not evaluated \\
ATM \citep{Lee_2025} & Adaptive layer merging in ViT encoder & ${>}30\%$ FLOP reduction & Depth-wise only; no query conditioning & Downstream routing was not evaluated \\
AdaMerge \citep{lee2026adamergesalienceawareadaptivetoken} & Salience-weighted merging in ViT encoder & ${\sim}2\times$ & Saliency heuristic may misfire on texture-heavy inputs & Effect of retained representations on routing is unknown \\
MaMe/MaRe \citep{huo2026mamemarematrixbased} & Matrix-decomposition merge in ViT and diffusion models & $31\%$ latency reduction & Designed for generative models; limited MLLM evaluation & Evaluation with MoE routers is unavailable \\
ToSa \citep{huang2025tosatokenmergingspatial} & Spatial-aware merging in ViT encoder & Comparable to ToMe & Spatial penalty adds overhead at high resolution & Downstream spatial reasoning requires broader evaluation \\
MustDrop \citep{liu2024multistagevisiontokendropping} & Multi-stage dropping in MLLM pipeline & Large FLOP reduction & Stage schedule requires per-model tuning & Downstream effects of altered attention context are unclear \\
HiRED \citep{arif2024hiredattentionguidedtokendropping} & Early-layer saliency pruning in MLLM prefill & Substantial throughput gain & Saliency from early layers may miss semantic importance & Later-stage robustness requires evaluation \\
GreedyPrune \citep{pei2025greedypruneretentingcriticalvisual} & Marginal-gain selection in MLLM encoder & High accuracy at aggressive ratios & Combinatorial search cost grows with token count & Retained-set distribution shifts were not measured \\
EvoComp \citep{song2026evocomplearningvisualtoken} & Evolutionary learned compression in MLLM & Avoids positional gaps & Additional training procedure required & Co-adaptation with downstream modules is unclear \\
CSVLM \citep{kiruluta2025csvlmcompressedsensingattention} & Compressed-sensing projection in MLLM encoder & Formal sparsity guarantees & Sparse-signal assumption may not fit dense scenes & Routing under projected inputs requires evaluation \\
FastMMoE \citep{xia2025fastmmoe} & Routing-aware pruning and dynamic expert activation in MoE MLLMs & $21.09$--$63.66\%$ FLOP reduction & Does not report baseline-to-pruned expert-assignment change & Routing guides pruning, but downstream reassignment is unreported \\
DynTok \citep{zhang2025dyntokdynamiccompressionvisual} & Dynamic frame budgeting in video MLLM & Higher action-segment quality & Complexity estimator adds per-clip overhead & Temporal routing effects are unknown \\
V2Drop \citep{chen2026variationawarevisiontokendropping} & Variation-based frame dropping in video MLLM & Large latency reduction for video & One-step window; no temporal memory & Routing continuity was not evaluated \\
\bottomrule
\end{tabular}
\end{table}
Table~\ref{tab:compression-methods} compares representative token-reduction methods and
marks cross-stage effects that remain untested.
\begin{reviewchange}
After the methodology and background, the survey follows the pipeline from token
compression through video, KV cache, MoE routing, quantization, edge deployment, and
benchmarking, then closes with open problems.
\end{reviewchange}

\begin{reviewchange}
\section{Review Methodology}
\label{sec:methodology}

We use a structured narrative review because models, hardware, and efficiency metrics
vary too widely for a defensible statistical meta-analysis.  The protocol fixes the
scope, search window, screening procedure, and evidential language used below.

Beyond the literature synthesis, we empirically examine two interactions highlighted by the framework: compression-induced routing disruption
(Section~\ref{sec:interface_b-empirical}) and TRC under sequence controls
(Section~\ref{sec:trc-empirical}).  Their datasets, controls, metrics, confidence
intervals, and scope limits are reported in those sections.

\subsection{Scope and Inclusion Criteria}

We include inference work on visual-token compression, KV-cache management, MoE
routing and serving, quantization, edge deployment, and relevant benchmarks when the
paper reports a mechanism or measurable quality or systems outcome.  We exclude
training-only acceleration and generic compression without an inference connection.
Text-only KV-cache and MoE-serving methods are included only when they provide a basis
for multimodal deployment, and their transfer to MLLMs is identified explicitly.

\subsection{Search Sources, Keywords, and Date Range}

We searched Google Scholar and Semantic Scholar, then verified primary records in
arXiv, OpenReview, ACL Anthology, PMLR, CVF Open Access, ACM Digital Library, and IEEE
Xplore.  Queries paired model terms (ViT, VLM/MLLM, video Transformer, and MoE) with
token compression, KV cache, routing, serving, quantization, edge/on-device inference,
latency, memory, energy, and benchmark terms.  Cross-stage queries included
``compression routing'', ``routing-consistent quantization'', ``multimodal KV cache'',
and ``hardware-aware MoE''.  The main window was January 2020--June 2026; earlier work
was added only when needed for a foundational mechanism or benchmark.

\subsection{Six-Domain Discovery Corpus}

Discovery used the six overlapping domains in Table~\ref{tab:domain-inventory}; a paper
could appear in more than one domain.

\begin{table}[t]
\centering
\caption{Six-domain discovery structure used for the structured narrative review.}
\label{tab:domain-inventory}
\small
\renewcommand{\arraystretch}{1.12}
\setlength{\tabcolsep}{4pt}
\begin{tabular}{p{0.12\linewidth} p{0.78\linewidth}}
\toprule
\textbf{Domain} & \textbf{Discovery scope} \\
\midrule
1 & Efficient MLLMs, visual-token reduction, and long-video compression \\
2 & Multimodal and vision MoE architectures, adaptation, and applications \\
3 & Edge/mobile LLM and MoE compression, serving, and deployment \\
4 & Video-LLM compression, temporal modelling, and heterogeneous experts \\
5 & MoE routing, load balancing, expert placement, and systems efficiency \\
6 & Multimodal reasoning, document, video, human-centric, search, and coding benchmarks \\
\bottomrule
\end{tabular}
\end{table}

Duplicates and multiple versions were consolidated.  The inventory is a discovery
pool, not a closed PRISMA corpus, because citation chaining continued during synthesis.

\subsection{Study Selection and Screening Procedure}

Screening proceeded through title/abstract relevance and deduplication, full-text
checks of the intervention, model, dataset, baseline, metric, and hardware, and
forward/backward citation chaining.  We prioritised recent, application-relevant work
with clear baselines and quantitative results; older or conceptual papers were retained
when foundational.  The most complete public version available by June 2026 was used,
and measured claims were tied to primary studies where possible.  Citation visibility
helped prioritise candidates but was not a threshold; because no frozen citation-count
snapshot was retained, that part of the ranking is not exactly reproducible.

\subsection{Evidence-Status Signalling}
\label{sec:evidence-classification}

Directly measured relationships are stated as findings, indirect combinations as
inferences, and untested mechanisms as hypotheses.  The label applies to each claim,
not an entire paper or interface.  Preprints are eligible, but their results are
attributed to the authors and are not treated as independent replication.
\end{reviewchange}

\section{Background and Preliminaries}
\label{sec:background}

Efficient multimodal deployment rests on three architectural facts: visual inputs are
converted into dense token sequences, multimodal LLMs pass those tokens through language
decoders, and MoE variants replace dense feed-forward computation with routed expert
banks.  These facts matter because they create the main efficiency levers studied in
this survey: visual tokens can be compressed, expert computation can be sparsified, and
hardware placement can reduce data movement.  On edge devices, however, the same levers
also create failure conditions: compression shifts the token distribution, sparse
routing can become imbalanced under shifted inputs, and memory hierarchies can make
expert and cache movement more expensive than arithmetic.  This section fixes the
architectural vocabulary used by the later failure analyses.

\subsection{Vision Transformers and Multimodal Architectures}

Vision Transformers (ViTs) apply the multi-head self-attention mechanism to sequences
of image patches, treating each patch as an independent token. The spatial redundancy of this representation is substantial: masked reconstruction pre-training produces competitive visual representations even when 75\%--85\% of input patches are hidden, because each visible patch carries information about its neighbours~\citep{he2021maskedautoencodersscalablevision}.
Video data exhibits even stronger redundancy, with effective pre-training possible at
masking ratios near $90\%$ because temporal continuity makes adjacent frames highly
predictable~\citep{tong2022videomaemaskedautoencodersdataefficient}.  An architectural alternative is to avoid full
quadratic attention over the input sequence altogether by cross-attending a compact
latent array to arbitrary high-dimensional inputs~\citep{jaegle2021perceivergeneralperceptioniterative}.

Modern MLLMs couple a visual encoder with a pre-trained LLM decoder through a
connector.  In common architectures, visual and textual positions both contribute to
sequence processing even though their information density may differ.  This motivates
compression.  Compression also changes the sequence or representations supplied to
later layers; its effects on MoE routing and quantization are classified separately in
Section~\ref{sec:codesign}.

\subsection{Mixture-of-Experts: Foundations and Fragility}

An MoE replaces a single FFN with $N$ expert subnetworks and a router that selects a
subset, usually Top-$k$, for each token.  Parameter count scales with $N$, while
per-token computation scales primarily with $k$~\citep{du2024revisitingmoedensespeedaccuracy}.
This conditional computation also creates routing, placement, batching, and
memory-transfer costs, so inference optimization cannot be reduced to FLOPs alone
~\citep{liu2025surveyinferenceoptimizationtechniques}.

Formally, for an input token representation $x$, a router computes logits
$h(x)=W_r x$ and selects the active expert set through Top-$k$ gating:
\[
\mathrm{Router}(x)=\mathrm{TopK}(\mathrm{softmax}(h(x)), k).
\]
The gated MoE output is then
\[
y(x)=\sum_{i\in \mathrm{TopK}(x)} g_i(x)E_i(x),
\qquad
g_i(x)=\mathrm{softmax}(h(x))_i,
\]
where $E_i$ is the $i$-th expert FFN and $g_i(x)=0$ for unselected experts.  A commonly
used reference point for balanced Top-$k$ routing is uniform expected utilisation,
\[
\mathbb{E}_{x}\!\left[\mathbf{1}\{i\in \mathrm{TopK}(x)\}\right]\approx \frac{k}{N}
\quad \forall i.
\]

Router behaviour is learned under a training distribution and may change under a
deployment shift.  {Under the low-bit settings evaluated by
\citet{fu2026eaquantenhancingposttrainingquantization}, quantizing router logits changes
expert assignments.  The compression-to-routing pathway and the joint effect of
compression and quantization are less extensively evaluated and remain important
cross-model and cross-modality questions.}

\begin{figure}[t]
\centering
\includegraphics[width=\textwidth]{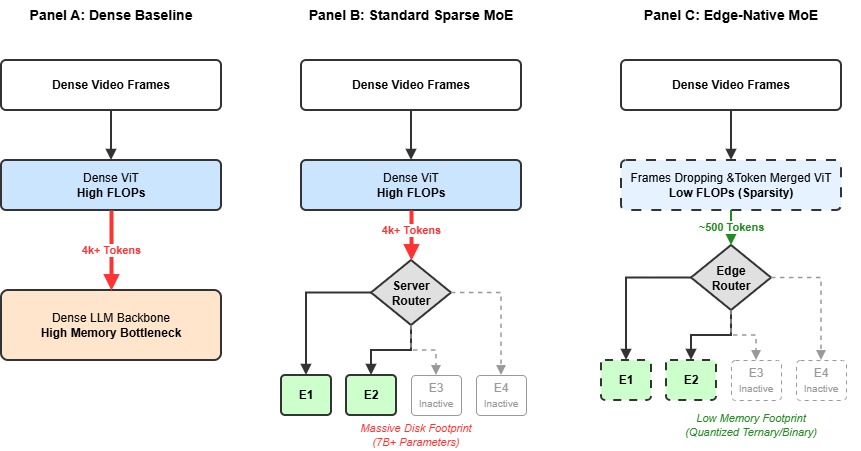}
\caption{Illustrative architectural paradigms for multimodal efficiency. \textbf{(Panel A)} Dense baselines can incur visual-token and memory costs. \textbf{(Panel B)} Server-side MoE designs reduce active expert computation while retaining the storage and movement costs of the expert bank. \textbf{(Panel C)} A proposed edge-oriented design combines upstream token compression with low-precision experts; its end-to-end benefit depends on the model, workload, and device.}
\label{fig:architectural-evolution}
\end{figure}

The primary training-time manifestation of this fragility is \emph{load imbalance}:
without regularisation, routers tend to route tokens to a small subset of experts, a
failure mode termed \emph{expert collapse}~\citep{sha2026sparsemixtureofexpertsroutingvisual,do2026eigenvectorsexpertstrainingfreenoncollapsing}.
As an operational diagnostic, this survey flags utilisation below the uniform
per-expert share expected under balanced Top-$k$ routing, $k/N$; for a 64-expert
model with $k=2$, $k/N=3.125\%$ (equivalently $1/N=1.5625\%$ only in the
special case $k=1$).  This reference is not a universal definition of collapse. In the evaluated setting, collapse can propagate layerwise,
with roughly one-third of sparse MoE layers drifting toward single-expert behaviour
~\citep{sha2026sparsemixtureofexpertsroutingvisual}.  Classical auxiliary load-balancing losses address collapse
through explicit regularisation, but they introduce training instability and
loss-weight tuning; recent work therefore explores auxiliary-loss-free balancing
~\citep{wang2024auxiliarylossfreeloadbalancingstrategy}.  Understanding how compression and quantization interact with
these already-fragile routing dynamics is one of the central open problems this survey
identifies.

\subsection{Edge Deployment: A Hierarchy of Constraints}

Deploying large models on resource-constrained hardware is not simply a scaled-down
version of cloud serving.  It is a qualitatively different problem governed by a
hierarchy of binding constraints that interact with every upstream design decision.

At the tightest level, on-chip memory capacity limits what can remain in fast storage
during a forward pass.  When an expert bank does not fit, weights may be loaded from a
slower memory tier.  DRAM bandwidth then influences transfer time.  On the memory-bound
workloads studied by \citet{chung2026joulesgodiagnosinginference}, data movement is a
major contributor to energy and latency; the balance is device- and workload-specific.
Reducing stored precision reduces bytes per weight, although realised latency and
energy gains also depend on kernel support and packing overhead.  Energy and thermal
limits can further make sustained throughput lower than a short peak measurement.

{These hardware constraints may also interact with upstream choices.}  Fewer
tokens mean fewer routing decisions, but the number of distinct expert loads also
depends on routing locality and cache state.  Compression-induced changes to that
locality have not been measured directly.  Quantization reduces bytes per stored
weight and can change expert selection in the cited MoE experiments
~\citep{fu2026eaquantenhancingposttrainingquantization}; whether that change causes
prefetch misses on a given device is an open systems question.  Consequently, a
configuration tuned on server hardware should be re-measured after composition on the
target edge device.  Evaluation resources include mobile compressed-model
benchmarking~\citep{li2025palmbenchcomprehensivebenchmarkcompressed}, architecture-level accelerator simulation
~\citep{sukthanker2024hwgptbenchhardwareawarearchitecturebenchmark}, direct power-performance characterisation~\citep{John_2024}, and
energy-aware measurement spanning $\mu$Watt embedded targets to MWatt datacentres
~\citep{tschand2025mlperfpowerbenchmarkingenergy}.  Figure~\ref{fig:architectural-evolution} summarises the
resulting architectural progression from dense baselines to server-side MoE models and
edge-native sparse systems that combine compression with low-precision experts.

\section{Conceptual Framework for Cross-Stage Interactions}
\label{sec:codesign}
{This section organizes potential interactions among token compression,
MoE routing, quantization, and edge execution into a conceptual framework rather than
a validated causal model.}  It identifies testable pathways through four operational
stages: compression, routing, precision selection, and hardware execution.  Each technique addresses a distinct
bottleneck and has often been evaluated under a separate objective.  When techniques
are composed, however, the output of one stage becomes part of the operating condition
of the next.  We call the resulting set of potential dependencies the \emph{proposed
failure-propagation framework}; evidence is attached to individual links rather than
to the chain as a whole.

\begin{figure}[t]
\centering
\includegraphics[width=\textwidth]{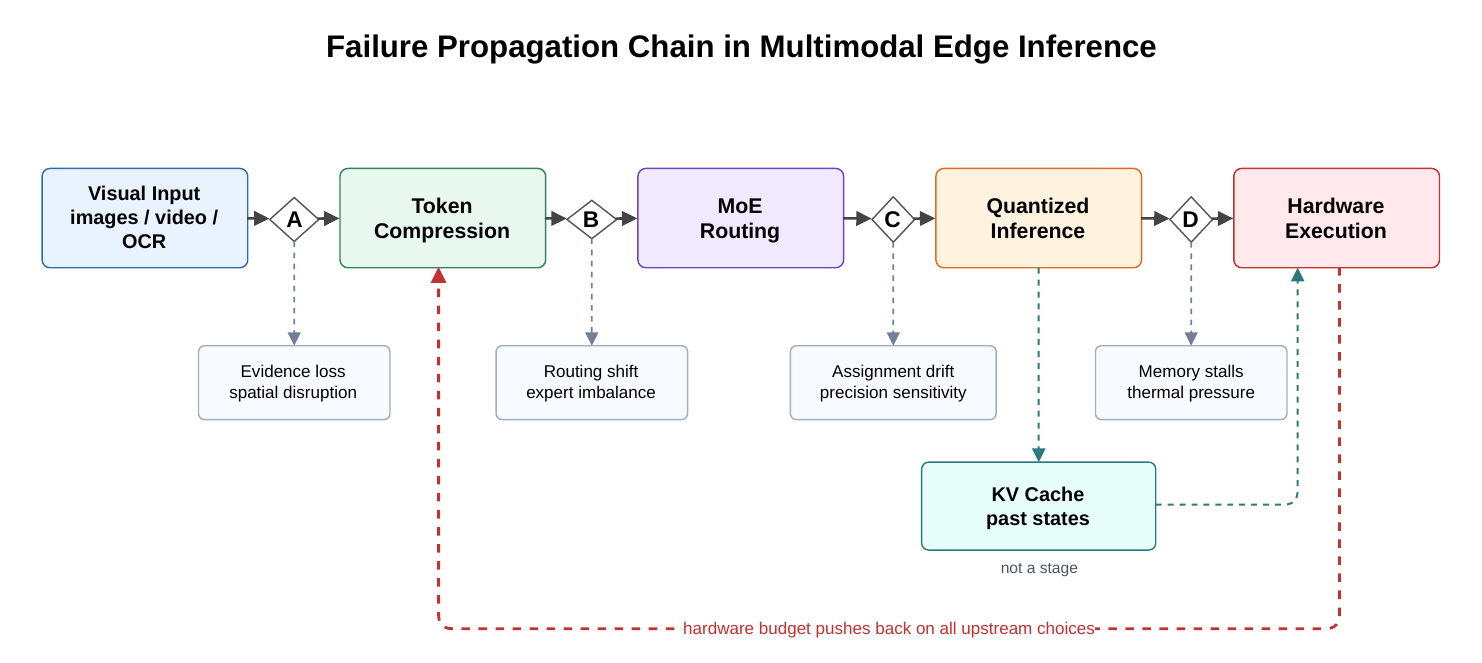}
\begin{reviewchange}
\caption{Proposed conceptual framework illustrating potential failure-propagation
pathways.  Interfaces A--D identify, respectively, the input-to-compression,
compression-to-routing, routing-to-precision, and model-to-hardware boundaries.  Solid
arrows denote directly studied stage relationships; dashed arrows denote inferred or
proposed interactions requiring joint evaluation.  The KV cache is shown as persistent
state whose size and retention policy feed hardware pressure.}
\label{fig:chain}
\end{reviewchange}
\end{figure}

\begin{reviewchange}
\subsection{Interfaces A--D: Definitions}
\label{sec:chain}

Figure~\ref{fig:chain} illustrates the four-stage inference pipeline and four
interfaces at which cross-technique failures may emerge.  The KV cache is drawn outside
the main left-to-right chain because it is persistent inference state rather than a
separate transformation stage; its footprint can nevertheless couple earlier token
and precision choices to later hardware pressure.
\end{reviewchange}

We write the composed inference path as

\begin{equation}
  \mathcal{P}
  \;=\;
  \mathcal{H}\!\circ\!\mathcal{Q}_{\boldsymbol{\pi}}
  \!\circ\!\mathcal{R}_{\boldsymbol{\theta}}
  \!\circ\!\mathcal{C}_{\boldsymbol{\phi}},
  \label{eq:pipeline}
\end{equation}

where $\mathcal{C}$, $\mathcal{R}$, $\mathcal{Q}$, and $\mathcal{H}$ denote compression,
routing, quantization, and hardware execution.  Their parameters cover the token
policy, router and expert capacity, precision and calibration, and device constraints.
Equation~\eqref{eq:pipeline} specifies order only; it makes no causal or independence
claim.  Each interface is summarized by what crosses it, how it can fail, and its
evidence status.

\paragraph{Interface A - Visual Input to Compression.}
Dense visual tokens cross this interface.  Dropping, merging, and positional changes
can remove spatial or task evidence; these effects are directly studied in
Sections~\ref{sec:redundancy}--\ref{sec:mlhm}.

\paragraph{Interface B - Token Compression to MoE Routing.}
A shorter or altered token sequence reaches the router and may change expert
assignments or load.  Before this survey, FastMMoE used routing similarity to guide
pruning but did not quantify downstream expert reassignment~\citep{xia2025fastmmoe}.
Section~\ref{sec:interface_b-empirical}
provides direct measurements on DeepSeek-VL2-Tiny and InternVL3.5-30B-A3B on
ScienceQA, and Section~\ref{sec:interface_b-larger-models} extends them to two
larger Qwen3.5 MoE-VLMs and two harder benchmarks; replication across further models,
modalities, and compression methods remains open.

\paragraph{Interface C - MoE Routing to Quantized Inference.}
Selected experts and gate weights determine which tensors are executed.  Quantized
router logits can change those selections, and routing-aware quantization recovers
$1.15\%$--$2.28\%$ average score in the evaluated settings
\citep{fu2026eaquantenhancingposttrainingquantization,park2026valueandstructurealignmentroutingconsistentquantization}.
Whether upstream pruning amplifies this effect remains unknown.

\begin{reviewchange}
\paragraph{Interface D - Quantized Inference to Hardware Execution.}
Quantized tensors and kernels cross this boundary.  Lower precision need not reduce
latency or energy when conversion, unsupported kernels, memory traffic, or thermal
limits dominate
\citep{chung2026joulesgodiagnosinginference,tschand2025mlperfpowerbenchmarkingenergy,sukthanker2024hwgptbenchhardwareawarearchitecturebenchmark}.
Results therefore require the precision, kernel, runtime, memory hierarchy, and device.
\end{reviewchange}

\begin{reviewchange}
Table~\ref{tab:evidence-map} records the evidence status of the major
cross-stage interactions used in the remainder of the survey.

\begin{table}[t]
\centering
\caption{Evidence map for the survey's major cross-stage interactions.  A label
describes the strength of evidence for the complete interaction, not for every
component mechanism.}
\label{tab:evidence-map}
\small
\renewcommand{\arraystretch}{1.12}
\setlength{\tabcolsep}{4pt}
\begin{tabular}{p{0.31\linewidth} p{0.25\linewidth} p{0.36\linewidth}}
\toprule
\textbf{Interaction} & \textbf{Evidence status} & \textbf{Basis and boundary} \\
\midrule
Visual-token reduction $\rightarrow$ task behaviour & Direct evidence & Direct token dropping and merging evaluations; effects remain model- and task-dependent \\
Compression $\rightarrow$ MoE routing & Direct evidence (limited setting)\textsuperscript{a} & FastMMoE~\citep{xia2025fastmmoe} couples routing signals to pruning; Section~\ref{sec:interface_b-empirical} directly measures assignment changes in two MoE-VLMs on ScienceQA; Section~\ref{sec:interface_b-larger-models} adds two larger MoE-VLMs on ScienceQA, MMMU-Pro, and MathVista \\
Quantization $\rightarrow$ MoE routing & Direct evidence & Routing-consistent quantization studies directly compare low-precision and full-precision assignments \\
Routing distribution $\rightarrow$ expert-wise precision allocation & Supported inference & Expert-aware quantization uses calibration traffic; compression-conditioned miscalibration is not directly tested \\
Quantized execution $\rightarrow$ device performance & Direct evidence & Latency, energy, and memory effects are measured for specified kernels and devices; transfer across platforms is not assumed \\
Compression/routing $\rightarrow$ expert-cache traffic & Open hypothesis & Proposed hardware pathway; requires end-to-end measurements on a specified device \\
Multimodal KV eviction $\rightarrow$ deferred-evidence loss & Direct evidence & Multimodal cache studies report deferred importance and modality-specific retention effects \\
TRC $\rightarrow$ video-understanding quality & Open hypothesis & Control sensitivity replicates in two MoE-VLMs, but natural TRC does not predict correctness in either evaluated sample and causal relevance is unvalidated \\
\bottomrule
\end{tabular}
\vspace{2pt}

\parbox{0.94\linewidth}{\footnotesize\textsuperscript{a} The evidence tier is upgraded
only for the DeepSeek-VL2-Tiny and InternVL3.5-30B-A3B ScienceQA settings measured in
Section~\ref{sec:interface_b-empirical} and the Qwen3.5-35B-A3B and Qwen3.5-122B-A10B settings
measured in Section~\ref{sec:interface_b-larger-models}; the model runs use different sample sizes,
precision, and preprocessing, and generalization to other models, modalities, and
compression methods remains an evidence-supported inference.}
\end{table}
\end{reviewchange}

\subsection{Interface B: The Compression-Routing Interaction}
\label{sec:interface_b}

\subsubsection{Mechanism and Prior Evidence}

Dropping shortens the sequence and removes attention context, whereas merging creates
new representations; either operation can change the hidden states presented to a
downstream router.  FastMMoE uses visual-token router-probability similarity to guide
pruning on DeepSeek-VL2 and InternVL3.5~\citep{xia2025fastmmoe}, but does not report
baseline-to-pruned downstream assignment-change rates.  The experiment below measures
that quantity in two MoE-VLMs on ScienceQA (Section~\ref{sec:interface_b-empirical}) and in two larger MoE-VLMs on three benchmarks (Section~\ref{sec:interface_b-larger-models}).  It does not test a routing-stability objective or
establish that preserving assignments improves task quality.

\begin{reviewchange}
\subsubsection{Empirical Validation: Compression-Induced Routing Disruption}
\label{sec:interface_b-empirical}

We evaluate DeepSeek-VL2-Tiny and InternVL3.5-30B-A3B on paired ScienceQA examples (1,000 for DeepSeek-VL2-Tiny and 256 for InternVL3.5-30B-A3B) at 25\%, 50\%, 70\%,
and 75\% visual-token retention, with additional 100\% L2 and random identity controls.
L2 retention keeps the largest-activation tokens; random retention uses one
fixed seed and is therefore a matched within-model control, not a multi-seed estimate.
The unpruned accuracies are $71.60\%$ and $93.36\%$, respectively.  Routing Assignment
Change Rate (RACR) is the fraction of baseline Top-$k$ assignments not preserved for
retained token-layer units; the Top-1 expert-change rate is the percentage of retained
token-layer units whose highest-ranked expert differs from the unpruned run.
This identity-preserving L2/random protocol is not a reproduction of FastMMoE; it is
designed specifically to compare retained tokens before and after pruning.

Within each model, all conditions use the same sampled example, image, question, answer
choices, prompt template, preprocessing output, and weights.  The image--question batch
is constructed once and each condition is derived from that visual-token sequence.
Generation is greedy, and the dataset seed is 20260818.  The reported random condition
uses replicate index 0, with an example-specific deterministic mask.  InternVL's routed expert matrices are
quantized to NF4 with FP16 computation, while router gates, attention, vision,
embeddings, and output heads remain FP16.  DeepSeek is evaluated in FP16.  The model
runs also have different sample sizes and preprocessing, so absolute accuracy and
routing values are not paired cross-model comparisons.  At 100\% retention, both
InternVL controls exactly reproduce the unpruned answers and routing.  The nominal
70\% InternVL setting retains 179 of 256 visual tokens (69.92\%).

Retained-token identity is aligned by original visual-token position.  Each compressed
route file stores the indices of the retained tokens in the unpruned visual sequence.
For every retained token and routed layer, its compressed Top-$k$ expert set is compared
with the unpruned Top-$k$ set at the same original index.  RACR is one minus the fraction
of baseline assignments preserved under this alignment, so RACR itself lies between
zero and one and cannot be negative.

The headline result replicates across the two evaluated architectures: pruning changes
the routing of tokens that remain.  At the common 50\% retention point, L2/random RACR
is $28.49\%/33.46\%$ for DeepSeek and $25.30\%/28.86\%$ for InternVL; the corresponding
Top-1 expert-change rates are $40.79\%/48.22\%$ and $33.28\%/37.69\%$.  Thus L2 retention
preserves routing better than random retention within both models.  The accuracy
ordering is not universal: relative to each model's unpruned accuracy, L2/random
reduce accuracy by $2.37\%/9.08\%$ for DeepSeek and $5.86\%/5.02\%$ for InternVL.
This dissociation
reinforces that lower routing change is not, by itself, evidence that routing
preservation causes better task performance.

For DeepSeek's additional budgets, RACR ranges from $25.6$--$29.1\%$ under L2 retention
and $27.2$--$42.1\%$ under random retention.  L2 accuracy is non-monotonic across the
tested budgets, and its paired confidence intervals include zero except at 75\%
retention; no monotonic accuracy claim is made.  Across InternVL's four pruning
budgets, L2 RACR ranges from $20.61\%$ to $30.80\%$, compared with $22.82\%$ to
$35.63\%$ for random retention.  L2 has lower RACR at every matched budget, while
random retention has higher observed accuracy at 25\%, 50\%, and 70\%, and the
methods tie at 75\%.  Table~\ref{tab:exp1-matched} reports
absolute condition values and percentage changes from each model's own baseline.
The accuracy change is $100(a_r/a_{\mathrm{base}}-1)$, where $a_r$ is accuracy under
retention $r$ and $a_{\mathrm{base}}$ is the same model's unpruned accuracy.  A negative
value indicates a reduction relative to that baseline.

\begin{figure}[t]
\centering
\includegraphics[width=\textwidth]{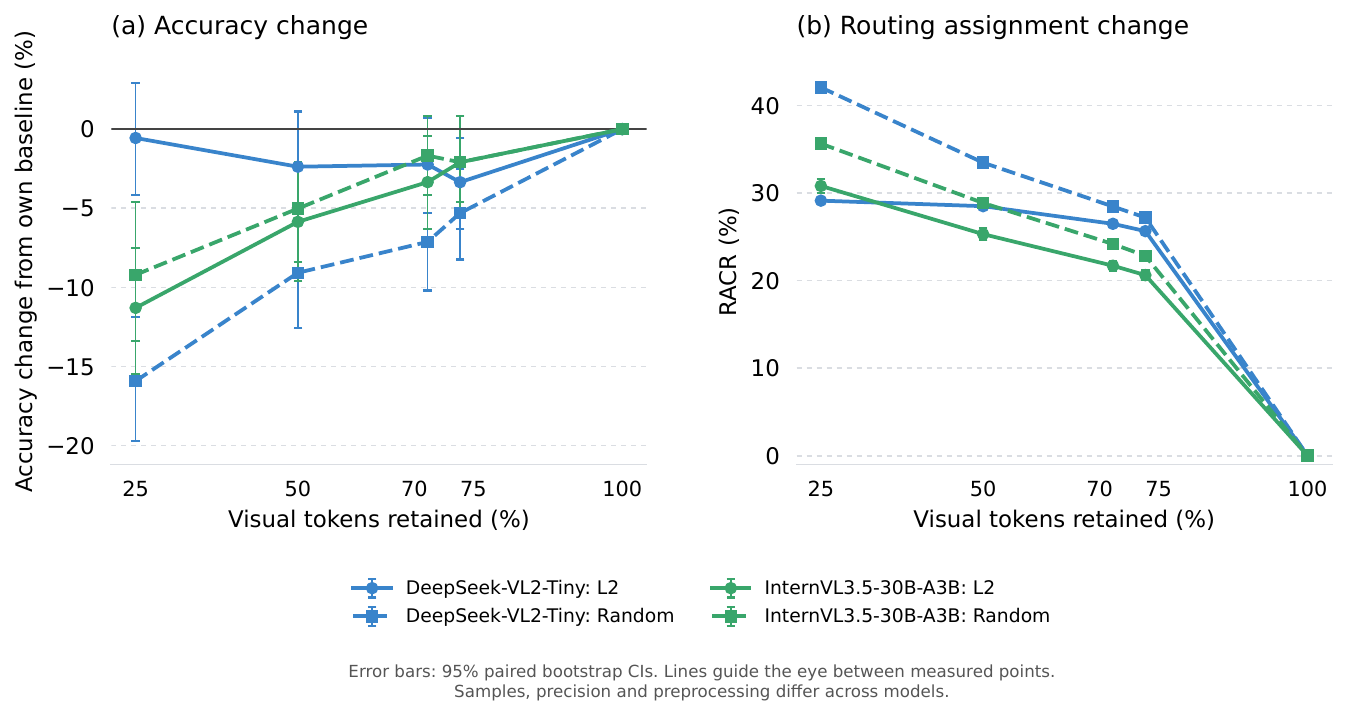}
\caption{ScienceQA accuracy change relative to each model's unpruned baseline.
(a) DeepSeek-VL2-Tiny (FP16) and InternVL3.5-30B-A3B (expert-NF4) at all four measured pruning budgets and the unpruned baseline;
connecting lines guide the eye between measured points.
(b) L2 and random pruning at the common 50\% retention point for DeepSeek and
InternVL3.5-30B-A3B (expert-NF4).  Change is $100(a_r/a_{\mathrm{base}}-1)$.
Different samples, precision, and preprocessing preclude a paired cross-model test;
normalization does not remove
these differences.}
\label{fig:exp1-model-comparison}
\end{figure}

\begin{table}[t]
\centering
\caption{Experiment~1 results by model and measured retention budget.
Accuracy change is a percentage of each model's unpruned accuracy,
$100(a_r/a_{\mathrm{base}}-1)$. RACR and Top-1 change are non-negative
routing-change rates; larger values mean more expert reassignment.
Baseline routing comparisons are not applicable; the measured 100\% InternVL
L2 and random controls have zero routing change.}
\label{tab:exp1-matched}
\small
\setlength{\tabcolsep}{3.5pt}
\resizebox{\textwidth}{!}{%
\begin{tabular}{lclrrrr}
\toprule
Model & Retention & Method & Accuracy (\%) & Change (\%) & RACR (\%) & Top-1 change (\%) \\
\midrule
DeepSeek-VL2-Tiny & 100\% & Baseline & 71.60 & 0.00 & -- & -- \\
 & 25\% & L2 & 71.20 & -0.56 & 29.12 & 41.14 \\
 & 25\% & Random & 60.20 & -15.92 & 42.05 & 58.01 \\
 & 50\% & L2 & 69.90 & -2.37 & 28.49 & 40.79 \\
 & 50\% & Random & 65.10 & -9.08 & 33.46 & 48.22 \\
 & 70\% & L2 & 70.00 & -2.23 & 26.48 & 38.58 \\
 & 70\% & Random & 66.50 & -7.12 & 28.46 & 41.61 \\
 & 75\% & L2 & 69.20 & -3.35 & 25.64 & 37.53 \\
 & 75\% & Random & 67.80 & -5.31 & 27.19 & 39.92 \\
\addlinespace
InternVL3.5-30B-A3B & 100\% & Baseline & 93.36 & 0.00 & -- & -- \\
 & 25\% & L2 & 82.81 & -11.30 & 30.80 & 40.72 \\
 & 25\% & Random & 84.77 & -9.21 & 35.63 & 45.61 \\
 & 50\% & L2 & 87.89 & -5.86 & 25.30 & 33.28 \\
 & 50\% & Random & 88.67 & -5.02 & 28.86 & 37.69 \\
 & 70\% & L2 & 90.23 & -3.35 & 21.69 & 28.59 \\
 & 70\% & Random & 91.80 & -1.67 & 24.17 & 31.90 \\
 & 75\% & L2 & 91.41 & -2.09 & 20.61 & 27.15 \\
 & 75\% & Random & 91.41 & -2.09 & 22.82 & 30.19 \\
 & 100\% & L2 & 93.36 & 0.00 & 0.00 & 0.00 \\
 & 100\% & Random & 93.36 & 0.00 & 0.00 & 0.00 \\
\addlinespace
\bottomrule
\end{tabular}%
}
\end{table}
\end{reviewchange}


\FloatBarrier
\subsubsection{Extension to Larger Models and Harder Benchmarks}
\label{sec:interface_b-larger-models}

For verifying if the compression-induced reassignment still holds in a larger scale and under more challenging tasks, we repeat the retained-token measurement on Qwen3.5-35B-A3B and Qwen3.5-122B-A10B, two MoE-VLMs with 256 experts for each routed layer and Top-8 routing (40 and 48 routed layers, respectively). We evaluate each model on three benchmarks:
1,000 ScienceQA test examples with images from dataset seed 20260818, the full MMMU-Pro standard (10-choice) test split (1,730 questions, up to seven images each), and the full MathVista testmini split (1,000 questions). The same questions from the same pinned dataset revisions are posed to both models.

Each question is run under 15 conditions: the unpruned baseline, L2 and random
identity controls at 100\% retention, L2 retention at 75\%, 50\%, and 25\%, and random
retention at each budget with three seeds. Pruning is applied after the vision
projector and before the language decoder. Budgets are applied per image, L2 ranks
post-projector activation norms, and retained tokens keep their original positional
coordinates. Routed expert matrices are quantized to NF4 with BF16 computation,
while routers, shared experts, attention, vision, embeddings, and the output head
remain BF16; unpruned and pruned conditions share the same loaded weights. Routing
is recorded during prompt prefill only. Decoding is greedy with direct-answer
prompts, answers are parsed strictly without an LLM judge, and unparseable answers
are scored as incorrect (unpruned valid-answer rates are 94.2--97.7\% for the 35B
model and 97.4--99.9\% for the 122B model). Confidence intervals use 2,000 paired
bootstrap resamples of questions. Because budgets, seed count, per-image selection,
and position handling differ from Section~\ref{sec:interface_b-empirical}, the values
in Table~\ref{tab:exp1-qwen-larger} are not paired comparisons with those in
Table~\ref{tab:exp1-matched}.

The effect persists in every setting (Table~\ref{tab:exp1-qwen-larger}). Under L2
retention, keeping 75\% of the visual tokens already changes 19.3--21.5\% of baseline
expert assignments on retained tokens in the 35B model and 15.4--16.3\% in the 122B
model, and RACR rises monotonically as the budget shrinks in all twelve
model--benchmark--method sequences. L2 retention yields lower RACR than random
retention in all 18 matched comparisons: the paired L2$-$random RACR differences
range from $-3.8$ to $-15.4$ percentage points, and every 95\% interval excludes zero.

Accuracy does not follow routing uniformly. On ScienceQA and MathVista, L2
retention preserves more accuracy than random retention at every budget in both
models (paired differences of $+1.1$ to $+7.7$ percentage points; all 95\% intervals
exclude zero). On MMMU-Pro, the interval for the L2$-$random accuracy difference
includes zero in five of the six comparisons, and at 25\% retention random retention
has the higher observed accuracy in both models (differences of $-1.19$ and $-1.04$
percentage points) even though L2 RACR is 10.9 and 13.3 points lower. Lower routing
change is therefore again not sufficient for better task accuracy, consistent with
the InternVL result in Section~\ref{sec:interface_b-empirical}.

RACR is lower for the 122B model than for the 35B model in every cell, but the two
checkpoints differ in depth, width, and training, so this is not evidence of a
parameter-scaling effect. These results extend the measured scope of
compression-induced reassignment to a 122B-class MoE-VLM and to multi-image and
mathematical-reasoning benchmarks. They remain limited to L2 and random token
dropping under expert-NF4 precision, and they do not show that routing change causes
the accuracy loss.

\begin{table}[t]
\centering
\caption{Extension of Experiment~1 to Qwen3.5-35B-A3B and Qwen3.5-122B-A10B
(expert-NF4) on three benchmarks. Paired cells give L2 / random retention(random
values are means over three seeds, averaged within each question). Change is a
percentage of each model's unpruned accuracy on that benchmark,
$100(a_r/a_{\mathrm{base}}-1)$, as in Table~\ref{tab:exp1-matched}. Unpruned is the
unpruned accuracy (\%) and Kept is the retention budget. RACR and Top-1 expert change
are routing-change rates on retained visual tokens. Both 100\% identity controls
have zero routing change and reproduce the baseline answers in all six runs.}
\label{tab:exp1-qwen-larger}
\small
\setlength{\tabcolsep}{3.5pt}
\renewcommand{\arraystretch}{1.12}
\resizebox{\textwidth}{!}{%
\begin{tabular}{lllcrrrr}
\toprule
Model & Benchmark & Unpruned & Kept & Accuracy (\%) & Change (\%) & RACR (\%) & Top-1 (\%) \\
\midrule
Qwen3.5-35B-A3B & ScienceQA ($n=1{,}000$) & 87.80 & 75\% & 88.60 / 87.23 & +0.91 / -0.65 & 19.33 / 24.09 & 21.83 / 27.78 \\
 & & & 50\% & 87.00 / 84.80 & -0.91 / -3.42 & 23.62 / 33.19 & 27.02 / 38.28 \\
 & & & 25\% & 84.10 / 80.60 & -4.21 / -8.20 & 30.19 / 41.95 & 34.79 / 48.11 \\
\cmidrule(l){2-8}
 & MMMU-Pro ($n=1{,}730$) & 44.86 & 75\% & 43.99 / 43.80 & -1.93 / -2.36 & 21.51 / 25.31 & 24.50 / 29.16 \\
 & & & 50\% & 41.79 / 41.71 & -6.83 / -7.00 & 27.02 / 34.76 & 31.28 / 40.20 \\
 & & & 25\% & 38.32 / 39.52 & -14.56 / -11.90 & 32.42 / 43.30 & 37.91 / 49.92 \\
\cmidrule(l){2-8}
 & MathVista ($n=1{,}000$) & 71.40 & 75\% & 70.30 / 67.43 & -1.54 / -5.56 & 20.56 / 26.10 & 23.94 / 30.75 \\
 & & & 50\% & 67.20 / 61.00 & -5.88 / -14.57 & 25.03 / 36.06 & 29.27 / 42.24 \\
 & & & 25\% & 57.30 / 50.93 & -19.75 / -28.66 & 30.94 / 45.40 & 36.40 / 52.70 \\
\midrule
Qwen3.5-122B-A10B & ScienceQA ($n=1{,}000$) & 94.60 & 75\% & 94.40 / 93.30 & -0.21 / -1.37 & 15.69 / 19.89 & 19.41 / 24.95 \\
 & & & 50\% & 93.30 / 90.73 & -1.37 / -4.09 & 19.14 / 28.85 & 23.73 / 35.86 \\
 & & & 25\% & 88.60 / 86.03 & -6.34 / -9.06 & 23.35 / 37.38 & 28.99 / 45.71 \\
\cmidrule(l){2-8}
 & MMMU-Pro ($n=1{,}730$) & 51.39 & 75\% & 50.92 / 50.33 & -0.90 / -2.06 & 16.25 / 20.14 & 20.27 / 25.40 \\
 & & & 50\% & 49.36 / 48.09 & -3.94 / -6.41 & 20.87 / 29.44 & 26.13 / 36.79 \\
 & & & 25\% & 43.87 / 44.91 & -14.62 / -12.60 & 24.67 / 37.95 & 31.01 / 46.73 \\
\cmidrule(l){2-8}
 & MathVista ($n=1{,}000$) & 73.70 & 75\% & 73.40 / 70.33 & -0.41 / -4.57 & 15.44 / 20.37 & 19.12 / 25.52 \\
 & & & 50\% & 71.00 / 63.67 & -3.66 / -13.61 & 19.81 / 29.88 & 24.71 / 37.09 \\
 & & & 25\% & 61.90 / 54.20 & -16.01 / -26.46 & 23.75 / 39.15 & 29.73 / 47.83 \\
\bottomrule
\end{tabular}%
}
\end{table}
\FloatBarrier

\subsection{Interface C: Routing Instability and Quantization
            Sensitivity}
\label{sec:interface_c}
Interface~C contains two claims with different evidence levels.
{Under the evaluated low-bit settings, quantization can alter MoE routing
by perturbing router logits
\citep{fu2026eaquantenhancingposttrainingquantization,park2026valueandstructurealignmentroutingconsistentquantization}.}
{Routing traffic can also inform which experts receive higher precision,
but the effect of a deployment-time routing shift on that allocation remains largely
unexplored.}

\subsubsection{How Routing Decisions Shape Quantization Sensitivity}

Several recent quantization methods allocate precision according to expert importance,
assigning higher bit-widths to experts that receive more routing traffic during
calibration \citep{deng2026gemqglobalexpertlevelmixedprecision}.  Applying that
allocation unchanged at deployment implicitly assumes that the calibration traffic is
representative.

{It is therefore reasonable to infer that, if the high-traffic experts
at calibration differ from those at deployment, an expert-wise precision allocation
based on calibration frequency may no longer match deployment usage.}  Upstream
compression is one possible source of such mismatch, but this pathway has not been
tested directly.

Second, the experts most affected by this miscalibration are
low-frequency experts---those activated only by rare or highly
specific input patterns.
\citet{fu2026eaquantenhancingposttrainingquantization} report that the assignments that change under W4A8 quantization
disproportionately involve low-frequency expert-token
combinations with high information content.
A compression policy that also changes these activations could therefore create a
compounding failure, but this extension is an evidence-supported inference rather than
a result of the cited experiment.

\subsubsection{Potential Compounding Across Interfaces}

{We therefore hypothesise that compression-induced and
quantization-induced routing changes may compound when both perturb the same
near-boundary assignments.}  No general inequality or amplification factor is justified
without measuring the joint distribution of assignment changes.  A suitable experiment
would report routing change rates for compression alone, quantization alone, and their
composition on the same model and inputs, together with task accuracy and confidence
intervals.  Until such evidence is available, ``compounding'' denotes a proposed
mechanism rather than an established quantitative relationship.

\subsection{Composed Optimization and the Pareto Frontier}
\label{sec:pareto}
Compression, routing, and quantization are commonly studied as separate
accuracy-efficiency trade-offs.  Compression changes token count and may lose
information; routing trades sparse activation against balance and systems overhead;
and quantization trades numerical precision against storage and execution cost.

{Because quantization-induced routing changes are measured and compression
changes the inputs presented downstream, independently chosen operating points need not
remain Pareto-optimal after composition.} We therefore treat
joint measurement of the full-system Pareto frontier as a co-design hypothesis: the
full-system frontier should be measured jointly rather than assumed to equal the
combination of component-level frontiers.  Joint optimisation is one response, but sequential
calibration, robustness constraints, or post-composition retuning may also be viable.

\subsection{Hardware Constraints as Additional Pareto Pressure}

{On the memory-bound workloads and devices studied by
\citet{chung2026joulesgodiagnosinginference}, data movement accounts for a major share
of inference energy.}  This is not a universal property of every model or device; the
balance depends on kernels, batch size, utilisation, and the memory hierarchy.
This transfer volume depends jointly on:
quantization precision (lower precision reduces bytes per
weight),
expert activation patterns (which experts are loaded determines
transfer volume),
and token count (which determines how many routing decisions
and expert activations occur per forward pass).
These quantities are influenced by the pipeline parameters
$(\boldsymbol{\phi}, \boldsymbol{\theta}, \boldsymbol{\pi})$, but their net cost must be
measured on the target hardware.

{A remaining hypothesis is that a compression policy which reduces token
count but disrupts routing locality could increase net transfer volume despite using
fewer tokens.}  Such a shift could force loading of a more diverse set of expert
weights per forward pass.
Similarly, a quantization policy that reduces per-weight
transfer cost but destabilises routing could increase transfer
volume by forcing prefetch misses on experts that routing
predictions failed to anticipate.
These are testable hardware pathways, not established outcomes; validating them
requires end-to-end traces of expert loads, cache hits, latency, and energy on the same
deployment.

\subsection{Framework Scope and Section Map}
\label{sec:chain_summary}

Table~\ref{tab:chain_map} connects later sections to the relevant stages and evidence
boundaries at Interfaces A--D.  The KV cache remains persistent state outside the four transformations:
it stores outputs of earlier decisions and reintroduces them as decoding-time memory,
retention, and hardware constraints.

\begin{table}[t]
  \centering
  \caption{%
    Map from survey sections to framework stages and interfaces.  An entry indicates
    relevance to an interface, not that the complete interaction has been measured.
  }
  \label{tab:chain_map}
  \small
  \renewcommand{\arraystretch}{1.15}
  \setlength{\tabcolsep}{3pt}
  \begin{tabular}{p{0.16\linewidth} p{0.22\linewidth} p{0.27\linewidth} p{0.25\linewidth}}
    \toprule
    \textbf{Section} &
    \textbf{Chain stage} &
    \textbf{Dominant bottleneck} &
    \textbf{Interaction informed} \\
    \midrule
    \S\ref{sec:redundancy} &
      Pre-compression evidence &
      Uneven information density &
      Interface A \\
    \S\ref{sec:vit}   &
      Stage 1: Compression &
      ViT spatial redundancy &
      Interface A, Interface B \\
    \S\ref{sec:mlhm}  &
      Stage 1: Compression &
      MLLM pipeline sensitivity &
      Interface A, Interface B \\
    \S\ref{sec:video} &
      Stage 1 + TRC &
      Temporal redundancy, expert jitter &
      Interface B + TRC \\
    \S\ref{sec:kvcache} &
      Cache-to-reasoning boundary &
      KV memory growth, evidence retention &
      Cache pressure + hardware feedback \\
    \S\ref{sec:moe} &
      Stage 2: Routing &
      Load imbalance, collapse &
      Interface B, Interface C \\
    \S\ref{sec:serving} &
      Routing-to-systems boundary &
      Expert movement, cache interference &
      Interfaces C and D \\
    \S\ref{sec:quant}  &
      Stage 3: Quantization &
      Activation outliers, routing drift &
      Interfaces C and D \\
    \S\ref{sec:edge}   &
      Stage 4: Hardware &
      SRAM, bandwidth, energy &
      Interface D \\
    \S\ref{sec:bench}  &
      Evaluation &
      Benchmark gaps &
      Interfaces A--D \\
    \bottomrule
  \end{tabular}
\end{table}

\section{Token Redundancy in Multimodal Transformers}
\label{sec:redundancy}

Token redundancy is the starting point for almost every efficiency method in
multimodal transformers.  The topic asks which visual, temporal, or cross-modal tokens
carry information that is predictable from context and can therefore be removed,
merged, or down-weighted.  Its importance for edge deployment is direct: redundant
visual tokens expand every downstream cost, including encoder attention, LLM prefilling,
KV-cache memory, and MoE routing traffic.  The key caveat is that redundancy is not
irrelevance: even low-density patches, OCR spans, or frames can become necessary for
grounding or reasoning.  This section therefore treats redundancy as input- and
task-dependent, not as permission to remove tokens indiscriminately.

\begin{reviewchange}
\subsection{Self-Supervised Evidence}

Reconstruction-based pre-training provides direct evidence that visual
patches are predictable from context: MAE reports competitive representation learning
with high masking ratios, and VideoMAE reports effective pre-training near $90\%$
masking~\citep{he2021maskedautoencodersscalablevision,tong2022videomaemaskedautoencodersdataefficient}.
These results establish redundancy for the cited pre-training objectives; they do
\emph{not} establish that the same fraction of tokens can be removed without
information loss during downstream multimodal inference.
Instead of detecting redundancy post hoc, latent-array models avoid full-resolution
token sequences by using iterative cross-attention to condense the input
~\citep{jaegle2021perceivergeneralperceptioniterative}.

\end{reviewchange}

\subsection{Cross-Modal Redundancy}

Video adds cross-modal redundancy because many visual tokens do not align with text;
they encode background, irrelevant transitions, or repeated information.  Redundancy-aware
coarse pre-training reduces alignment
effort for non-relevant visual tokens and enhances retrieval by reducing noisy signals
generated by non-aligning video tokens \citep{wu2022rapredundancyawarevideolanguagepretraining}.
Answering questions demands fine alignment, since many frames do not contain information
relevant to the specific question posed even if those frames have saliency independent
of the task. Question-guided frame selection helps focus computation on relevant frames
that influence the answer \citep{Li_2023}.
This progression from sentence-to-video alignment to query-to-token alignment corresponds
to the move from global token constraints in ViT to query conditioning in MLLMs.

\subsection{The Information Density Framework}

Recently, new theories have brought together the three forms of redundancy
mentioned above under the heading of \emph{token information density}, which is defined
as the information content of a token when conditioned on its context~\citep{shao2026surveytokencompressionefficient}.
When a token's information density drops below a threshold, it may be
eligible for elimination; however, both the threshold and the density itself depend on
the modality (visuospatial tokens have lower average densities compared to text),
and the context (background tokens tend to have lower density than object boundaries).
The decision to apply these theories has become a choice between eliminating low
density tokens or keeping them but with diminished weight; attention-based learning
is a way to implement deletion, while position rescaling achieves the opposite
conservatively~\citep{mahajan2025attentionguidedalignmentefficient}.
\section{Token Compression in Vision Transformers}
\label{sec:vit}

Token compression in Vision Transformers converts redundancy into raw efficiency gains
within the visual encoder.  The rationale behind this is that ViT self-attention depends
on the number of image patches, and compressing visual sequences before they enter the LLM
might lead to more efficient encoders as well as shorter multimodal context lengths.  Yet
compression on edge devices usually needs to be extreme, leading to unique sources of error,
such as information loss due to dropping tokens, conflicts due to merging tokens, and overly
harsh compression due to fixed compression limits.  This is why approaches for token
compression discussed in this section revolve around one key question: How does one compress
tokens while not destroying the feature representation that other layers rely on?

\subsection{From Token Dropping to Token Merging: A Design Dichotomy}

\begin{figure}[h]
\centering
\includegraphics[width=0.9\textwidth]{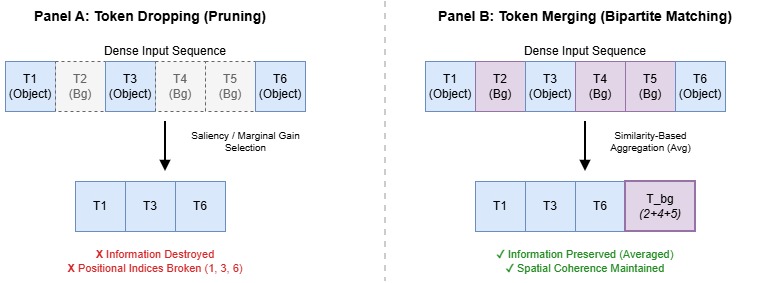}
\begin{reviewchange}
\caption{Illustration of two visual-token reduction mechanisms. \textbf{(A)} Token
dropping permanently removes selected representations and may require careful position
handling. \textbf{(B)} Token merging uses bipartite matching to replace selected token
pairs with aggregate representations.  The comparison describes information flow and
does not assert that one mechanism is universally superior.}
\label{fig:token-compression}
\end{reviewchange}
\end{figure}

The basic conflict in ViT token compression is between irreversibility, efficiency, and
representation fidelity.  Token pruning removes selected elements, while token merging
replaces several representations with an aggregate; neither operation guarantees
preservation of all downstream-relevant information.  Training-free bipartite soft
matching provides one practical balance,
where tokens are split into two subsets and similar token pairs from each subset are
merged, leading to a speedup of up to $2\!\times$ while suffering an accuracy loss
of only $0.2\%$--$0.3\%$ on ImageNet for the evaluated DeiT, MAE, and CLIP-like
models~\citep{bolya2023tokenmergingvitfaster}.  Merging retains an aggregate of source
tokens, whereas dropping removes selected representations; the relative advantage is
task- and method-dependent.  Figure~\ref{fig:token-compression} illustrates the two
mechanisms rather than establishing a universal ranking.

\subsection{Layer-Wise and Spatial Failure Modes in Token Merging}

A fixed global similarity threshold creates two orthogonal failure modes.  First,
thresholds that are safe in late semantic layers can over-merge early layers where
fine-grained texture differences still carry local information.  Layer-specific
adaptive thresholds address this depth-wise mismatch and report more than $30\%$ DeiT
FLOP reduction without a measured accuracy decrease in the cited setting
~\citep{Lee_2025}.  Second,
feature similarity alone can merge visually distant regions that happen to share
embedding statistics, such as sky and sea patches with similar colour distributions.
Adding spatial distance to the merge criterion protects layout information and is
especially important for tasks such as scene understanding and visual question
answering~\citep{huang2025tosatokenmergingspatial}.  These results suggest evaluating
both depth-wise calibration and spatial coherence rather than treating either property
as sufficient in every setting.

\subsection{Extending Compression Beyond Accuracy: Generativity and Throughput}

Once basic merging is effective, the next trade-off is whether to improve the
accuracy-throughput frontier by protecting salient tokens, changing the algebra of
merging, or sparsifying attention without changing the token count.  Salience-weighted
similarity scoring protects foreground and edge tokens while adaptive per-layer ratios
control how aggressively each depth is compressed, reducing high-compression Top-1
degradation to about half that of fixed matching baselines~\citep{lee2026adamergesalienceawareadaptivetoken}.  A
matrix-decomposition formulation shifts the bottleneck from pairwise matching to
GPU-friendly batched operations and can pair token merging with restoration, reducing
Stable Diffusion generation latency by $31\%$ while preserving visual quality
~\citep{huo2026mamemarematrixbased}.  Attention sparsification follows a complementary axis: rather than
reducing token count, it reshapes attention-score statistics and retains only
significant token-token connections, lowering attention density over the existing
sequence~\citep{chen2025rettentionultrasparsevisual}.

\section{Token Management in Multimodal LLMs}
\label{sec:mlhm}

Token management for multimodal LLMs is concerned with choosing which visual tokens
to preserve, in what order, and after how many have been fed into the language model
pipeline. This subject is important since the LLM considers visual and textual tokens
as one single context, meaning that all saved visual tokens will take up attention
compute and KV-cache memory.  The intervention stage therefore matters: early
compression offers more potential compute savings but may remove visual detail, while
late compression retains more upstream computation; position handling can also affect
spatial tasks.

\subsection{The Pipeline Sensitivity Problem: Where to Compress?}

{MLLM studies show that the effect of a fixed compression level depends
on where token pruning is applied in the processing pipeline.}  Encoder-stage dropping targets visual
compute, while later dropping changes the context available to the language model.
Multistage strategies exploit this asymmetry by using different budgets across stages
~\citep{liu2024multistagevisiontokendropping}.  High-resolution tiling adds pressure
because a tiled scene can contain thousands of visual tokens.  HiRED uses early
attention to estimate saliency before later ViT computation
~\citep{arif2024hiredattentionguidedtokendropping}.

\subsection{The Compression Approach Spectrum: Greedy, Structural, and Learned}

Given a fixed token budget, three broad approaches are greedy selection, structural or
signal-theoretic projection, and learned selection.  Formal preservation claims for a
projection hold only under the method's sparsity assumptions; they do not imply that no
task-relevant information is ever lost.  Greedy selection maximises an estimated
marginal gain and reports favourable accuracy under aggressive pruning
~\citep{pei2025greedypruneretentingcriticalvisual}; query-conditioned
sparsification creates a budget that is dependent on the ongoing query of the text 
by employing cross-modal attention for determining relevant visual tokens~\citep{zhang2025sparsevlmvisualtokensparsification}; while
compressive-sensing projections offer formal preservation guarantees under the cited
method's sparse-signal assumptions~\citep{kiruluta2025csvlmcompressedsensingattention}.
Finally, learned compression is another dimension that occupies the spectrum:
evolutionary pseudo-labeling trains a compressor whose decisions follow task performance
without creating positional gaps through token dropping~\citep{song2026evocomplearningvisualtoken}.
The resulting trade-off is clear:
greedy, query-conditioned, and structural methods are largely training-free but less
adaptive, while learned compressors adapt to task statistics at the cost of an
additional training procedure.

\subsection{Positional Integrity Under Compression}
Compression can also change the relationship between retained content and positional
indices, with measured consequences for spatial tasks in the cited work.
This can be addressed by encoding spatiotemporal information in the compressed tokens
rather than renumbering surviving tokens for better performance in token dropping
regimes \citep{huang2026ppepositionalpreservationembedding}.  In general,
a training free approach that combines positionality with the use of viewpoint tiles and
merging tokens based on text input becomes especially relevant during intensive
compression and spatial queries.
~\citep{yu2026visiontrimunifiedvisiontoken}.  A complementary alternative leaves indices intact but changes
positional encoding magnitudes, amplifying informative tokens so they exert greater
influence on attention scores~\citep{huang2026modixtrainingfreemultimodalinformationdriven}.
These methods support reporting positional handling explicitly in MLLM compression.

\section{Video Understanding and Temporal Routing Consistency}
\label{sec:video}

Video understanding extends token redundancy from space into time.  The topic is
important because video inputs can multiply image-token costs by hundreds or thousands
of frames, often making full-token processing impractical for constrained devices.  Yet
temporal data also changes the failure landscape: repeated background can be
compressible, while motion, event boundaries, and temporally separated evidence may be
lost if compression is applied frame by frame without respecting sequence structure.
When MoE routing is added, adjacent frames may also activate different expert subsets.
This section reviews the established video-compression evidence and then introduces TRC
as a proposed diagnostic for describing that routing variation.

\subsection{Why Video Compression Requires More Than Static Token Reduction}

{Video compression studies directly exploit temporal repetition and
variation because independently applying an image policy to every frame ignores
cross-frame structure~\citep{zhang2025dyntokdynamiccompressionvisual,chen2026variationawarevisiontokendropping}.}
A region that is unchanged across neighbouring frames may be a good compression
candidate, whereas motion or an event boundary may be task relevant.  Which regions can
be removed safely remains dependent on the query, scene, and evaluation task.

\subsection{The Temporal Sequencing Challenge: MoE Parallel Routing vs.\ Recurrent
Constraints}

{The architectural evidence suggests that token-level MoE routers
generally select experts from a token's current representation without explicitly
conditioning on the expert selected for a neighbouring frame.}  By contrast, an LSTM state at time $t+1$ is explicitly a
function of the prior recurrent state~\citep{10.1162/neco.1997.9.8.1735}.  This
architectural contrast suggests a useful measurement question: how much do expert
assignments change between adjacent frames?  It does not establish that changing
experts harms temporal reasoning; Transformer residual streams and attention can carry
temporal information even when routed expert sets differ.

Published video methods provide adjacent but not direct evidence.  Motion and content
can be represented separately~\citep{zhang2025vqtokenneuraldiscretetoken}; adaptive
compression assigns larger budgets to complex action segments
~\citep{zhang2025dyntokdynamiccompressionvisual}; and variation-aware dropping removes
features that change little across neighbouring frames
~\citep{chen2026variationawarevisiontokendropping}.  None of these results establishes
that a video MoE should preserve the same experts across frames.  Likewise,
quantization-induced routing changes are established in MoE studies
~\citep{fu2026eaquantenhancingposttrainingquantization}, but compression-induced
temporal routing changes have not been directly measured.

The image-domain measurement in Section~\ref{sec:interface_b-empirical} does not answer
this temporal question: whether video compression perturbs routing across frames, and
whether that perturbation changes TRC, remains untested.

State-space video backbones provide a compact recurrent state, whereas attention-based
MoE models may carry temporal information through their shared representations.  Which
architecture yields more stable expert assignments, and whether such stability is
useful, remain empirical questions.

\begin{reviewchange}
\subsection{Temporal Routing Consistency}
\label{sec:trc}

Existing evaluations of efficient multimodal models primarily measure final-task
accuracy and computational cost.  These metrics do not reveal whether changes to
sequential visual inputs alter the internal routing behaviour of MoE models.  We
therefore propose Temporal Routing Consistency (TRC) as an intermediate diagnostic
that quantifies the consistency of expert assignments across adjacent visual inputs.
Its purpose is not to serve as a proxy for task accuracy, but to make routing behaviour
observable under controlled temporal and visual perturbations.  TRC consequently
answers one narrow question: \emph{do adjacent frames send their visual-token traffic
to the same experts?}

At MoE layer $\ell$, aggregate the routing weights of all visual tokens in frame $t$
into a normalized expert-mass vector $\mathbf{m}_t^{(\ell)}$.  Let
$E_{t,M}^{(\ell)}$ contain the $M$ experts with the largest mass.  TRC@$M$ is the mean
Jaccard overlap between the top-$M$ sets of adjacent frames:
\begin{equation}
  \mathrm{TRC@M}^{(\ell)}
  = \frac{1}{T-1}\sum_{t=1}^{T-1}
  \frac{\left|E_{t,M}^{(\ell)} \cap E_{t+1,M}^{(\ell)}\right|}
       {\left|E_{t,M}^{(\ell)} \cup E_{t+1,M}^{(\ell)}\right|}.
  \label{eq:trc}
\end{equation}
Zero means no top experts are shared; one means the sets are identical.  We use
$M=8$ as the primary setting.  Weighted TRC replaces the sets with the complete mass
vectors and averages $\sum_e\min(m_{t,e},m_{t+1,e})/\sum_e\max(m_{t,e},m_{t+1,e})$,
so the result is not tied to a hard top-$M$ cutoff.  Always-active shared experts are
excluded.  Both metrics must be reported by layer or with a prespecified aggregation.
High TRC means stable expert use; it does not by itself mean that the model understood
the video or that the routing was desirable.

\begin{reviewchange}
\subsection{Empirical Validation: TRC Under Sequence Controls}
\label{sec:trc-empirical}

We evaluate DeepSeek-VL2-Tiny and InternVL3.5-30B-A3B on 256 NExT-QA questions each,
presenting eight uniformly sampled frames as a multi-image sequence.  This is
sampled-frame inference, not native-video inference.  InternVL uses the same pinned
expert-NF4 checkpoint as Experiment~1 and receives separately encoded images with frame
labels; DeepSeek uses its FP16 multi-image preprocessing.  The sample manifests are not
available for a paired cross-model analysis, and the models differ in preprocessing,
expert count, precision, and routing space.  We therefore test whether the control
ordering replicates within each model rather than comparing their absolute TRC levels.

For each frame, we rank experts by the total routing mass they receive across that
frame's visual tokens and keep the top 8.  TRC@8 is then the mean Jaccard overlap of
this top-8 set between adjacent frames: 1 means the same eight dominant experts are
chosen for both frames, 0 means none are shared.  We use 8 as the primary cutoff because
it is a natural middle ground between a stricter Top-1 view and a looser view that risks
approaching the full expert set; TRC@4 and TRC@16 are robustness checks.

The five conditions isolate different sources of routing change.  Duplicate frames
repeat one frame eight times and should maximize consistency, since the visual input
does not change.  Chronological and shuffled sequences use the same eight frames but
differ only in order, isolating the effect of temporal order alone.  Cross-clip
sequences replace some frames with unrelated content, removing visual continuity.
Router permutation randomly relabels which expert is which, independently per frame,
giving a floor for overlap that reflects only chance.

Figure~\ref{fig:trc-conditions} gives the main result.  For both models, TRC@8 and
weighted TRC (which additionally accounts for how much routing mass each expert
receives, not just whether it is in the top 8) follow the same ordering: duplicate
$>$ chronological $>$ shuffled $>$ cross-clip $>$ router permutation.  For DeepSeek,
chronological TRC@8 is $0.6367$; shuffling lowers it to $0.6148$, a paired difference
of $0.0219$ (95\% CI $[0.0181,0.0258]$), while the chronological--cross-clip gap is
$0.1934$ ($[0.1862,0.2002]$).  For InternVL, the corresponding values are $0.5089$,
$0.4760$, and $0.2800$: the paired chronological--shuffled difference is $0.0329$
($[0.0281,0.0379]$), and the chronological--cross-clip difference is $0.2290$
($[0.2183,0.2393]$).  Thus both architectures detect temporal order and show a larger
response when adjacent frames no longer belong to the same clip.

\begin{figure}[t]
\centering
\includegraphics[width=\textwidth]{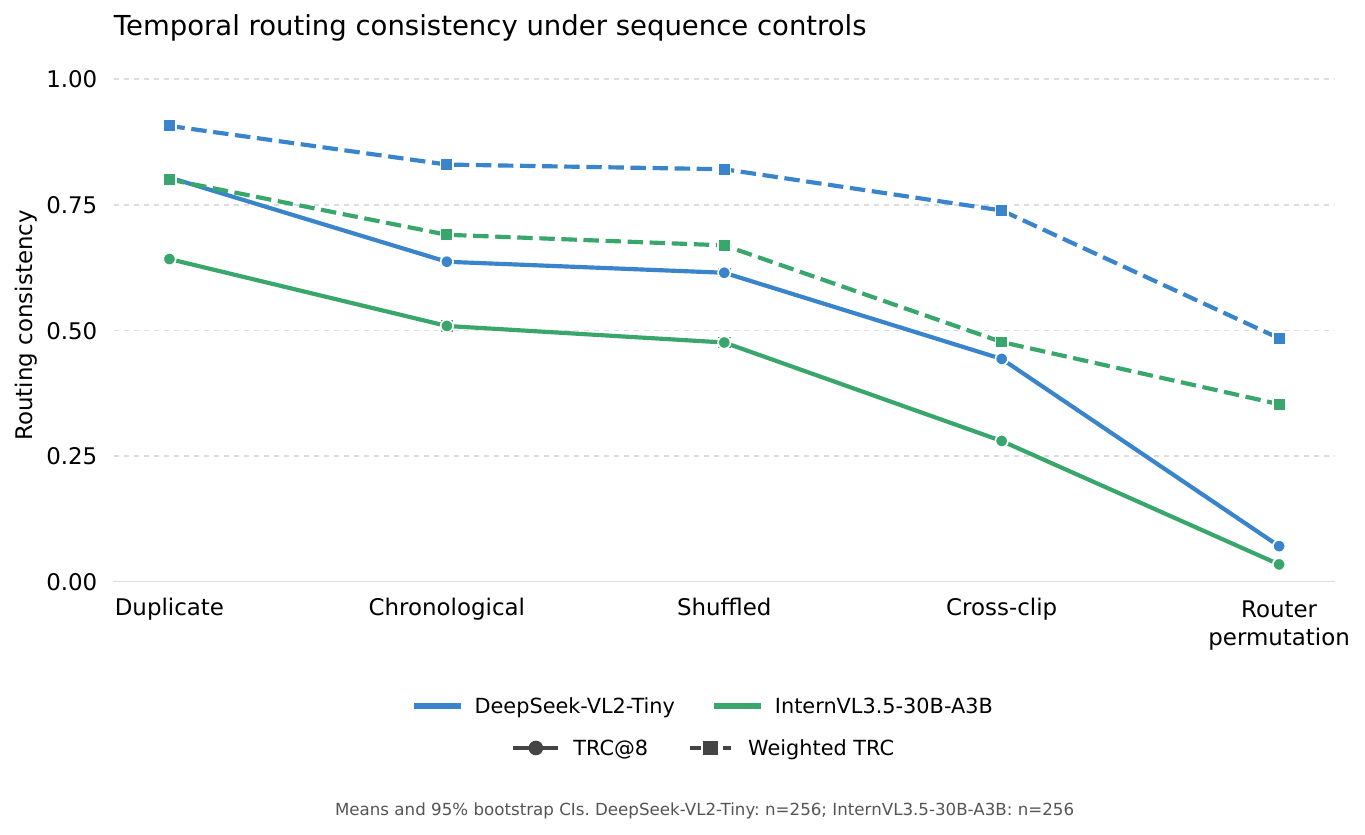}
\caption{TRC control results for DeepSeek-VL2-Tiny and InternVL3.5-30B-A3B on 256
NExT-QA questions per model.  Points are means and bars are 95\% clip-bootstrap
confidence intervals.  TRC@8 (solid lines) and weighted TRC (dashed lines) reproduce
the same within-model control ordering.  Absolute levels are not treated as directly
comparable because the models have different routing spaces, preprocessing, and
precision.}
\label{fig:trc-conditions}
\end{figure}

\begin{table}[t]
\centering
\caption{Experiment~2 routing-consistency means and NExT-QA accuracy under the
five sequence controls ($n=256$ per model).  \textit{n/m}: answer correctness was
not measured for that condition in the saved run.  \textit{n/a}: expert-label
permutation is a routing-only diagnostic, without a separate answer-generation
condition.  Measured accuracies use each model's own evaluation sample.}
\label{tab:trc-models}
\small
\setlength{\tabcolsep}{4pt}
\resizebox{\textwidth}{!}{%
\begin{tabular}{llrrr}
\toprule
Model & Condition & TRC@8 & Weighted TRC & Accuracy (\%) \\
\midrule
DeepSeek-VL2-Tiny & Duplicate & 0.8045 & 0.9072 & 57.81 \\
 & Chronological & 0.6367 & 0.8302 & 60.93 \\
 & Shuffled & 0.6148 & 0.8209 & 59.77 \\
 & Cross-clip & 0.4433 & 0.7389 & 36.33 \\
 & Router permutation & 0.0706 & 0.4844 & - \\
\addlinespace
InternVL3.5-30B-A3B & Duplicate & 0.6422 & 0.8004 & 74.61 \\
 & Chronological & 0.5089 & 0.6905 & 87.89 \\
 & Shuffled & 0.4760 & 0.6694 & 86.72 \\
 & Cross-clip & 0.2800 & 0.4770 & 44.92 \\
 & Router permutation & 0.0343 & 0.3532 & - \\
\addlinespace
\bottomrule
\end{tabular}%
}
\end{table}

For DeepSeek, the ordering holds at all 11 routed layers, so the effect is not an
artifact of one layer, although its size varies by depth: chronological TRC@8 ranges
from $0.5580$ at layer 0 to $0.7689$ at layer 5.  InternVL's layer-aggregated TRC@4,
TRC@8, TRC@16, and weighted TRC all reproduce the expected paired-control directions.

Natural chronological TRC does not separate correct from incorrect answers in either
model.  For DeepSeek, TRC@8 averages $0.6366$ for correct answers and $0.6370$ for
incorrect answers, a difference of $-0.0005$ (95\% CI $[-0.0142,+0.0136]$;
point-biserial $r=-0.004$).  For InternVL, the corresponding means are $0.5086$ and
$0.5115$, a difference of $-0.0029$ ($[-0.0327,+0.0286]$; $r=-0.012$).  InternVL's
weighted-TRC difference is likewise indistinguishable from zero ($+0.0024$,
$[-0.0150,+0.0217]$).  Moreover, shuffling InternVL frames produces a clear TRC change
but only a $-1.33\%$ accuracy change relative to chronological accuracy; its paired
95\% interval, scaled by the observed chronological accuracy, is
$[-4.44\%,+1.33\%]$ and includes zero.  Duplicate and cross-clip inputs reduce
InternVL accuracy by $15.11\%$ and $48.89\%$, respectively, relative to chronological
input.  These conditions also remove or replace visual evidence, so they cannot
isolate TRC as the causal mechanism.  DeepSeek's saved run records correctness only
for chronological input ($63.28\%$), so accuracy under its other input controls
cannot be recovered from the routing measurements.

Across both models, TRC is sensitive to controlled changes in sequence ordering and,
more strongly, to disruptions in visual continuity.  This establishes TRC as a useful
routing diagnostic in the evaluated sampled-frame settings.  It does not establish
that higher TRC predicts answer correctness or causes better task performance.
\end{reviewchange}

\subsection{What TRC Does and Does Not Establish: Limitations}
\label{sec:trc-limitations}

TRC has four main limits.  First, low overlap can be the correct response to a scene
change, while high overlap can reflect expert collapse; neither value is inherently
good.  Second, chance overlap depends on the number of experts, $M$, token count, and
expert popularity, so comparisons require matched budgets, per-layer reporting, and a
permutation baseline.  Third, set overlap ignores routing weights, which is why weighted
TRC is reported alongside TRC@8.  Fourth, both evaluations use sampled frames rather
than native-video inference, and their different expert counts, precision, preprocessing,
and frame encodings make absolute cross-model TRC levels non-comparable.  TRC should
therefore be interpreted with task quality, routing entropy, expert load, and scene
boundaries, not used as a training objective on its own.

\end{reviewchange}

\subsection{Long-Form Video: From Frame-Level Compression to Scene-Level Abstraction}

As video length extends from seconds to hours, frame-level token compression becomes
insufficient: even compressing each frame to a single token would leave thousands of
tokens for an hour-long video.  Long-video efficiency therefore shifts from local
token pruning toward hierarchical and memory-augmented abstraction~\citep{zou2024secondshoursreviewingmultimodal}.
Scene-level hierarchies segment videos into semantic episodes, summarise each episode
into a compact representation, and perform long-range temporal reasoning over the
episode sequence rather than raw frames~\citep{faure2025hermestemporalcoherentlongformunderstanding}.  This
hierarchy introduces a new sequencing constraint: episode order must be preserved for
causal reasoning, and summaries must retain enough scene context for downstream
queries.  Multi-granularity representations extend the hierarchy by combining frame-,
clip-, and scene-level features into a unified multi-scale token sequence, trading
richer context for higher token cost~\citep{li2026mmvirmultimodalmultigranularityrepresentation}.

The efficiency-understanding tension at the long-form scale is sharpest for real-time
video.  Online pruning can bound latency by using a lightweight preliminary attention
pass to estimate token importance before committing to full-attention computation
~\citep{jin2025streamingassistantefficientvisualtoken}.  A complementary approach compresses the temporal token
sequence directly inside the VLM context window without retraining, exploiting the
fact that abstract predictive video representations already encode temporal
redundancy structurally and are therefore cheaper to compress~\citep{wang2025lvclightweightcompressionframework,drozdov2024videorepresentationlearningjointembedding}.

\subsection{Self-Supervised Foundations for Efficient Video Representations}

The efficiency of downstream compression depends critically on the structure of
pre-trained video representations.  Reconstruction-based pre-training with $90\%$
masking forces each visible token to carry high information content, structurally
reducing temporal redundancy before any downstream pruning is applied
~\citep{tong2022videomaemaskedautoencodersdataefficient}.  Predicting abstract representations rather than pixels pushes
this effect toward semantically organised temporal encodings that are inherently more
compressible~\citep{drozdov2024videorepresentationlearningjointembedding}.  Cross-modal redundancy can also be reduced during
video-language pre-training by weakening alignment to visual regions that carry little
text-relevant information~\citep{wu2022rapredundancyawarevideolanguagepretraining}.

\section{KV Cache Optimisation}
\label{sec:kvcache}

\begin{figure}[t]
\centering
\includegraphics[width=0.92\textwidth]{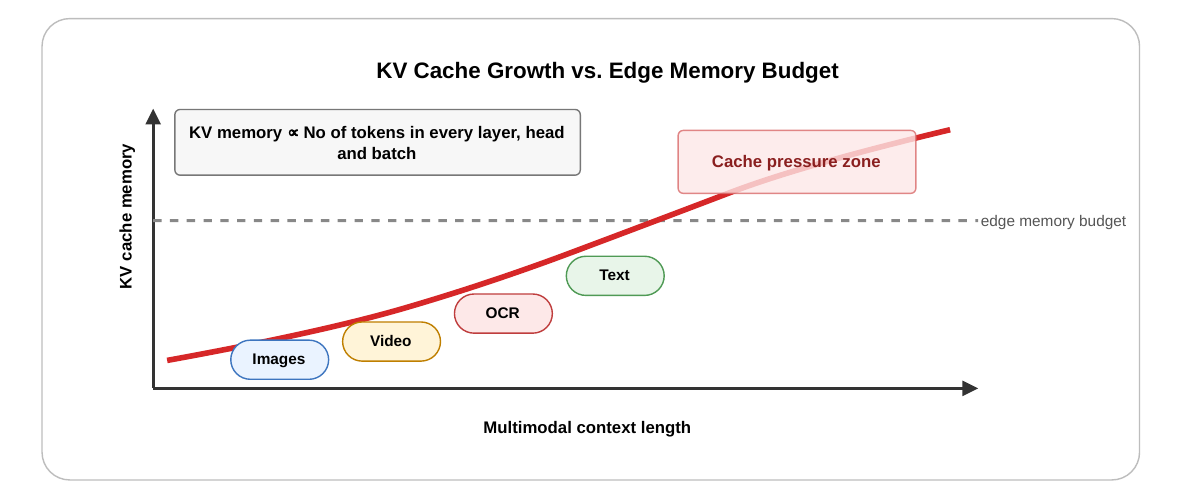}
\begin{reviewchange}
\caption{Illustrative KV-cache growth under an edge-memory budget.  For a standard
dense cache, retained tokens contribute key and value states across layers, KV heads,
batch elements, and numerical precision, producing the linear scaling in the text.
The dashed line represents a device-specific budget.  Crossing it may require
offloading, recomputation, compression, or eviction; the resulting latency and energy
change depends on the implementation and hardware.}
\label{fig:kv-cache-growth-budget}
\end{reviewchange}
\end{figure}

The KV cache enables efficient autoregressive generation by storing the key and value
states produced by causal self-attention~\citep{vaswani2023attentionneed}.  This avoids recomputing
attention projections for the full prefix at every decoding step, but it moves the
bottleneck from arithmetic to memory residency.  In multimodal edge inference this
topic is crucial because image patches, video frames, OCR spans, and text tokens all
leave cached states behind during generation.  For a decoder with batch size $B$,
retained token length $T$, number of layers $N$, KV heads $H$, head dimension $d$, and
element size $p$ bytes, the cache footprint is approximately
\[
2BTNHdp \ \text{bytes},
\]
where the factor of two comes from storing both keys and values.  In words, KV memory
grows linearly with tokens, layers, heads, hidden dimension, precision, and batch size.
Long-sequence transformer variants reduce attention complexity architecturally
~\citep{beltagy2020longformerlongdocumenttransformer,zaheer2021bigbirdtransformerslonger}, and efficient-NLP surveys place such methods within a
broader compression toolkit~\citep{treviso2023efficientmethodsnaturallanguage}; however, they do not by themselves solve
the deployment-time failure condition that every retained prefix token still occupies
cache memory.  KV-cache optimisation therefore asks which cached states must remain
exactly addressable during prefill and decoding, and which can be virtualised,
compressed, or evicted before the edge memory budget is exceeded.
In the terminology of Section~\ref{sec:codesign}, this is why KV-cache optimisation sits
at the cache-to-reasoning boundary rather than inside the four-stage chain: it stores
states created after upstream compression, routing, and precision decisions, then turns
those earlier choices into later memory and evidence-retention constraints.
Figure~\ref{fig:kv-cache-growth-budget} visualises this memory-pressure curve for
multimodal contexts under an edge-device budget.

\subsection{Serving-Level Cache Virtualisation}

Before making decisions about which tokens should be pruned out, the serving
system must also ensure that its efforts in preserving tokens do not waste any
cache resources on those very tokens it decides to keep. Contiguous cache
allocation does not fit the autoregressive serving paradigm well because the length
of the incoming requests increases dynamically, stops at some point in time, and
usually shares prefixes. The idea behind PagedAttention~\citep{kwon2023efficientmemorymanagementlarge}
is that in order to overcome such limitation of the serving infrastructure, KV
caches could behave more similarly to virtual memory: the caches for different
requests can be partitioned into blocks of equal sizes and allocated
non-contiguously, allowing dynamic allocation, prefix sharing, and copy-on-write
mechanisms in decoding branches. This way, the semantics of cache contents remain
intact while fragmentation and duplication of data in cache are avoided.

\subsection{Streaming Retention and Heavy-Hitter Eviction}

Having dealt with efficient memory allocation, another issue is that token age is a bad
measure for token importance. While the naive approach uses the sliding window which
always keeps the latest state while removing the oldest one, streaming generation
becomes difficult when the early position of tokens functions as an \emph{attention sink}.
The combination of keeping both the initial attention sink tokens and the sliding window
helps improve streaming performance without tuning~\citep{xiao2024efficientstreaminglanguagemodels}.
This is still not enough, because tokens that have historical importance do not necessarily
come from the front part of the prompt. The heavy hitter eviction approach keeps the small number
of historical tokens that attract large amounts of attention mass along with the sliding window~\citep{zhang2023h2oheavyhitteroracleefficient}.
This leads us to a more accurate principle: the KV eviction should focus on attentional structure,
not only recency.

{Transferring that attention-based retention principle to multimodal
reasoning may be fragile and has not been established by the text-only evidence.}
Multi-hop benchmarks such as
HotpotQA, MuSiQue, and 2WikiMultiHopQA require evidence to be composed across separated
facts~\citep{yang2018hotpotqadatasetdiverseexplainable,trivedi2022musiquemultihopquestionssinglehop,ho2020constructingmultihopqadataset}; the analogous MLLM case may
involve a small visual region, an OCR token, and a distant text query.  Such evidence
can be low-frequency or temporarily low-attention before it becomes decisive.  Thus the
  same attention-based heuristics that reduce cache size could delete latent evidence
before the model reaches the reasoning step that requires it.  This multimodal
extension should not be attributed to the text-only H$_2$O evaluation.

\subsection{Prompt-Aware and Layerwise Cache Compression}

Global eviction rules also ignore that attention behaviour varies across heads and
layers.  If a short observation window near the end of the prompt already reveals which
positions a head will rely on during generation, then compression can be made
prompt-aware rather than static.  SnapKV operationalises this observation by clustering
and retaining important KV positions per attention head, reducing long-context memory
and decoding cost without fine-tuning~\citep{li2024snapkvllmknowslooking}.  Yet head-level selection still
allocates cache uniformly across depth, even though lower layers tend to distribute
attention broadly while upper layers funnel information into fewer critical positions.
PyramidKV turns this layerwise asymmetry into a cache schedule, retaining more entries
in lower layers and fewer in upper layers~\citep{cai2025pyramidkvdynamickvcache}.

The other weakness is that of granularity. Token-based eviction may retain individual
salient states while breaking the semantic units they form.  ChunkKV uses semantic
chunks rather than individual tokens as the basis for compression, but its original
evaluation is text-only~\citep{liu2025chunkkvsemanticpreservingkvcache}.
{Chunk-level retention may also be relevant to visual patches, OCR
spans, subtitles, and dialogue turns, but that multimodal transfer requires direct
evaluation.}

\begin{reviewchange}
\subsection{Text-Only vs Multimodal Methods}
\label{sec:kv-domains}

The original evaluation domain materially limits what can be
claimed for a cache method.  PagedAttention, StreamingLLM, H$_2$O, SnapKV, PyramidKV,
ChunkKV, and GraphKV were introduced and evaluated for text-only LLM workloads; their
mechanisms may be reusable in an MLLM stack, but the original results do not establish
retention of visual or cross-modal evidence.  ZipVL, MadaKV, and RetentiveKV evaluate
multimodal models directly and can support multimodal-native claims within their stated
benchmarks.  Table~\ref{tab:kv-domain} makes this boundary explicit.

\begin{table}[t]
\centering
\caption{Original evaluation domain of representative KV-cache methods.  ``Text-only''
and ``multimodal'' describe the evidence in the cited paper, not all possible later
uses of an implementation.}
\label{tab:kv-domain}
\small
\renewcommand{\arraystretch}{1.12}
\setlength{\tabcolsep}{3pt}
\begin{tabular}{p{0.17\linewidth} p{0.25\linewidth} p{0.18\linewidth} p{0.31\linewidth}}
\toprule
\textbf{Method} & \textbf{Primary mechanism} & \textbf{Original evaluation domain} & \textbf{Transfer boundary} \\
\midrule
PagedAttention~\citep{kwon2023efficientmemorymanagementlarge} & Block-based cache virtualisation & Text-only LLM serving & Systems result; does not test visual-evidence retention \\
StreamingLLM~\citep{xiao2024efficientstreaminglanguagemodels} & Attention sinks plus recent window & Text-only language models & Multimodal use is unvalidated \\
H$_2$O~\citep{zhang2023h2oheavyhitteroracleefficient} & Heavy-hitter plus recent-token eviction & Text-only LLMs & Attention importance may differ for visual tokens \\
SnapKV~\citep{li2024snapkvllmknowslooking} & Head-wise prompt observation and clustering & Text-only LLMs & Original long-context results are not MLLM evidence \\
PyramidKV~\citep{cai2025pyramidkvdynamickvcache} & Layer-dependent cache budget & Text-only LLMs & Layer pattern requires multimodal revalidation \\
ChunkKV~\citep{liu2025chunkkvsemanticpreservingkvcache} & Semantic-chunk retention & Text-only LLMs & Visual/OCR chunk transfer is inferred \\
GraphKV~\citep{li2025graphkvbreakingstaticselection} & Dependency-graph eviction & Text-only long-context LLMs & Cross-modal graph construction is untested \\
ZipVL~\citep{he2024zipvlefficientlargevisionlanguage} & Joint prefilling sparsification and cache compression & Multimodal LLMs & Direct MLLM evidence on reported VQA settings \\
MadaKV~\citep{li2025madakvadaptivemodalityperceptionkv} & Modality-adaptive eviction & Multimodal LLMs & Direct MLLM long-context evidence \\
RetentiveKV~\citep{liu2026retentivekvstatespacememoryuncertaintyaware} & Entropy-guided state-space memory & Multimodal LLMs & Direct evidence on reported multimodal benchmarks \\
\bottomrule
\end{tabular}
\end{table}
\end{reviewchange}

\subsection{Graph-Based and Hybrid Eviction}

Chunkwise retention can still miss non-local dependencies in a flattened sequence.
GraphKV models tokens as graph nodes and updates their importance through graph
relations~\citep{li2025graphkvbreakingstaticselection}.
{Because GraphKV's original evidence is text-only, constructing a graph
that reliably links visual patches, OCR spans, and question tokens remains a plausible
multimodal extension rather than an established capability.}

{The same calibration issue may also arise across optimisation stages.}
If upstream compression has already changed the retained visual-token distribution, a
fixed downstream eviction rule may no longer be calibrated to it.  Direct evidence for
joint visual sparsification and cache compression comes from ZipVL, which reports
$2.3\!\times$ prefilling acceleration and $2.8\!\times$ decoding-throughput improvement
with a $0.5\%$ VQAv2 accuracy reduction in its evaluated setting
~\citep{he2024zipvlefficientlargevisionlanguage}; this result does not establish the
broader over-eviction mechanism for all fixed policies.

\subsection{Cross-Modal and State-Space Cache Compression}

{The MadaKV evaluation directly reports that visual and textual cache
entries exhibit different attention patterns.}  Its modality-adaptive eviction reports a
$1.3\!\times$--$1.5\!\times$ decoding-latency reduction on the evaluated multimodal
long-context tasks~\citep{li2025madakvadaptivemodalityperceptionkv}.  The result supports
modality-specific cache allocation in those settings.

Finally, if exact KV storage remains too expensive even after modality-aware eviction,
the cache need not be represented entirely as discrete key-value states.  A compact
state-space memory can absorb part of the history, while entropy-based eviction
preserves entries with the most unique information.  This hybrid cache-memory design
achieves $5\!\times$ KV compression with $1.5\!\times$ decoding acceleration on MLLM
benchmarks~\citep{liu2026retentivekvstatespacememoryuncertaintyaware}.  These methods
show that cache policy changes which states remain exactly addressable during
generation; the downstream reasoning effect remains conditional on the task.

The reported speedups and compression ratios throughout this section---$2.3\!\times$
prefilling acceleration with $2.8\!\times$ decoding-throughput improvement at a $0.5\%$
VQAv2 loss~\citep{he2024zipvlefficientlargevisionlanguage},
$1.3\!\times$--$1.5\!\times$ decoding gains~\citep{li2025madakvadaptivemodalityperceptionkv}, and $5\!\times$ compression
with $1.5\!\times$ acceleration~\citep{liu2026retentivekvstatespacememoryuncertaintyaware}---are typically measured as
benchmark-level losses or accuracy parity on the benchmarks used by each method.  This
is narrower than ``no information loss.''  {A discarded state may still
matter for a held-out multi-hop or fine-grained query, so cache efficiency should be
interpreted relative to the evaluation distribution and tested with targeted
evidence-retention probes.}

\section{Mixture-of-Experts: Architectures and Routing}
\label{sec:moe}

Mixture-of-Experts architectures improve scaling by activating only a small subset of
expert feed-forward networks for each token.  Their appeal is that sparse activation can
increase model capacity without increasing per-token arithmetic in the same proportion,
making MoE models attractive for efficient multimodal inference.  The edge failure case
is that routing is not merely a mathematical selection rule: each
expert assignment becomes a memory-access and scheduling decision, and shifted token
distributions can produce load imbalance, expert collapse, or unstable specialisation.
This section therefore examines routing not only as an accuracy mechanism, but as the
control surface that determines whether sparse capacity remains usable under
compression, quantization, and hardware constraints.

\subsection{The Routing Design Space: Differentiability, Balance, and Collapse Avoidance}

The main difficulty with MoE routing is balancing three somewhat conflicting goals,
namely \emph{differentiability} (gradients must flow through the routing layer in order
to facilitate proper specialisation of the experts), \emph{load balance} (all experts need
to be roughly equally active in terms of token stream traffic to avoid inefficiency),
and \emph{non-collapse} (the router cannot fall into the habit of choosing just a few
experts all the time).  The classical Top-$k$ router with additional balancing losses
achieves the latter goal using regularisation, but fails in differentiability due to
the discontinuity introduced by the argmax operation.

Different routing approaches address different parts of this design space.  Continuous
ReLU Routing replaces hard Top-$k$ selection with a differentiable construction
~\citep{wang2025remoefullydifferentiablemixtureofexperts}; auxiliary-loss-free methods
target balance without a tuned auxiliary loss
~\citep{wang2024auxiliarylossfreeloadbalancingstrategy}; and eigenvector routing is
designed to provide non-collapsing assignments under its stated construction
~\citep{do2026eigenvectorsexpertstrainingfreenoncollapsing}.  A unified routing system could combine continuous
optimisation, structural load balance, and collapse-resistant expert geometry rather
than treating them as separate fixes.

\subsection{Expert Collapse: Quantitative Diagnosis and Failure Mode Taxonomy}
\label{subsec:collapse}

Expert collapse---the router repeatedly selects a small number of experts while
underutilising others---is a key failure mode in MoE training.  It is
critical to understand what collapse means in terms of defining its mitigation
mechanisms, since ``collapse'' can refer to different concepts with varying
thresholds and implications.  This survey uses utilisation below the balanced Top-$k$ reference share $k/N$
as a diagnostic flag, not as a universal collapse threshold.  For 64 experts with
$k=2$, the reference share is $3.125\%$.  Empirically, collapse can spread across depth, with roughly one-third
of sparse MoE layers drifting toward single-expert routing when balancing is weak
\citep{sha2026sparsemixtureofexpertsroutingvisual}.  In conditional generation,
\citet{sha2026sparsemixtureofexpertsroutingvisual} report a form termed
\emph{selective deadlock}, in which class-conditioned traffic concentrates on an expert
and can cause capacity overflow.  Post-hoc utilisation statistics reveal concentration
after it occurs; spectral and Gini-based diagnostics have therefore been proposed as
earlier indicators~\citep{delibasoglu2026spectralmanifoldregularizationstable}.  Their
thresholds should be calibrated to the architecture rather than treated as universal.

Hierarchical expert architectures provide one mitigation: gate
networks based on task specialization create soft limits that enhance load balancing
and specialized task performance~\citep{wang2024homehierarchymultigateexperts}; they do
not by themselves guarantee balanced routing for every workload.

\subsection{Representation Collapse in Tokenisation}
Analogous collapse behaviors emerge within vector quantization (VQ) architectures, where codebook collapse refers
to the phenomenon wherein a few codebook elements become the exclusive choice of
the model while the rest remain unused. Codebook collapse exhibits a rich-get-richer
behavior, whereby codebook elements with assignment probabilities greater than the
mean assignment probability experience norm growth, receive an increasing number of
assignments, and thereby further contribute to the disparity~\citep{zhao2024representationcollapsingproblemsvector}.
The use of a simple linear layer on codebook elements may counteract such a
feedback mechanism by ensuring linear independence among codebook elements
~\citep{zhu2025addressingrepresentationcollapsevector}.  Nevertheless, this static solution does not
carry over into the domain of MoE since the MoE utilization problem involves
determining a probability distribution dynamically over expert weight matrices based
on the input token mini-batch; unlike the fixed codebook structure of VQ, no
combination of expert weight matrices and the token mini-batch can produce an
orthogonal or linearly independent set of experts.

\subsection{MoE in Vision and Multimodal Models}

Whether MoE routing benefits visual tasks depends on the computational regime.  The
cited study reports benefits when specialised experts complement the backbone and costs
when memory access or load imbalance offsets sparse computation
~\citep{sun2026doessparsemoehelp}.  {Strong compression may change that
balance, but a general advantage for compressed over uncompressed backbones has not
been established.}
CosMoES proposes compact sparse architectures to minimize MoE routing overhead when operating with few parameters~\citep{huber2025cosmoescompactsparsemixture},
and lookup-based experts, which share an embedding table to drastically lower expert parameter costs~\citep{jie2025mixturelookupexperts}.

\subsection{High-Throughput MoE Serving}
\label{sec:serving}

High-performance MoE serving turns routing into a systems problem.  Sparse activation
does not guarantee low latency: realised performance also depends on expert placement,
batching, cache residency, communication, and the target hardware.  These concerns are
especially relevant when active weights move through a constrained memory hierarchy.

\subsubsection{System-Level Analysis and Scheduling}

MoE-Lens models the transition between compute-bound and memory-bound regimes and
reports up to $25.5\!\times$ speedup ($4.6\!\times$ on average) relative to the
baseline used in that study~\citep{yuan2025moelenshardwarelimithighthroughput}.  These
figures are workload- and baseline-specific rather than general serving limits.

Memory stalls can be reduced either by predicting routing paths early enough to
prefetch required experts, or by continuously adapting the prefetch policy as workload
statistics change.  Predictive routing and adaptive prefetching report latency and
memory benefits under their evaluated workloads
~\citep{2024arXiv241017954H,shen2025expertflowadaptiveexpertscheduling}; realised gains
depend on prediction accuracy, cache capacity, and transfer bandwidth.

Serving efficiency can also be improved before memory management begins.  Sensitivity-
based gating skips low-sensitivity experts for compatible token types, reducing the
average number of activated experts by $25\%$ and achieving a $1.35\!\times$ speedup
without memory-management overhead~\citep{Zhong_2024}.  When workload heterogeneity is
the dominant issue, task-aware expert loading adapts memory allocation to request
characteristics and improves utilisation across diverse tasks~\citep{tairin2025emoetaskawarememoryefficient}.

\subsubsection{Disaggregated Serving}

Co-locating attention and expert modules creates cache interference because attention
benefits from full KV residency while experts benefit from parallel weight movement.
Disaggregating the two module types onto separate GPU sub-clusters preserves fast KV
access for attention and lets expert execution proceed without cache interference,
achieving up to $3.9\!\times$ higher per-GPU throughput than co-located serving
~\citep{zhang2026janusdisaggregatingattentionexperts}.

\section{Quantization for Efficient Inference}
\label{sec:quant}

Quantization represents weights or activations at lower precision, reducing nominal
storage and bytes transferred.  Realised compute, latency, and energy benefits depend on
kernel and hardware support.  In multimodal and MoE models, published studies identify
additional risks: low-bit numerical degradation, changes in router decisions, and
different sensitivity across modalities.
\begin{table}[t]
\centering
\caption{Taxonomy of quantization failure modes in VLMs and MoE models. Rec. = recoverable via calibration/fine-tuning without architectural changes ($\sim$ = partial).}
\label{tab:quant-failures}
\small
\renewcommand{\arraystretch}{1.3}
\begin{tabular}{p{0.24\linewidth} p{0.22\linewidth} p{0.05\linewidth} p{0.27\linewidth} p{0.12\linewidth}}
\toprule
\textbf{Failure mode} & \textbf{Representative evidence} & \textbf{Rec.} & \textbf{Mitigation} & \textbf{Scope} \\
\midrule
Signal degradation \citep{zhou2026signaldegradationcomputationcollapse} & Gradual degradation in the 4-bit regime & \checkmark & Calibration or fine-tuning & VLM, MoE \\
Computation collapse \citep{zhou2026signaldegradationcomputationcollapse} & Catastrophic cliff in the 2-bit regime & $\times$ & Mixed precision / clipping / structural repair & VLM, MoE \\
Routing inconsistency \citep{fu2026eaquantenhancingposttrainingquantization} & Routing-aware PTQ recovers $1.15\%$--$2.28\%$ average score at W4A4/W3A4 & $\sim$ & Calibration- \citep{fu2026eaquantenhancingposttrainingquantization} or constraint-centric \citep{park2026valueandstructurealignmentroutingconsistentquantization} & MoE only \\
Modality-heterogeneous degradation \citep{zhong2026breakingmodalityheterogeneitylowbit} & SplitQ retains $93.5\%$ of FP16 at W3A3 ($69.5$ vs.\ $74.3$) & $\sim$ & Modality-separated quantization \citep{xue2026vlmqtokensaliencydrivenposttraining} & VLM only \\
Speculative decoding interaction \citep{zhang2025speculativedecodingmeetsquantization} & Hierarchical design reaches $2.78\!\times$ speedup and $1.31\!\times$ over EAGLE-2 on 4-bit Llama-3-70B & \checkmark & Joint drafter--target design & Text LLM; multimodal transfer untested \\
\bottomrule
\end{tabular}
\end{table}
Table~\ref{tab:quant-failures} condenses the main quantization failure regimes
discussed in this section and makes explicit that not all failures are recoverable by
the same kind of calibration strategy.
\subsection{Quantization Failure Modes: A Quantitative Taxonomy}
\label{subsec:qfail}

Before exploring quantization techniques separately, we first need to understand their
failure mechanisms using evidence that the cited papers actually report. One useful
distinction separates gradual signal degradation from catastrophic computation collapse,
because the two modes require different remedies~\citep{zhou2026signaldegradationcomputationcollapse}.  Signal
degradation is the regime in which quantization perturbs computation while preserving
the overall processing pattern, whereas computation collapse is the regime in which key
operations fail and the forward pass breaks down. For LLaMA-style transformers,
\citet{zhou2026signaldegradationcomputationcollapse} report a workable 4-bit regime and
a severe 2-bit degradation regime for the evaluated LLaMA-style models.  These are not
universal bit-width thresholds.  The practical
significance of this taxonomy is that gradual degradation can often be mitigated by
calibration or lightweight fine-tuning, whereas collapse usually requires architectural
intervention such as clipping, mixed precision, or structural repair.

\begin{figure}[t]
\centering
\includegraphics[width=0.8\textwidth]{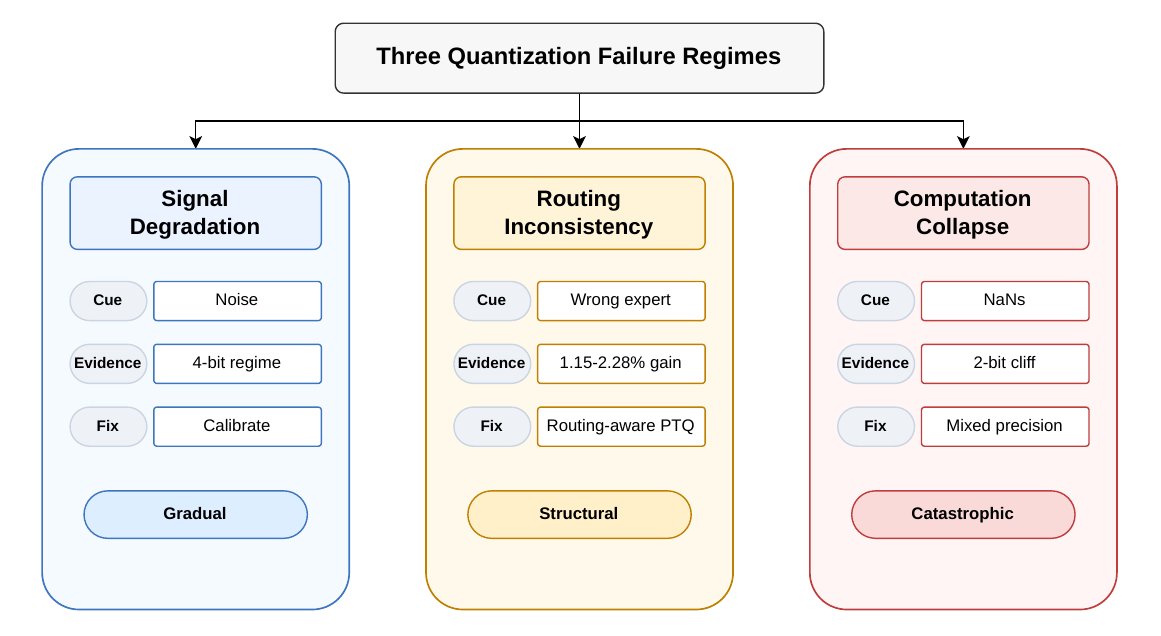}
\begin{reviewchange}
\caption{Comparison of three quantization failure regimes in VLM and MoE inference.
Signal degradation is gradual loss of fidelity; routing inconsistency is a structural
MoE-specific change in expert selection; and computation collapse is catastrophic
numerical instability.  The categories and mitigations are tied to the evaluated
precision settings discussed in the text rather than presented as universal bit-width
thresholds.}
\label{fig:quant-failure-flowchart}
\end{reviewchange}
\end{figure}

MoE quantization adds a third failure mode: \emph{routing inconsistency}, where noise
in router logits changes expert assignments post-deployment.  Rather than claiming a
universal assignment-change threshold, the clearest source-backed evidence is
outcome-based: routing-aware post-training quantization recovers $1.15\%$--$2.28\%$
average score across three MoE architectures under W4A4 and W3A4 settings.  This
establishes task relevance for those settings without implying a universal routing
threshold
~\citep{fu2026eaquantenhancingposttrainingquantization}.  Figure~\ref{fig:quant-failure-flowchart} recasts these
failure modes as a side-by-side comparison of three regimes. Rather than presenting a
decision path, the figure highlights what breaks in each case, what evidence marks the
regime, and which mitigation is most appropriate, making clear that quantization can
perturb signal fidelity, sparse routing, or numerical stability at different levels of
severity.

\subsection{Addressing Routing Inconsistency: Structure-Preserving Quantization}

Two complementary strategies address routing inconsistency.  Calibration-centric
quantization exposes the router to token distributions that most frequently produce
assignment changes and then minimises logit perturbation on those distributions
~\citep{fu2026eaquantenhancingposttrainingquantization}.  Constraint-centric quantization instead preserves the full-
precision Top-$k$ assignment structure on calibration data, penalising quantized
parameter configurations that change expert assignments
~\citep{park2026valueandstructurealignmentroutingconsistentquantization}.  Both papers
report benefits under their calibration and evaluation settings; robustness to unseen
modalities or upstream compression remains open.

\subsection{Global and Mixed-Precision Quantization for MoE}

Mixed-precision MoE quantization shifts the allocation unit from layers to experts.
Global expert-importance scoring assigns higher precision to rare but critical experts
and aggressively compresses routine experts, substantially reducing memory demand while
maintaining accuracy~\citep{deng2026gemqglobalexpertlevelmixedprecision}.  Low-rank compensation complements
this allocation by learning fine-grained corrections for the most critical experts,
improving the bandwidth-accuracy trade-off of offloaded MoE inference~\citep{liu2025bandwidthefficientadaptivemixtureofexpertslowrank}.

At extremely low precision, the central trade-off becomes representation expressivity
versus bitwise efficiency.  Binary experts combine $\{-1,+1\}$ weights with learned
compensation and bitwise kernels~\citep{zhao2026mobieefficientinferencemixture}.
Ternary experts add zero, matching sparse activation more naturally than a generic
bit-width choice~\citep{wang2026motemixtureternaryexperts}.  Scaling-law analysis helps
identify when binary, ternary, or 4-bit architectures are preferable
~\citep{liu2025paretoqimprovingscalinglaws}.

\subsection{Modality-Heterogeneous Quantization in VLMs}

{VLM quantization studies directly report different activation
distributions and precision sensitivities for visual and textual tokens.}  In the cited
W3A3 evaluation, modality-separated SplitQ retains $93.5\%$ of the FP16 score (69.5
versus 74.3)~\citep{zhong2026breakingmodalityheterogeneitylowbit}.  This setting-specific
result supports modality-aware treatment without establishing universal variance or
outlier thresholds.  Second-order curvature information can identify weight
dimensions that are sensitive to quantization noise and retain higher precision for
them~\citep{xue2026vlmqtokensaliencydrivenposttraining}.  Binary and 2-bit VLMs
need architecture changes beyond post-training calibration, since
calibration alone is not enough~\citep{wang2025bivlmpushingultralowprecision,guo2025speedqstagedprocessingenhanced}.

\subsection{Cross-Technique Interactions: Speculative Decoding and Quantization}

{The cited speculative-decoding study directly reports that quantizing
the target can change drafter acceptance and that joint design recovers efficiency,
including a reported $2.78\!\times$ speedup and $1.31\!\times$ improvement over its
EAGLE-2 comparison on 4-bit Llama-3-70B
~\citep{zhang2025speculativedecodingmeetsquantization}.}  This
interaction is a further established instance of the cross-component pattern in
Section~\ref{sec:codesign}:
a precision change applied to the target model propagates into a separate component
(the drafter) whose calibration assumptions were not designed to account for it,
producing a system-level efficiency loss that single-component evaluation would miss.

\section{Edge and On-Device Deployment}
\label{sec:edge}

Edge deployment exposes device-specific limits in fast memory, bandwidth, transfer
latency, power, and sustained thermal operation.  An MoE expert bank may exceed fast
on-chip capacity, in which case selected experts are cached, prefetched, or loaded from
a slower tier.  The frequency and cost of those transfers depend on routing reuse,
placement, and the serving implementation.

\subsection{Expert Storage Hierarchies}

Hot-cold hierarchies control expert placement within a device.  In the cited design,
shared components and selected experts occupy different memory tiers, and expert FFN
weights are prefetched using routing predictions.  The authors report speedups over
their all-in-storage baseline~\citep{10906629}; the result is
specific to that hierarchy and workload.

Prefetch predictors and tiered storage strategies can increase the depth of the hierarchy to support larger sparsity-based models on mobile devices \citep{10906629}.

Offloading techniques can be complemented by precision-aware nested representations; when weights are stored in a quantized manner, experts can be accessed at varying degrees of precision based on their usage, thereby saving memory space while not being limited to one level of precision~\citep{Wang_2025}.

\begin{figure}[t]
\centering
\includegraphics[width=\textwidth]{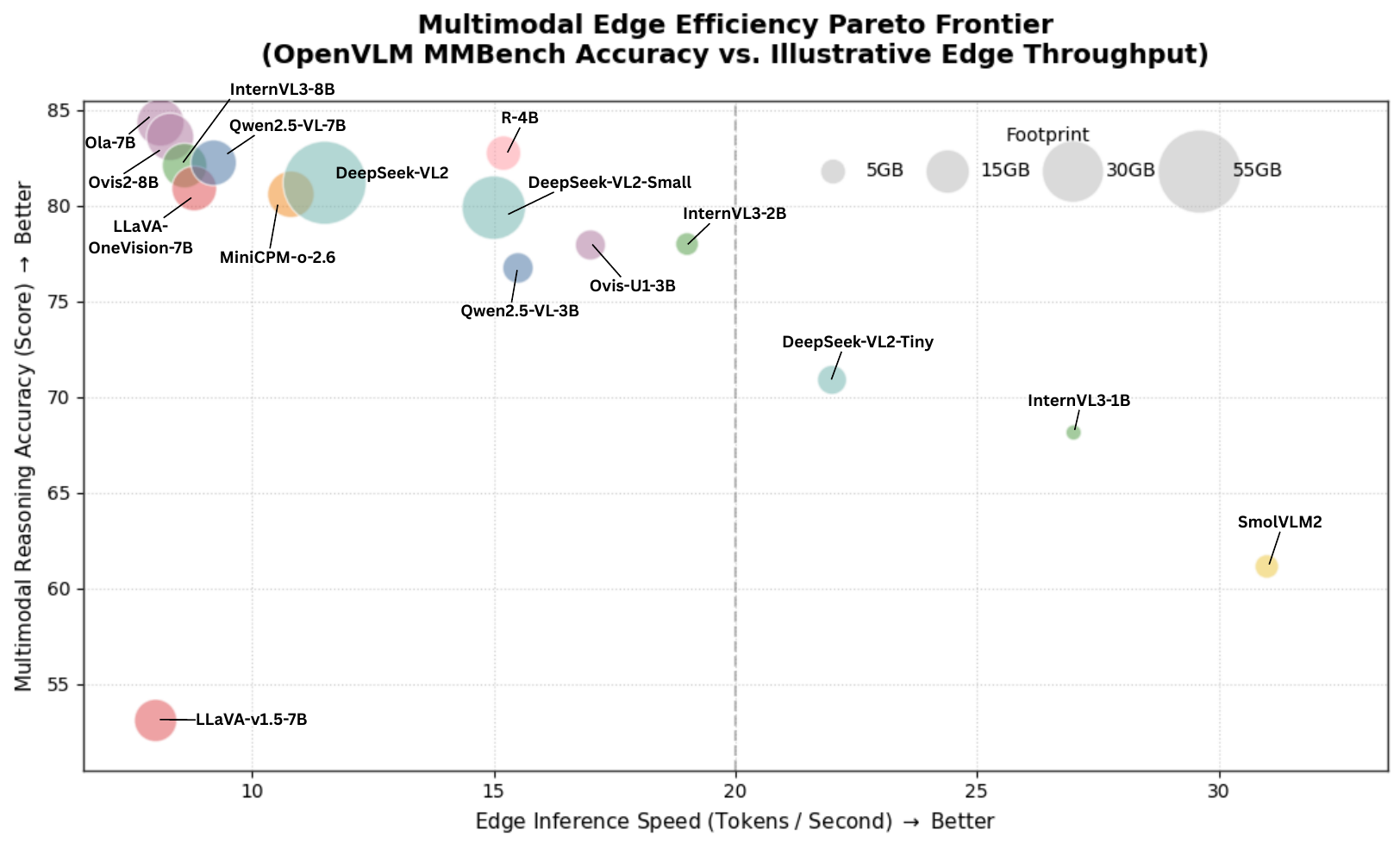}
\caption{Illustrative multimodal edge-efficiency Pareto view.  Vertical-axis MMBench
V1.1 scores were compiled from the OpenCompass OpenVLM leaderboard, and bubble areas
approximate FP16 parameter footprints from public model cards.  Horizontal positions
are illustrative design placements rather than measured cross-device throughput.
The empty upper-right region is therefore a conceptual target, not an empirical claim.}
\label{fig:edge-pareto-frontier}
\end{figure}
\noindent\emph{Figure 10 value provenance.}  Benchmark scores were taken from the
OpenCompass OpenVLM Leaderboard
(\url{https://opencompass-open-vlm-leaderboard.hf.space/}, accessed 22 August 2026),
and approximate model sizes from the corresponding public model cards.  No plotted
throughput value is reported as a measurement from a paper or leaderboard.

\subsection{Lossless Compression and Scheduling}

Lossless expert-weight compression and cache-aware scheduling tackle two separate aspects of the device-side inference latency chain. The former aims to lower transfer costs,
while the latter maximizes cache hits through allocation of commonly triggered experts to 
adjacent cache addresses, resulting in an up to $72.77\%$ reduction in inference latency compared to the unoptimized MoE serving process \citep{yang2026zipmoeefficientondevicemoe}.

Heterogeneous expert quantization assigns different precision to experts based on
importance and device limits.  The cited work reports an improved accuracy--deployment
cost trade-off in its setting~\citep{11527015}; generalisation to other devices requires
measurement.

{Many throughput evaluations use a warmed, steady-state serving loop,
whereas sporadic mobile use may incur cold-start loading of the backbone and
experts~\citep{10906629}.  For short interactions,
that startup component could be a material fraction of total latency.}  A compressor or
quantizer that lowers steady-state transfer cost need not minimise cold-start latency,
because storage reads, deserialisation, kernel initialisation, and cache warming may
contribute differently.
Measuring the loading time and the inference latency separately, and assessing
both with an appropriate intermittent-use workload, is thus important for making edge
deployment claims representative; this aspect of loading latency has been
overlooked in most of the throughput-centric benchmarks summarized in
Section~\ref{sec:bench}. 
The very same memory hierarchy that influences the steady-state cost for
loading an expert dictates how a device can host all of the experts or needs to
coordinate with other devices, and this is one motivation behind the
distributed/federated deployment scenario considered in Section~\ref{sec:challenges}.

{Figure~\ref{fig:edge-pareto-frontier} should therefore be read as a
conceptual view of the edge-deployment trade-off rather than a controlled Pareto
study.}  A defensible frontier would
require measuring all models with the same runtime, precision, input distribution,
thermal protocol, and device.  The upper-right region is therefore a target for future
evaluation rather than a claimed ``hardware wall.''

\section{Benchmarking and Evaluation Frameworks}
\label{sec:bench}

Benchmarking determines whether claimed efficiency survives contact with real models,
real prompts, and real hardware.  This matters because compression, quantization,
routing, cache management, and edge scheduling can each improve one metric while
degrading another.  The current failure condition is evaluative fragmentation:
existing suites measure accuracy, latency, memory, power, routing cost, or localisation
in isolation, but rarely expose how an upstream optimisation creates downstream
reasoning or deployment failure.  This section therefore separates three questions:
what current benchmarks can measure, what they cannot measure, and what a unified
multimodal edge benchmark should report.

\begin{table}[t]
\centering
\caption{Current benchmark coverage for efficient multimodal inference.  Existing
benchmarks provide useful single-axis measurements, but no benchmark jointly measures
compression, routing, quantization, cache residency, and hardware behaviour under one
deployment protocol.}
\label{tab:benchmarks}
\small
\setlength{\tabcolsep}{4pt}
\begin{tabular}{p{0.2\linewidth} p{0.35\linewidth} p{0.35\linewidth}}
\toprule
\textbf{Benchmark} & \textbf{What it measures} & \textbf{Remaining gap} \\
\midrule
ImageNet & Image classification accuracy and throughput & Weak proxy for multimodal reasoning \\
DeiT / MAE / CLIP eval suites & ViT representation quality under token merging & Limited downstream MLLM coverage \\
VQA benchmarks & Answer accuracy under visual-language tasks & Often miss spatial and OCR degradation \\
Mixtral-class reasoning benchmarks & Language reasoning after MoE quantization & Do not expose visual-token routing shift \\
LLaVA-class eval suites & VLM accuracy and activation statistics & Limited hardware and energy reporting \\
OpenVLM MMBench V1.1 & Multimodal reasoning accuracy & No standard edge throughput protocol \\
LLMInferenceBench & Accelerator latency, throughput, and memory & Mostly hardware-centric, not multimodal-failure-centric \\
PalmBench & Mobile compressed-model accuracy-efficiency tradeoffs & Limited MoE and VLM interaction coverage \\
HW-GPT-Bench & Hardware-aware architecture-search comparisons & Simulated design space, limited deployment traces \\
CarAML & Power-performance characterisation & Not specialised to MLLM pipelines \\
MLPerf Power & Standardised power measurement & Energy axis only, not reasoning failure diagnosis \\
MoE-CAP & MoE cost-accuracy-performance tradeoffs & Does not include visual compression or modality-aware cache effects \\
LocateBench & Spatial localisation robustness & Capability-specific, not full deployment-system evaluation \\
\bottomrule
\end{tabular}
\end{table}
Table~\ref{tab:benchmarks} makes the evaluative gap concrete: existing suites measure
important slices of the problem, but none of them jointly stress compression, routing,
KV-cache behaviour, quantization, and hardware under a single deployment protocol.

\subsection{What Current Benchmarks Can Measure}

Current benchmarks are strongest when the measurement target is isolated.  Accelerator-
level suites standardise throughput, latency, and memory utilisation across AI hardware
~\citep{chittyvenkata2024llminferencebenchinferencebenchmarkinglarge}; mobile-focused suites evaluate compressed-model
accuracy-efficiency trade-offs on device classes where memory and thermal limits
dominate~\citep{li2025palmbenchcomprehensivebenchmarkcompressed}; hardware-aware architecture benchmarks support NAS
comparisons across simulated configurations~\citep{sukthanker2024hwgptbenchhardwareawarearchitecturebenchmark}; and power-performance
characterisation captures accelerator efficiency beyond latency alone~\citep{John_2024}.
On the model-quality side, VQA and OpenVLM-style leaderboards measure multimodal
reasoning accuracy, while localisation-specific benchmarks reveal failures hidden by
aggregate VQA scores~\citep{chiang2024locatebenchevaluatinglocatingability}.  For sparse models, cost-accuracy-
performance evaluation makes routing and expert movement visible rather than treating
MoE systems as ordinary dense models~\citep{jiang2025moecapbenchmarkingcostaccuracy}.

\subsection{What Current Benchmarks Cannot Measure}

{The surveyed benchmarks measure useful slices of the pipeline but do
not by themselves attribute a result to a cross-stage pathway.}  A VQA
answer can remain correct while localisation degrades
~\citep{chiang2024locatebenchevaluatinglocatingability}; throughput or power can improve
without revealing which token, cache, routing, or precision choice caused the change;
and expert-centric metrics expose movement cost without necessarily isolating why a
route changed~\citep{tschand2025mlperfpowerbenchmarkingenergy,jiang2025moecapbenchmarkingcostaccuracy}.

This matters because energy and latency can be shaped by memory movement as well as
arithmetic.  On the memory-bound settings studied by
\citet{chung2026joulesgodiagnosinginference}, a FLOP reduction alone need not reduce
wall-clock energy.  Because inference energy can scale sharply with accuracy targets in
the cited analysis~\citep{yang2024doubleexponentialincreasesinferenceenergy}, a
full-system evaluation should include sustained throughput, thermal behaviour,
memory traffic, and answer quality under one protocol.  Existing tools cover pieces of
this space---deadline-aware frequency scaling~\citep{yan2024polythrottleenergyefficientneuralnetwork}, training-side
power measurement~\citep{Latif_2025}, kernel-level NAS~\citep{la2025kernellevelenergyefficientneuralarchitecture}, and
energy-latency-accuracy trilemma formulations~\citep{tian2024energyefficiencynavigatingtrilemmaenergy}---but they do not
jointly diagnose multimodal compression-routing-quantization failures.

\subsection{Toward a Unified Multimodal Edge Benchmark}

{To address this gap, we propose that a benchmark for the systems
surveyed here report a deployment tuple rather than a single number:}
\[
\big(
\mathrm{accuracy},\ \mathrm{latency},\ \mathrm{energy},\ \mathrm{peak\ memory},
\mathrm{KV\ footprint},\ \mathrm{expert\ traffic},\ \mathrm{routing\ stability}
\big).
\]
The protocol should define hardware targets, prompt sizes, vision-token budget,
precision format, and decoding method, then measure systems and reasoning performance.
For dense VLMs, it should compare model size and visual-token quality under a common
compute budget~\citep{li2025inferenceoptimalvlmsneed}.  For MoE VLMs, it should report
expert use and transfer volume, routing before and after quantization, and cache hit rate
under visual-token compression.

The suggested benchmark could also feature stress tests aimed at activating the
interfaces discussed here, namely pre-routing compression, post-routing quantization,
KV-cache stress under long multimodal prompts, and sustained edge inference within the
thermal envelope.  For vision-language models using video MoEs, the routing stability component of
the deployment tuple could include \emph{Temporal Routing Consistency}
(TRC, Section~\ref{sec:trc}) for the video input before and after compression and quantization,
as an exploratory diagnostic beside downstream accuracy, not as a validated proxy for
temporal reasoning.  Evaluation of performance scalability through probing allows comparing
capacities across model sizes and architectures~\citep{castro2023scalableperformanceanalysisvisionlanguage},
and security analysis adds resistance to OOD or backdoor visual inputs~\citep{lyu2025backdooringvisionlanguagemodelsoutofdistribution}.
Efficiency improvements are most interpretable when reported with any measured quality
or systems regressions.

\section{Open Challenges and Future Directions}
\label{sec:challenges}

The evidence map in Table~\ref{tab:evidence-map} makes the main research boundary
explicit: several component-level effects are established, whereas most complete
compression--routing--quantization--hardware pathways remain untested.  The priorities
below therefore separate empirical gaps from mechanisms proposed by this survey.

{\textbf{Compression-Routing Interaction in MoE.}}  Priorities are
multi-seed controls; replication beyond the two evaluated MoE-VLMs, including
full-precision InternVL and additional modalities and pruning or merging methods such as
FastMMoE~\citep{xia2025fastmmoe}; causal tests separating routing
preservation from information preservation; and a joint compression--quantization
experiment that measures assignments, load balance, task quality, and systems cost
under all component and composed conditions.

\begin{reviewchange}
\textbf{TRC Validation for Video MoE.}  The control experiment establishes measurement
sensitivity in two sampled-frame MoE-VLM settings, but not task relevance or native-video
generalization.  Open questions include whether TRC correlates
with temporal reasoning after controlling for scene transitions,
token budget, expert count, Top-$k$, layer, and expert collapse; whether a weighted or
chance-adjusted variant is more informative; and whether compression or quantization
changes TRC and downstream accuracy together.  Only after those tests would a routing
consistency loss or recurrent routing state be justified.
\end{reviewchange}

{\textbf{Unified Cross-Modal Compression.}  Current token compression
methods largely treat visual and textual modalities independently.}  The modality heterogeneity
challenge identified by~\citep{zhong2026breakingmodalityheterogeneitylowbit} and~\citep{li2025madakvadaptivemodalityperceptionkv} points toward
modality-specific strategies, but a principled framework for joint cross-modal
compression---accounting for complementary inter-modal information---remains open.

\textbf{Theoretical Foundations of Expert Collapse.}  While multiple empirical
mitigation strategies have been proposed~\citep{wang2024auxiliarylossfreeloadbalancingstrategy,do2026eigenvectorsexpertstrainingfreenoncollapsing,delibasoglu2026spectralmanifoldregularizationstable}, and practical empirical markers such as utilisation below the $k/N$ balanced-routing reference and layerwise drift toward single-expert routing are beginning to emerge, a theoretical characterisation of when and why collapse occurs is
incomplete.  Analogies from vector quantization collapse~\citep{zhao2024representationcollapsingproblemsvector,zhu2025addressingrepresentationcollapsevector} are informative but do not directly transfer to
continuous-routing MoE contexts.

\textbf{Proactive Failure Mode Detection.}  The quantization failure modes identified
by~\citep{zhou2026signaldegradationcomputationcollapse} and routing collapse modes by~\citep{sha2026sparsemixtureofexpertsroutingvisual} are
currently detected only post-deployment.  The spectral Gini threshold~\citep{delibasoglu2026spectralmanifoldregularizationstable}
together with activation-level monitoring for collapse-prone low-bit layers~\citep{zhou2026signaldegradationcomputationcollapse} represent first steps
toward proactive monitoring, but neither is integrated into standard training pipelines.
Systematic incorporation of collapse and degradation monitors into training
dashboards, with automatic corrective interventions, represents a practically important
and theoretically interesting challenge.

\textbf{Long-Video Efficiency Benchmarks.}  Long-form video understanding~\citep{zou2024secondshoursreviewingmultimodal,faure2025hermestemporalcoherentlongformunderstanding} lacks benchmarks that jointly measure temporal reasoning
quality and computational efficiency.  Existing benchmarks reward accuracy without
penalising compute cost; future benchmarks should reward high-quality answers at low
inference cost.

{\textbf{Hardware-Aware Co-Design.}  The energy literature shows that
data movement is important in the evaluated memory-bound settings
~\citep{chung2026joulesgodiagnosinginference,yang2024doubleexponentialincreasesinferenceenergy,tschand2025mlperfpowerbenchmarkingenergy}.}
This motivates jointly measuring architecture, precision, memory traffic, and hardware
configuration.  It does not establish that quantization will outperform pruning on
every device; that comparison should be made with device-level latency and energy
measurements under a common workload.

{\textbf{Federated and Distributed Edge Inference.}  When a model does
not fit or meet latency targets on one device, coordination with another device is one
option alongside compression or offloading.}  Distributed inference introduces load
balancing, communication, and privacy trade-offs that merit device-level evaluation
~\citep{10906629,Wang_2025}.

{\textbf{Hallucination and Factual Accuracy Under Compression.} Token
compression, quantization, and KV-cache eviction have been assessed primarily by
accuracy, latency, and memory metrics in this survey, yet deleting or modifying visual
input and cached context could also influence what the model \emph{says} regardless of
changes in aggregate benchmark performance.} While a model that no longer has the ability to see
deleted visual input or reference the evicted KV buffer may choose to say "I don't
know," it might instead formulate a coherent response based on contextual input
without evidence. The impact of extreme compression, quantization, or eviction on the
rate of hallucination generation and whether such instances are skewed toward
infrequent but informationally dense tokens or experts like those that failed at
Interfaces~B and~C has yet to be fully characterized.

\begin{reviewchange}
\section{Conclusion}
\label{sec:conclusion}

This survey examines visual token compression, KV-cache management, MoE routing,
quantization, and edge deployment as connected stages of multimodal inference.
Its cross-stage framework separates directly measured relationships from supported
inferences and open hypotheses, identifying where a local efficiency intervention
may change the conditions encountered by downstream stages.

The two empirical studies provide evidence for specific parts of this framework.
On ScienceQA, visual-token pruning changes expert assignments among retained tokens
in both DeepSeek-VL2-Tiny and InternVL3.5-30B-A3B. Across all four measured pruning
budgets, L2 selection produces lower RACR than random selection within each model.
Accuracy does not follow the same ordering across models: InternVL's random
selection has higher observed accuracy at three budgets and ties L2 at the fourth.
Both InternVL identity controls reproduce baseline answers and routes at 100\%
retention. The reassignment and the lower RACR under L2 selection persist in
Qwen3.5-35B-A3B and Qwen3.5-122B-A10B on ScienceQA, MMMU-Pro, and MathVista,
where random selection is averaged over three seeds; on MMMU-Pro the two selections
are not reliably separated in accuracy. These results establish compression-induced reassignment, but do not
identify routing change as the cause of accuracy loss. On NExT-QA, TRC@8 and weighted
TRC respond to the expected sequence controls in both models, yet neither shows a
reliable association with correctness in the evaluated samples. TRC therefore
remains a diagnostic of routing continuity rather than an established measure of
video understanding.

The findings motivate reporting downstream routing behaviour alongside task quality
when evaluating multimodal efficiency. Their scope remains limited to the measured pruning and sampled-frame protocols, with model-specific 
precision and preprocessing.
Multi-seed replication of the single-seed runs, native-video evaluations, and joint 
measurements of pruning,
quantization, cache behaviour, and hardware cost are needed 
before broader causal or deployment claims can be made.

\newpage
\end{reviewchange}

\bibliographystyle{tmlr}
\bibliography{references}

@misc{arif2024hiredattentionguidedtokendropping,
      title={HiRED: Attention-Guided Token Dropping for Efficient Inference of High-Resolution Vision-Language Models}, 
      author={Kazi Hasan Ibn Arif and JinYi Yoon and Dimitrios S. Nikolopoulos and Hans Vandierendonck and Deepu John and Bo Ji},
      year={2024},
      eprint={2408.10945},
      archivePrefix={arXiv},
      primaryClass={cs.CV},
      url={https://arxiv.org/abs/2408.10945}, 
}

@misc{beltagy2020longformerlongdocumenttransformer,
      title={Longformer: The Long-Document Transformer}, 
      author={Iz Beltagy and Matthew E. Peters and Arman Cohan},
      year={2020},
      eprint={2004.05150},
      archivePrefix={arXiv},
      primaryClass={cs.CL},
      url={https://arxiv.org/abs/2004.05150}, 
}

@inproceedings{bolya2023tokenmergingvitfaster,
      title={Token Merging: Your ViT But Faster},
      author={Daniel Bolya and Cheng-Yang Fu and Xiaoliang Dai and Peizhao Zhang and Christoph Feichtenhofer and Judy Hoffman},
      booktitle={International Conference on Learning Representations},
      year={2023},
      note={ICLR 2023 notable top 5\%},
      url={https://openreview.net/forum?id=JroZRaRw7Eu},
}

@misc{cai2025pyramidkvdynamickvcache,
      title={PyramidKV: Dynamic KV Cache Compression based on Pyramidal Information Funneling}, 
      author={Zefan Cai and Yichi Zhang and Bofei Gao and Yuliang Liu and Yucheng Li and Tianyu Liu and Keming Lu and Wayne Xiong and Yue Dong and Junjie Hu and Wen Xiao},
      year={2025},
      eprint={2406.02069},
      archivePrefix={arXiv},
      primaryClass={cs.CL},
      url={https://arxiv.org/abs/2406.02069}, 
}

@misc{castro2023scalableperformanceanalysisvisionlanguage,
      title={Scalable Performance Analysis for Vision-Language Models}, 
      author={Santiago Castro and Oana Ignat and Rada Mihalcea},
      year={2023},
      eprint={2305.18786},
      archivePrefix={arXiv},
      primaryClass={cs.CV},
      url={https://arxiv.org/abs/2305.18786}, 
}

@misc{chen2025rettentionultrasparsevisual,
      title={Re-ttention: Ultra Sparse Visual Generation via Attention Statistical Reshape}, 
      author={Ruichen Chen and Keith G. Mills and Liyao Jiang and Chao Gao and Di Niu},
      year={2025},
      eprint={2505.22918},
      archivePrefix={arXiv},
      primaryClass={cs.CV},
      url={https://arxiv.org/abs/2505.22918}, 
}

@misc{chen2026variationawarevisiontokendropping,
      title={Variation-aware Vision Token Dropping for Faster Large Vision-Language Models}, 
      author={Junjie Chen and Xuyang Liu and Zichen Wen and Yiyu Wang and Siteng Huang and Honggang Chen},
      year={2026},
      eprint={2509.01552},
      archivePrefix={arXiv},
      primaryClass={cs.CV},
      url={https://arxiv.org/abs/2509.01552}, 
}

@misc{chiang2024locatebenchevaluatinglocatingability,
      title={LocateBench: Evaluating the Locating Ability of Vision Language Models}, 
      author={Ting-Rui Chiang and Joshua Robinson and Xinyan Velocity Yu and Dani Yogatama},
      year={2024},
      eprint={2410.19808},
      archivePrefix={arXiv},
      primaryClass={cs.CV},
      url={https://arxiv.org/abs/2410.19808}, 
}

@misc{chittyvenkata2024llminferencebenchinferencebenchmarkinglarge,
      title={LLM-Inference-Bench: Inference Benchmarking of Large Language Models on AI Accelerators}, 
      author={Krishna Teja Chitty-Venkata and Siddhisanket Raskar and Bharat Kale and Farah Ferdaus and Aditya Tanikanti and Ken Raffenetti and Valerie Taylor and Murali Emani and Venkatram Vishwanath},
      year={2024},
      eprint={2411.00136},
      archivePrefix={arXiv},
      primaryClass={cs.LG},
      url={https://arxiv.org/abs/2411.00136}, 
}

@misc{chung2026joulesgodiagnosinginference,
      title={Where Do the Joules Go? Diagnosing Inference Energy Consumption}, 
      author={Jae-Won Chung and Ruofan Wu and Jeff J. Ma and Mosharaf Chowdhury},
      year={2026},
      eprint={2601.22076},
      archivePrefix={arXiv},
      primaryClass={cs.LG},
      url={https://arxiv.org/abs/2601.22076}, 
}

@misc{delibasoglu2026spectralmanifoldregularizationstable,
      title={Spectral Manifold Regularization for Stable and Modular Routing in Deep MoE Architectures}, 
      author={Ibrahim Delibasoglu},
      year={2026},
      eprint={2601.03889},
      archivePrefix={arXiv},
      primaryClass={cs.LG},
      url={https://arxiv.org/abs/2601.03889}, 
}

@misc{deng2026gemqglobalexpertlevelmixedprecision,
      title={GEMQ: Global Expert-Level Mixed-Precision Quantization for MoE LLMs}, 
      author={Jianing Deng and Song Wang and Dongwei Wang and Zijie Liu and Tianlong Chen and Huanrui Yang and Jingtong Hu},
      year={2026},
      eprint={2605.23078},
      archivePrefix={arXiv},
      primaryClass={cs.LG},
      url={https://arxiv.org/abs/2605.23078}, 
}

@misc{do2026eigenvectorsexpertstrainingfreenoncollapsing,
      title={Eigenvectors of Experts are Training-free Non-collapsing Routers}, 
      author={Giang Do and Hung Le and Truyen Tran},
      year={2026},
      eprint={2605.30992},
      archivePrefix={arXiv},
      primaryClass={cs.LG},
      url={https://arxiv.org/abs/2605.30992}, 
}

@misc{drozdov2024videorepresentationlearningjointembedding,
      title={Video Representation Learning with Joint-Embedding Predictive Architectures}, 
      author={Katrina Drozdov and Ravid Shwartz-Ziv and Yann LeCun},
      year={2024},
      eprint={2412.10925},
      archivePrefix={arXiv},
      primaryClass={cs.CV},
      url={https://arxiv.org/abs/2412.10925}, 
}

@misc{du2024revisitingmoedensespeedaccuracy,
      title={Revisiting MoE and Dense Speed-Accuracy Comparisons for LLM Training}, 
      author={Xianzhi Du and Tom Gunter and Xiang Kong and Mark Lee and Zirui Wang and Aonan Zhang and Nan Du and Ruoming Pang},
      year={2024},
      eprint={2405.15052},
      archivePrefix={arXiv},
      primaryClass={cs.LG},
      url={https://arxiv.org/abs/2405.15052}, 
}

@misc{faure2025hermestemporalcoherentlongformunderstanding,
      title={HERMES: temporal-coHERent long-forM understanding with Episodes and Semantics}, 
      author={Gueter Josmy Faure and Jia-Fong Yeh and Min-Hung Chen and Hung-Ting Su and Shang-Hong Lai and Winston H. Hsu},
      year={2025},
      eprint={2408.17443},
      archivePrefix={arXiv},
      primaryClass={cs.CV},
      url={https://arxiv.org/abs/2408.17443}, 
}

@misc{fu2026eaquantenhancingposttrainingquantization,
      title={EAQuant: Enhancing Post-Training Quantization for MoE Models via Expert-Aware Optimization}, 
      author={Zhongqian Fu and Tianyi Zhao and Ning Ding and Xianzhi Yu and Xiaosong Li and Yehui Tang and Yunhe Wang},
      year={2026},
      eprint={2506.13329},
      archivePrefix={arXiv},
      primaryClass={cs.CL},
      url={https://arxiv.org/abs/2506.13329}, 
}

@misc{guo2025speedqstagedprocessingenhanced,
      title={SPEED-Q: Staged Processing with Enhanced Distillation towards Efficient Low-bit On-device VLM Quantization}, 
      author={Tianyu Guo and Shanwei Zhao and Shiai Zhu and Chenguang Ma},
      year={2025},
      eprint={2511.08914},
      archivePrefix={arXiv},
      primaryClass={cs.CV},
      url={https://arxiv.org/abs/2511.08914}, 
}

@inproceedings{he2021maskedautoencodersscalablevision,
      title={Masked Autoencoders Are Scalable Vision Learners},
      author={Kaiming He and Xinlei Chen and Saining Xie and Yanghao Li and Piotr Dollár and Ross Girshick},
      booktitle={Proceedings of the IEEE/CVF Conference on Computer Vision and Pattern Recognition},
      pages={16000--16009},
      year={2022},
      url={https://openaccess.thecvf.com/content/CVPR2022/html/He_Masked_Autoencoders_Are_Scalable_Vision_Learners_CVPR_2022_paper.html},
}

@ARTICLE{2024arXiv241017954H,
       author = {{He}, Xin and {Zhang}, Shunkang and {Tang}, Kaijie and {Shi}, Shaohuai and {Wang}, Yuxin and {Zeng}, Zihao and {Tang}, Zhenheng and {Chu}, Xiaowen and {Yin}, Haiyan and {Tsang}, Ivor W. and {Ong}, Yew Soon},
        title = "{ExpertFlow: Efficient Mixture-of-Experts Inference via Predictive Expert Caching and Token Scheduling}",
      journal = {arXiv e-prints},
         year = 2024,
        month = oct,
          eid = {arXiv:2410.17954},
        pages = {arXiv:2410.17954},
          doi = {10.48550/arXiv.2410.17954},
archivePrefix = {arXiv},
       eprint = {2410.17954},
 primaryClass = {cs.AI},
       adsurl = {https://ui.adsabs.harvard.edu/abs/2024arXiv241017954H}
}

@misc{he2024zipvlefficientlargevisionlanguage,
      title={ZipVL: Efficient Large Vision-Language Models with Dynamic Token Sparsification}, 
      author={Yefei He and Feng Chen and Jing Liu and Wenqi Shao and Hong Zhou and Kaipeng Zhang and Bohan Zhuang},
      year={2024},
      eprint={2410.08584},
      archivePrefix={arXiv},
      primaryClass={cs.CV},
      url={https://arxiv.org/abs/2410.08584}, 
}

@misc{ho2020constructingmultihopqadataset,
      title={Constructing A Multi-hop QA Dataset for Comprehensive Evaluation of Reasoning Steps}, 
      author={Xanh Ho and Anh-Khoa Duong Nguyen and Saku Sugawara and Akiko Aizawa},
      year={2020},
      eprint={2011.01060},
      archivePrefix={arXiv},
      primaryClass={cs.CL},
      url={https://arxiv.org/abs/2011.01060}, 
}

@article{10.1162/neco.1997.9.8.1735,
    author = {Hochreiter, Sepp and Schmidhuber, Jürgen},
    title = {Long Short-Term Memory},
    journal = {Neural Computation},
    volume = {9},
    number = {8},
    pages = {1735-1780},
    year = {1997},
    month = {11},
    issn = {0899-7667},
    doi = {10.1162/neco.1997.9.8.1735},
    url = {https://doi.org/10.1162/neco.1997.9.8.1735},
    eprint = {https://direct.mit.edu/neco/article-pdf/9/8/1735/813796/neco.1997.9.8.1735.pdf},
}

@misc{huang2025tosatokenmergingspatial,
      title={ToSA: Token Merging with Spatial Awareness}, 
      author={Hsiang-Wei Huang and Wenhao Chai and Kuang-Ming Chen and Cheng-Yen Yang and Jenq-Neng Hwang},
      year={2025},
      eprint={2506.20066},
      archivePrefix={arXiv},
      primaryClass={cs.CV},
      url={https://arxiv.org/abs/2506.20066}, 
}

@misc{huang2026modixtrainingfreemultimodalinformationdriven,
      title={MODIX: A Training-Free Multimodal Information-Driven Positional Index Scaling for Vision-Language Models}, 
      author={Ruoxiang Huang and Zhen Yuan},
      year={2026},
      eprint={2604.12537},
      archivePrefix={arXiv},
      primaryClass={cs.CV},
      url={https://arxiv.org/abs/2604.12537}, 
}

@misc{huang2026ppepositionalpreservationembedding,
      title={PPE: Positional Preservation Embedding for Token Compression in Multimodal Large Language Models}, 
      author={Mouxiao Huang and Borui Jiang and Dehua Zheng and Hailin Hu and Kai Han and Xinghao Chen},
      year={2026},
      eprint={2510.22936},
      archivePrefix={arXiv},
      primaryClass={cs.CV},
      url={https://arxiv.org/abs/2510.22936}, 
}

@misc{huber2025cosmoescompactsparsemixture,
      title={CoSMoEs: Compact Sparse Mixture of Experts}, 
      author={Patrick Huber and Akshat Shrivastava and Ernie Chang and Chinnadhurai Sankar and Ahmed Aly and Adithya Sagar},
      year={2025},
      eprint={2503.00245},
      archivePrefix={arXiv},
      primaryClass={cs.LG},
      url={https://arxiv.org/abs/2503.00245}, 
}

@misc{huo2026mamemarematrixbased,
      title={MaMe \& MaRe: Matrix-Based Token Merging and Restoration for Efficient Visual Perception and Synthesis}, 
      author={Simin Huo and Ning Li},
      year={2026},
      eprint={2604.13432},
      archivePrefix={arXiv},
      primaryClass={cs.CV},
      url={https://arxiv.org/abs/2604.13432}, 
}

@inproceedings{jaegle2021perceivergeneralperceptioniterative,
      title={Perceiver: General Perception with Iterative Attention},
      author={Andrew Jaegle and Felix Gimeno and Andrew Brock and Andrew Zisserman and Oriol Vinyals and Joao Carreira},
      booktitle={Proceedings of the 38th International Conference on Machine Learning},
      series={Proceedings of Machine Learning Research},
      volume={139},
      pages={4651--4664},
      publisher={PMLR},
      year={2021},
      url={https://proceedings.mlr.press/v139/jaegle21a.html},
}

@misc{jiang2025moecapbenchmarkingcostaccuracy,
      title={MoE-CAP: Benchmarking Cost, Accuracy and Performance of Sparse Mixture-of-Experts Systems}, 
      author={Yinsicheng Jiang and Yao Fu and Yeqi Huang and Ping Nie and Zhan Lu and Leyang Xue and Congjie He and Man-Kit Sit and Jilong Xue and Li Dong and Ziming Miao and Dayou Du and Tairan Xu and Kai Zou and Edoardo Ponti and Luo Mai},
      year={2025},
      eprint={2412.07067},
      archivePrefix={arXiv},
      primaryClass={cs.LG},
      url={https://arxiv.org/abs/2412.07067}, 
}

@misc{jie2025mixturelookupexperts,
      title={Mixture of Lookup Experts}, 
      author={Shibo Jie and Yehui Tang and Kai Han and Yitong Li and Duyu Tang and Zhi-Hong Deng and Yunhe Wang},
      year={2025},
      eprint={2503.15798},
      archivePrefix={arXiv},
      primaryClass={cs.LG},
      url={https://arxiv.org/abs/2503.15798}, 
}

@misc{jin2025streamingassistantefficientvisualtoken,
      title={StreamingAssistant: Efficient Visual Token Pruning for Accelerating Online Video Understanding}, 
      author={Xinqi Jin and Hanxun Yu and Bohan Yu and Kebin Liu and Jian Liu and Keda Tao and Yixuan Pei and Huan Wang and Fan Dang and Jiangchuan Liu and Weiqiang Wang},
      year={2025},
      eprint={2512.12560},
      archivePrefix={arXiv},
      primaryClass={cs.CV},
      url={https://arxiv.org/abs/2512.12560}, 
}

@inproceedings{John_2024,
   title={Performance and Power: Systematic Evaluation of AI Workloads on Accelerators with CARAML},
   url={http://dx.doi.org/10.1109/SCW63240.2024.00158},
   DOI={10.1109/scw63240.2024.00158},
   booktitle={SC24-W: Workshops of the International Conference for High Performance Computing, Networking, Storage and Analysis},
   publisher={IEEE},
   author={John, Chelsea Maria and Nassyr, Stepan and Penke, Carolin and Herten, Andreas},
   year={2024},
   month=Nov, pages={1164–1176} 
}

@misc{kim2023mixturequantizedexpertsmoqe,
      title={Mixture of Quantized Experts (MoQE): Complementary Effect of Low-bit Quantization and Robustness}, 
      author={Young Jin Kim and Raffy Fahim and Hany Hassan Awadalla},
      year={2023},
      eprint={2310.02410},
      archivePrefix={arXiv},
      primaryClass={cs.LG},
      url={https://arxiv.org/abs/2310.02410}, 
}

@misc{kiruluta2025csvlmcompressedsensingattention,
      title={CS-VLM: Compressed Sensing Attention for Efficient Vision-Language Representation Learning}, 
      author={Andrew Kiruluta and Preethi Raju and Priscilla Burity},
      year={2025},
      eprint={2507.02957},
      archivePrefix={arXiv},
      primaryClass={cs.CV},
      url={https://arxiv.org/abs/2507.02957}, 
}

@inproceedings{kwon2023efficientmemorymanagementlarge,
      title={Efficient Memory Management for Large Language Model Serving with PagedAttention},
      author={Woosuk Kwon and Zhuohan Li and Siyuan Zhuang and Ying Sheng and Lianmin Zheng and Cody Hao Yu and Joseph E. Gonzalez and Hao Zhang and Ion Stoica},
      booktitle={Proceedings of the 29th ACM Symposium on Operating Systems Principles},
      pages={611--626},
      publisher={ACM},
      year={2023},
      doi={10.1145/3600006.3613165},
      url={https://doi.org/10.1145/3600006.3613165},
}

@misc{la2025kernellevelenergyefficientneuralarchitecture,
      title={Kernel-Level Energy-Efficient Neural Architecture Search for Tabular Dataset}, 
      author={Hoang-Loc La and Phuong Hoai Ha},
      year={2025},
      eprint={2504.08359},
      archivePrefix={arXiv},
      primaryClass={cs.LG},
      url={https://arxiv.org/abs/2504.08359}, 
}

@article{Latif_2025,
   title={Single-Node Power Demand During AI Training: Measurements on an 8-GPU NVIDIA H100 System},
   volume={13},
   ISSN={2169-3536},
   url={http://dx.doi.org/10.1109/ACCESS.2025.3554728},
   DOI={10.1109/access.2025.3554728},
   journal={IEEE Access},
   publisher={Institute of Electrical and Electronics Engineers (IEEE)},
   author={Latif, Imran and Newkirk, Alex C. and Carbone, Matthew R. and Munir, Arslan and Lin, Yuewei and Koomey, Jonathan and Yu, Xi and Dong, Zhihua},
   year={2025},
   pages={61740–61747}
}

@inbook{Lee_2025,
   title={Lossless Token Merging Even Without Fine-Tuning in Vision Transformers},
   ISBN={9781643686318},
   ISSN={1879-8314},
   url={http://dx.doi.org/10.3233/FAIA250832},
   DOI={10.3233/faia250832},
   booktitle={ECAI 2025},
   publisher={IOS Press},
   author={Lee, Jaeyeon and Choi, Dong-Wan},
   year={2025},
   month=Oct 
}

@misc{lee2026adamergesalienceawareadaptivetoken,
      title={AdaMerge: Salience-Aware Adaptive Token Merging for Training-Free Acceleration of Vision Transformers}, 
      author={Semi Lee and Hyejin Go and Hyesong Choi},
      year={2026},
      eprint={2605.27465},
      archivePrefix={arXiv},
      primaryClass={cs.CV},
      url={https://arxiv.org/abs/2605.27465}, 
}

@inproceedings{Li_2023, series={MM ’23},
   title={Redundancy-aware Transformer for Video Question Answering},
   url={http://dx.doi.org/10.1145/3581783.3612577},
   DOI={10.1145/3581783.3612577},
   booktitle={Proceedings of the 31st ACM International Conference on Multimedia},
   publisher={ACM},
   author={Li, Yicong and Yang, Xun and Zhang, An and Feng, Chun and Wang, Xiang and Chua, Tat-Seng},
   year={2023},
   month=Oct, pages={3172–3180},
   collection={MM ’23} 
}

@misc{li2024snapkvllmknowslooking,
      title={SnapKV: LLM Knows What You are Looking for Before Generation}, 
      author={Yuhong Li and Yingbing Huang and Bowen Yang and Bharat Venkitesh and Acyr Locatelli and Hanchen Ye and Tianle Cai and Patrick Lewis and Deming Chen},
      year={2024},
      eprint={2404.14469},
      archivePrefix={arXiv},
      primaryClass={cs.CL},
      url={https://arxiv.org/abs/2404.14469}, 
}

@misc{li2025graphkvbreakingstaticselection,
      title={GraphKV: Breaking the Static Selection Paradigm with Graph-Based KV Cache Eviction}, 
      author={Xuelin Li and Xiangqi Jin and Linfeng Zhang},
      year={2025},
      eprint={2509.00388},
      archivePrefix={arXiv},
      primaryClass={cs.CL},
      url={https://arxiv.org/abs/2509.00388}, 
}

@misc{li2025inferenceoptimalvlmsneed,
      title={Inference Optimal VLMs Need Fewer Visual Tokens and More Parameters}, 
      author={Kevin Y. Li and Sachin Goyal and Joao D. Semedo and J. Zico Kolter},
      year={2025},
      eprint={2411.03312},
      archivePrefix={arXiv},
      primaryClass={cs.CV},
      url={https://arxiv.org/abs/2411.03312}, 
}

@misc{li2025madakvadaptivemodalityperceptionkv,
      title={MadaKV: Adaptive Modality-Perception KV Cache Eviction for Efficient Multimodal Long-Context Inference}, 
      author={Kunxi Li and Zhonghua Jiang and Zhouzhou Shen and Zhaode Wang and Chengfei Lv and Shengyu Zhang and Fan Wu and Fei Wu},
      year={2025},
      eprint={2506.15724},
      archivePrefix={arXiv},
      primaryClass={cs.LG},
      url={https://arxiv.org/abs/2506.15724}, 
}

@misc{li2025palmbenchcomprehensivebenchmarkcompressed,
      title={PalmBench: A Comprehensive Benchmark of Compressed Large Language Models on Mobile Platforms}, 
      author={Yilong Li and Jingyu Liu and Hao Zhang and M Badri Narayanan and Utkarsh Sharma and Shuai Zhang and Pan Hu and Yijing Zeng and Jayaram Raghuram and Suman Banerjee},
      year={2025},
      eprint={2410.05315},
      archivePrefix={arXiv},
      primaryClass={cs.LG},
      url={https://arxiv.org/abs/2410.05315}, 
}

@misc{li2026mmvirmultimodalmultigranularityrepresentation,
      title={MMViR: A Multi-Modal and Multi-Granularity Representation for Long-range Video Understanding}, 
      author={Zizhong Li and Haopeng Zhang and Jiawei Zhang},
      year={2026},
      eprint={2601.05495},
      archivePrefix={arXiv},
      primaryClass={cs.CV},
      url={https://arxiv.org/abs/2601.05495}, 
}

@misc{liu2024multistagevisiontokendropping,
      title={Multi-Stage Vision Token Dropping: Towards Efficient Multimodal Large Language Model}, 
      author={Ting Liu and Liangtao Shi and Richang Hong and Yue Hu and Quanjun Yin and Linfeng Zhang},
      year={2024},
      eprint={2411.10803},
      archivePrefix={arXiv},
      primaryClass={cs.CV},
      url={https://arxiv.org/abs/2411.10803}, 
}

@misc{liu2025surveyinferenceoptimizationtechniques,
      title={A Survey on Inference Optimization Techniques for Mixture of Experts Models}, 
      author={Jiacheng Liu and Peng Tang and Wenfeng Wang and Yuhang Ren and Xiaofeng Hou and Pheng-Ann Heng and Minyi Guo and Chao Li},
      year={2025},
      eprint={2412.14219},
      archivePrefix={arXiv},
      primaryClass={cs.LG},
      url={https://arxiv.org/abs/2412.14219}, 
}

@misc{liu2025bandwidthefficientadaptivemixtureofexpertslowrank,
      title={Bandwidth-Efficient Adaptive Mixture-of-Experts via Low-Rank Compensation}, 
      author={Zhenyu Liu and Yunzhen Liu and Zehao Fan and Garrett Gagnon and Yayue Hou and Nan Wu and Yangwook Kang and Liu Liu},
      year={2025},
      eprint={2512.17073},
      archivePrefix={arXiv},
      primaryClass={cs.LG},
      url={https://arxiv.org/abs/2512.17073}, 
}

@misc{liu2025chunkkvsemanticpreservingkvcache,
      title={ChunkKV: Semantic-Preserving KV Cache Compression for Efficient Long-Context LLM Inference}, 
      author={Xiang Liu and Zhenheng Tang and Peijie Dong and Zeyu Li and Yue Liu and Bo Li and Xuming Hu and Xiaowen Chu},
      year={2025},
      eprint={2502.00299},
      archivePrefix={arXiv},
      primaryClass={cs.CL},
      url={https://arxiv.org/abs/2502.00299}, 
}

@misc{liu2025paretoqimprovingscalinglaws,
      title={ParetoQ: Improving Scaling Laws in Extremely Low-bit LLM Quantization}, 
      author={Zechun Liu and Changsheng Zhao and Hanxian Huang and Sijia Chen and Jing Zhang and Jiawei Zhao and Scott Roy and Lisa Jin and Yunyang Xiong and Yangyang Shi and Lin Xiao and Yuandong Tian and Bilge Soran and Raghuraman Krishnamoorthi and Tijmen Blankevoort and Vikas Chandra},
      year={2025},
      eprint={2502.02631},
      archivePrefix={arXiv},
      primaryClass={cs.LG},
      url={https://arxiv.org/abs/2502.02631}, 
}

@ARTICLE{11527015,
  author={Liu, Tao and Xie, Qipeng and Han, Zhaoyang and Wang, Weizheng and Pan, Shengli and Zhang, Xiang and Su, Chunhua and Wu, Kaishun},
  journal={IEEE Transactions on Cognitive Communications and Networking}, 
  title={Efficient Edge Deployment of Mixture-of-Experts Models with Hybrid Expert Quantization}, 
  year={2026},
  volume={},
  number={},
  pages={1-1},
  doi={10.1109/TCCN.2026.3694872}
}

@misc{liu2026retentivekvstatespacememoryuncertaintyaware,
      title={RetentiveKV: State-Space Memory for Uncertainty-Aware Multimodal KV Cache Eviction}, 
      author={Sihao Liu and YuFan Xiong and Zhonghua Jiang and Zhaode Wang and chengfei lv Shengyu Zhang},
      year={2026},
      eprint={2605.04075},
      archivePrefix={arXiv},
      primaryClass={cs.LG},
      url={https://arxiv.org/abs/2605.04075}, 
}

@misc{lyu2025backdooringvisionlanguagemodelsoutofdistribution,
      title={Backdooring Vision-Language Models with Out-Of-Distribution Data}, 
      author={Weimin Lyu and Jiachen Yao and Saumya Gupta and Lu Pang and Tao Sun and Lingjie Yi and Lijie Hu and Haibin Ling and Chao Chen},
      year={2025},
      eprint={2410.01264},
      archivePrefix={arXiv},
      primaryClass={cs.CV},
      url={https://arxiv.org/abs/2410.01264}, 
}

@misc{mahajan2025attentionguidedalignmentefficient,
      title={Attention Guided Alignment in Efficient Vision-Language Models}, 
      author={Shweta Mahajan and Hoang Le and Hyojin Park and Farzad Farhadzadeh and Munawar Hayat and Fatih Porikli},
      year={2025},
      eprint={2511.17793},
      archivePrefix={arXiv},
      primaryClass={cs.CV},
      url={https://arxiv.org/abs/2511.17793}, 
}

@misc{park2026valueandstructurealignmentroutingconsistentquantization,
      title={Value-and-Structure Alignment for Routing-Consistent Quantization of Mixture-of-Experts Models}, 
      author={Hancheol Park and Geonho Lee and Tairen Piao and Tae-Ho Kim},
      year={2026},
      eprint={2606.05688},
      archivePrefix={arXiv},
      primaryClass={cs.CL},
      url={https://arxiv.org/abs/2606.05688}, 
}

@misc{pei2025greedypruneretentingcriticalvisual,
      title={GreedyPrune: Retenting Critical Visual Token Set for Large Vision Language Models}, 
      author={Ruiguang Pei and Weiqing Sun and Zhihui Fu and Jun Wang},
      year={2025},
      eprint={2506.13166},
      archivePrefix={arXiv},
      primaryClass={cs.CV},
      url={https://arxiv.org/abs/2506.13166}, 
}

@misc{sha2026sparsemixtureofexpertsroutingvisual,
      title={Sparse Mixture-of-Experts Routing in Visual Diffusion Transformers:Diagnosis, Boundary Calibration and Evolutionary Roadmap from Routing Collapse to Selective Deadlock}, 
      author={Haiying Sha},
      year={2026},
      eprint={2605.19378},
      archivePrefix={arXiv},
      primaryClass={cs.CV},
      url={https://arxiv.org/abs/2605.19378}, 
}

@misc{shao2026surveytokencompressionefficient,
      title={A Survey of Token Compression for Efficient Multimodal Large Language Models}, 
      author={Kele Shao and Keda Tao and Kejia Zhang and Sicheng Feng and Mu Cai and Yuzhang Shang and Haoxuan You and Can Qin and Yang Sui and Huan Wang},
      year={2026},
      eprint={2507.20198},
      archivePrefix={arXiv},
      primaryClass={cs.CV},
      url={https://arxiv.org/abs/2507.20198}, 
}

@misc{shen2025expertflowadaptiveexpertscheduling,
      title={ExpertFlow: Adaptive Expert Scheduling and Memory Coordination for Efficient MoE Inference}, 
      author={Zixu Shen and Kexin Chu and Yifan Zhang and Dawei Xiang and Runxin Wu and Wei Zhang},
      year={2025},
      eprint={2510.26730},
      archivePrefix={arXiv},
      primaryClass={cs.DC},
      url={https://arxiv.org/abs/2510.26730}, 
}

@misc{shen2025fastviddynamicdensitypruning,
      title={FastVID: Dynamic Density Pruning for Fast Video Large Language Models}, 
      author={Leqi Shen and Guoqiang Gong and Tao He and Yifeng Zhang and Pengzhang Liu and Sicheng Zhao and Guiguang Ding},
      year={2025},
      eprint={2503.11187},
      archivePrefix={arXiv},
      primaryClass={cs.CV},
      url={https://arxiv.org/abs/2503.11187}, 
}

@misc{song2026evocomplearningvisualtoken,
      title={EvoComp: Learning Visual Token Compression for Multimodal Large Language Models via Semantic-Guided Evolutionary Labeling}, 
      author={Jiafei Song and Fengwei Zhou and Jin Qu and Wenjin Jason Li and Tong Wu and Gengjian Xue and Zhikang Zhao and Daomin Wei and Yichao Lu and Bailin Na},
      year={2026},
      eprint={2604.17087},
      archivePrefix={arXiv},
      primaryClass={cs.CV},
      url={https://arxiv.org/abs/2604.17087}, 
}

@misc{sukthanker2024hwgptbenchhardwareawarearchitecturebenchmark,
      title={HW-GPT-Bench: Hardware-Aware Architecture Benchmark for Language Models}, 
      author={Rhea Sanjay Sukthanker and Arber Zela and Benedikt Staffler and Aaron Klein and Lennart Purucker and Joerg K. H. Franke and Frank Hutter},
      year={2024},
      eprint={2405.10299},
      archivePrefix={arXiv},
      primaryClass={cs.LG},
      url={https://arxiv.org/abs/2405.10299}, 
}

@misc{sun2026doessparsemoehelp,
      title={When Does Sparse MoE Help in Vision? The Role of Backbone Compute Leverage in Sparse Routing}, 
      author={Libo Sun and Po-wei Harn and Peixiong He and Xiao Qin},
      year={2026},
      eprint={2605.15484},
      archivePrefix={arXiv},
      primaryClass={cs.CV},
      url={https://arxiv.org/abs/2605.15484}, 
}

@misc{tairin2025emoetaskawarememoryefficient,
      title={eMoE: Task-aware Memory Efficient Mixture-of-Experts-Based (MoE) Model Inference}, 
      author={Suraiya Tairin and Shohaib Mahmud and Haiying Shen and Anand Iyer},
      year={2025},
      eprint={2503.06823},
      archivePrefix={arXiv},
      primaryClass={cs.LG},
      url={https://arxiv.org/abs/2503.06823}, 
}

@misc{tian2024energyefficiencynavigatingtrilemmaenergy,
      title={Towards Energy-Efficiency by Navigating the Trilemma of Energy, Latency, and Accuracy}, 
      author={Boyuan Tian and Yihan Pang and Muhammad Huzaifa and Shenlong Wang and Sarita Adve},
      year={2024},
      eprint={2409.04018},
      archivePrefix={arXiv},
      primaryClass={cs.CV},
      url={https://arxiv.org/abs/2409.04018}, 
}

@inproceedings{tong2022videomaemaskedautoencodersdataefficient,
      title={VideoMAE: Masked Autoencoders are Data-Efficient Learners for Self-Supervised Video Pre-Training},
      author={Zhan Tong and Yibing Song and Jue Wang and Limin Wang},
      booktitle={Advances in Neural Information Processing Systems},
      volume={35},
      year={2022},
      url={https://papers.nips.cc/paper_files/paper/2022/hash/416f9cb3276121c42eebb86352a4354a-Abstract-Conference.html},
}

@misc{treviso2023efficientmethodsnaturallanguage,
      title={Efficient Methods for Natural Language Processing: A Survey}, 
      author={Marcos Treviso and Ji-Ung Lee and Tianchu Ji and Betty van Aken and Qingqing Cao and Manuel R. Ciosici and Michael Hassid and Kenneth Heafield and Sara Hooker and Colin Raffel and Pedro H. Martins and André F. T. Martins and Jessica Zosa Forde and Peter Milder and Edwin Simpson and Noam Slonim and Jesse Dodge and Emma Strubell and Niranjan Balasubramanian and Leon Derczynski and Iryna Gurevych and Roy Schwartz},
      year={2023},
      eprint={2209.00099},
      archivePrefix={arXiv},
      primaryClass={cs.CL},
      url={https://arxiv.org/abs/2209.00099}, 
}

@article{trivedi2022musiquemultihopquestionssinglehop,
      title={MuSiQue: Multihop Questions via Single-hop Question Composition},
      author={Harsh Trivedi and Niranjan Balasubramanian and Tushar Khot and Ashish Sabharwal},
      journal={Transactions of the Association for Computational Linguistics},
      volume={10},
      pages={539--554},
      publisher={MIT Press},
      year={2022},
      doi={10.1162/tacl_a_00475},
      url={https://aclanthology.org/2022.tacl-1.31/},
}

@misc{tschand2025mlperfpowerbenchmarkingenergy,
      title={MLPerf Power: Benchmarking the Energy Efficiency of Machine Learning Systems from Microwatts to Megawatts for Sustainable AI}, 
      author={Arya Tschand and Arun Tejusve Raghunath Rajan and Sachin Idgunji and Anirban Ghosh and Jeremy Holleman and Csaba Kiraly and Pawan Ambalkar and Ritika Borkar and Ramesh Chukka and Trevor Cockrell and Oliver Curtis and Grigori Fursin and Miro Hodak and Hiwot Kassa and Anton Lokhmotov and Dejan Miskovic and Yuechao Pan and Manu Prasad Manmathan and Liz Raymond and Tom St. John and Arjun Suresh and Rowan Taubitz and Sean Zhan and Scott Wasson and David Kanter and Vijay Janapa Reddi},
      year={2025},
      eprint={2410.12032},
      archivePrefix={arXiv},
      primaryClass={cs.AR},
      url={https://arxiv.org/abs/2410.12032}, 
}

@inproceedings{vaswani2023attentionneed,
      title={Attention Is All You Need},
      author={Ashish Vaswani and Noam Shazeer and Niki Parmar and Jakob Uszkoreit and Llion Jones and Aidan N. Gomez and Lukasz Kaiser and Illia Polosukhin},
      booktitle={Advances in Neural Information Processing Systems},
      volume={30},
      year={2017},
      url={https://papers.nips.cc/paper_files/paper/2017/hash/3f5ee243547dee91fbd053c1c4a845aa-Abstract.html},
}

@misc{wang2024auxiliarylossfreeloadbalancingstrategy,
      title={Auxiliary-Loss-Free Load Balancing Strategy for Mixture-of-Experts}, 
      author={Lean Wang and Huazuo Gao and Chenggang Zhao and Xu Sun and Damai Dai},
      year={2024},
      eprint={2408.15664},
      archivePrefix={arXiv},
      primaryClass={cs.LG},
      url={https://arxiv.org/abs/2408.15664}, 
}

@misc{wang2024homehierarchymultigateexperts,
      title={HoME: Hierarchy of Multi-Gate Experts for Multi-Task Learning at Kuaishou}, 
      author={Xu Wang and Jiangxia Cao and Zhiyi Fu and Kun Gai and Guorui Zhou},
      year={2024},
      eprint={2408.05430},
      archivePrefix={arXiv},
      primaryClass={cs.IR},
      url={https://arxiv.org/abs/2408.05430}, 
}

@misc{wang2025bivlmpushingultralowprecision,
      title={Bi-VLM: Pushing Ultra-Low Precision Post-Training Quantization Boundaries in Vision-Language Models}, 
      author={Xijun Wang and Junyun Huang and Rayyan Abdalla and Chengyuan Zhang and Ruiqi Xian and Dinesh Manocha},
      year={2025},
      eprint={2509.18763},
      archivePrefix={arXiv},
      primaryClass={cs.CV},
      url={https://arxiv.org/abs/2509.18763}, 
}

@inproceedings{Wang_2025, series={ACM MOBICOM ’25},
   title={D2MoE: Dual Routing and Dynamic Scheduling for Efficient On-Device MoE-based LLM Serving},
   url={http://dx.doi.org/10.1145/3680207.3723493},
   DOI={10.1145/3680207.3723493},
   booktitle={Proceedings of the 31st Annual International Conference on Mobile Computing and Networking},
   publisher={ACM},
   author={Wang, Haodong and Zhou, Qihua and Hong, Zicong and Guo, Song},
   year={2025},
   month=Nov, pages={574–588},
   collection={ACM MOBICOM ’25} 
}

@misc{wang2025lvclightweightcompressionframework,
      title={LVC: A Lightweight Compression Framework for Enhancing VLMs in Long Video Understanding}, 
      author={Ziyi Wang and Haoran Wu and Yiming Rong and Deyang Jiang and Yixin Zhang and Yunlong Zhao and Shuang Xu and Bo XU},
      year={2025},
      eprint={2504.06835},
      archivePrefix={arXiv},
      primaryClass={cs.CV},
      url={https://arxiv.org/abs/2504.06835}, 
}

@misc{wang2025remoefullydifferentiablemixtureofexperts,
      title={ReMoE: Fully Differentiable Mixture-of-Experts with ReLU Routing}, 
      author={Ziteng Wang and Jun Zhu and Jianfei Chen},
      year={2025},
      eprint={2412.14711},
      archivePrefix={arXiv},
      primaryClass={cs.LG},
      url={https://arxiv.org/abs/2412.14711}, 
}

@misc{wang2026motemixtureternaryexperts,
      title={MoTE: Mixture of Ternary Experts for Memory-efficient Large Multimodal Models}, 
      author={Hongyu Wang and Jiayu Xu and Ruiping Wang and Yan Feng and Yitao Zhai and Peng Pei and Xunliang Cai and Xilin Chen},
      year={2026},
      eprint={2506.14435},
      archivePrefix={arXiv},
      primaryClass={cs.CV},
      url={https://arxiv.org/abs/2506.14435}, 
}

@misc{wu2022rapredundancyawarevideolanguagepretraining,
      title={RaP: Redundancy-aware Video-language Pre-training for Text-Video Retrieval}, 
      author={Xing Wu and Chaochen Gao and Zijia Lin and Zhongyuan Wang and Jizhong Han and Songlin Hu},
      year={2022},
      eprint={2210.06881},
      archivePrefix={arXiv},
      primaryClass={cs.CV},
      url={https://arxiv.org/abs/2210.06881}, 
}

@inproceedings{xiao2024efficientstreaminglanguagemodels,
      title={Efficient Streaming Language Models with Attention Sinks},
      author={Guangxuan Xiao and Yuandong Tian and Beidi Chen and Song Han and Mike Lewis},
      booktitle={International Conference on Learning Representations},
      year={2024},
      url={https://openreview.net/forum?id=NG7sS51zVF},
}

@misc{xia2025fastmmoe,
      title={FastMMoE: Accelerating Multimodal Large Language Models through Dynamic Expert Activation and Routing-Aware Token Pruning},
      author={Guoyang Xia and Yifeng Ding and Fengfa Li and Lei Ren and Wei Chen and Fangxiang Feng and Xiaojie Wang},
      year={2025},
      eprint={2511.17885},
      archivePrefix={arXiv},
      primaryClass={cs.CV},
      url={https://arxiv.org/abs/2511.17885},
}

@misc{xue2026vlmqtokensaliencydrivenposttraining,
      title={VLMQ: Token Saliency-Driven Post-Training Quantization for Vision-language Models}, 
      author={Yufei Xue and Yushi Huang and Jiawei Shao and Lunjie Zhu and Chi Zhang and Xuelong Li and Jun Zhang},
      year={2026},
      eprint={2508.03351},
      archivePrefix={arXiv},
      primaryClass={cs.CV},
      url={https://arxiv.org/abs/2508.03351}, 
}

@misc{yan2024polythrottleenergyefficientneuralnetwork,
      title={PolyThrottle: Energy-efficient Neural Network Inference on Edge Devices}, 
      author={Minghao Yan and Hongyi Wang and Shivaram Venkataraman},
      year={2024},
      eprint={2310.19991},
      archivePrefix={arXiv},
      primaryClass={cs.LG},
      url={https://arxiv.org/abs/2310.19991}, 
}

@inproceedings{yang2018hotpotqadatasetdiverseexplainable,
      title={HotpotQA: A Dataset for Diverse, Explainable Multi-hop Question Answering},
      author={Zhilin Yang and Peng Qi and Saizheng Zhang and Yoshua Bengio and William W. Cohen and Ruslan Salakhutdinov and Christopher D. Manning},
      booktitle={Proceedings of the 2018 Conference on Empirical Methods in Natural Language Processing},
      pages={2369--2380},
      address={Brussels, Belgium},
      publisher={Association for Computational Linguistics},
      year={2018},
      doi={10.18653/v1/D18-1259},
      url={https://aclanthology.org/D18-1259/},
}

@misc{yang2024doubleexponentialincreasesinferenceenergy,
      title={Double-Exponential Increases in Inference Energy: The Cost of the Race for Accuracy}, 
      author={Zeyu Yang and Karel Adamek and Wesley Armour},
      year={2024},
      eprint={2412.09731},
      archivePrefix={arXiv},
      primaryClass={cs.CV},
      url={https://arxiv.org/abs/2412.09731}, 
}

@misc{yang2026zipmoeefficientondevicemoe,
      title={ZipMoE: Efficient On-Device MoE Serving via Lossless Compression and Cache-Affinity Scheduling}, 
      author={Yuchen Yang and Yaru Zhao and Pu Yang and Shaowei Wang and Zhi-Hua Zhou},
      year={2026},
      eprint={2601.21198},
      archivePrefix={arXiv},
      primaryClass={cs.DC},
      url={https://arxiv.org/abs/2601.21198}, 
}

@ARTICLE{10906629,
  author={Yi, Rongjie and Guo, Liwei and Wei, Shiyun and Zhou, Ao and Wang, Shangguang and Xu, Mengwei},
  journal={IEEE Transactions on Mobile Computing}, 
  title={EdgeMoE: Empowering Sparse Large Language Models on Mobile Devices}, 
  year={2025},
  volume={24},
  number={8},
  pages={7059-7073},
  doi={10.1109/TMC.2025.3546466}
}

@misc{yi2025edgemoeempoweringsparselarge,
      title={EdgeMoE: Empowering Sparse Large Language Models on Mobile Devices}, 
      author={Rongjie Yi and Liwei Guo and Shiyun Wei and Ao Zhou and Shangguang Wang and Mengwei Xu},
      year={2025},
      eprint={2308.14352},
      archivePrefix={arXiv},
      primaryClass={cs.LG},
      url={https://arxiv.org/abs/2308.14352}, 
}

@misc{yu2026visiontrimunifiedvisiontoken,
      title={VisionTrim: Unified Vision Token Compression for Training-Free MLLM Acceleration}, 
      author={Hanxun Yu and Wentong Li and Xuan Qu and Song Wang and Junbo Chen and Jianke Zhu},
      year={2026},
      eprint={2601.22674},
      archivePrefix={arXiv},
      primaryClass={cs.CV},
      url={https://arxiv.org/abs/2601.22674}, 
}

@misc{yuan2025moelenshardwarelimithighthroughput,
      title={MoE-Lens: Towards the Hardware Limit of High-Throughput MoE LLM Serving Under Resource Constraints}, 
      author={Yichao Yuan and Lin Ma and Nishil Talati},
      year={2025},
      eprint={2504.09345},
      archivePrefix={arXiv},
      primaryClass={cs.DC},
      url={https://arxiv.org/abs/2504.09345}, 
}

@inproceedings{zaheer2021bigbirdtransformerslonger,
      title={Big Bird: Transformers for Longer Sequences},
      author={Manzil Zaheer and Guru Guruganesh and Kumar Avinava Dubey and Joshua Ainslie and Chris Alberti and Santiago Ontanon and Philip Pham and Anirudh Ravula and Qifan Wang and Li Yang and Amr Ahmed},
      booktitle={Advances in Neural Information Processing Systems},
      volume={33},
      year={2020},
      url={https://papers.nips.cc/paper/2020/hash/c8512d142a2d849725f31a9a7a361ab9-Abstract.html},
}

@misc{zhang2023h2oheavyhitteroracleefficient,
      title={H$_2$O: Heavy-Hitter Oracle for Efficient Generative Inference of Large Language Models}, 
      author={Zhenyu Zhang and Ying Sheng and Tianyi Zhou and Tianlong Chen and Lianmin Zheng and Ruisi Cai and Zhao Song and Yuandong Tian and Christopher Ré and Clark Barrett and Zhangyang Wang and Beidi Chen},
      year={2023},
      eprint={2306.14048},
      archivePrefix={arXiv},
      primaryClass={cs.LG},
      url={https://arxiv.org/abs/2306.14048}, 
}

@misc{zhang2025dyntokdynamiccompressionvisual,
      title={DynTok: Dynamic Compression of Visual Tokens for Efficient and Effective Video Understanding}, 
      author={Hongzhi Zhang and Jingyuan Zhang and Xingguang Ji and Qi Wang and Fuzheng Zhang},
      year={2025},
      eprint={2506.03990},
      archivePrefix={arXiv},
      primaryClass={cs.CL},
      url={https://arxiv.org/abs/2506.03990}, 
}

@misc{zhang2025sparsevlmvisualtokensparsification,
      title={SparseVLM: Visual Token Sparsification for Efficient Vision-Language Model Inference}, 
      author={Yuan Zhang and Chun-Kai Fan and Junpeng Ma and Wenzhao Zheng and Tao Huang and Kuan Cheng and Denis Gudovskiy and Tomoyuki Okuno and Yohei Nakata and Kurt Keutzer and Shanghang Zhang},
      year={2025},
      eprint={2410.04417},
      archivePrefix={arXiv},
      primaryClass={cs.CV},
      url={https://arxiv.org/abs/2410.04417}, 
}

@misc{zhang2025speculativedecodingmeetsquantization,
      title={Speculative Decoding Meets Quantization: Compatibility Evaluation and Hierarchical Framework Design}, 
      author={Yudi Zhang and Weilin Zhao and Xu Han and Tiejun Zhao and Wang Xu and Hailong Cao and Conghui Zhu},
      year={2025},
      eprint={2505.22179},
      archivePrefix={arXiv},
      primaryClass={cs.CL},
      url={https://arxiv.org/abs/2505.22179}, 
}

@misc{zhang2025vqtokenneuraldiscretetoken,
      title={VQToken: Neural Discrete Token Representation Learning for Extreme Token Reduction in Video Large Language Models}, 
      author={Haichao Zhang and Yun Fu},
      year={2025},
      eprint={2503.16980},
      archivePrefix={arXiv},
      primaryClass={cs.CV},
      url={https://arxiv.org/abs/2503.16980}, 
}

@misc{zhang2026janusdisaggregatingattentionexperts,
      title={Janus: Disaggregating Attention and Experts for Scalable MoE Inference}, 
      author={Zhexiang Zhang and Ye Wang and Yumiao Zhao and Jiayu Xiao and Qianjing Yang and Xiangyu Wang and Jingzhe Jiang and Qizhen Weng and Ruichuan Chen and Shaohuai Shi and Adel N. Toosi and Yin Chen and Minchen Yu},
      year={2026},
      eprint={2512.13525},
      archivePrefix={arXiv},
      primaryClass={cs.DC},
      url={https://arxiv.org/abs/2512.13525}, 
}

@misc{zhao2024representationcollapsingproblemsvector,
      title={Representation Collapsing Problems in Vector Quantization}, 
      author={Wenhao Zhao and Qiran Zou and Rushi Shah and Dianbo Liu},
      year={2024},
      eprint={2411.16550},
      archivePrefix={arXiv},
      primaryClass={cs.LG},
      url={https://arxiv.org/abs/2411.16550}, 
}

@misc{zhao2026mobieefficientinferencemixture,
      title={MoBiE: Efficient Inference of Mixture of Binary Experts under Post-Training Quantization}, 
      author={Zhixiong Zhao and Zukang Xu and Zhixuan Chen and Dawei Yang},
      year={2026},
      eprint={2604.06798},
      archivePrefix={arXiv},
      primaryClass={cs.LG},
      url={https://arxiv.org/abs/2604.06798}, 
}

@inproceedings{Zhong_2024, series={ICCAD ’24},
   title={AdapMoE: Adaptive Sensitivity-based Expert Gating and Management for Efficient MoE Inference},
   url={http://dx.doi.org/10.1145/3676536.3676741},
   DOI={10.1145/3676536.3676741},
   booktitle={Proceedings of the 43rd IEEE/ACM International Conference on Computer-Aided Design},
   publisher={ACM},
   author={Zhong, Shuzhang and Liang, Ling and Wang, Yuan and Wang, Runsheng and Huang, Ru and Li, Meng},
   year={2024},
   month=Oct, pages={1–9},
   collection={ICCAD ’24} 
}

@misc{zhong2026breakingmodalityheterogeneitylowbit,
      title={Breaking Modality Heterogeneity in Low-Bit Quantization for Large Vision-Language Models}, 
      author={Yi Zhong and Haotong Qin and Xindong Zhang and Lei Zhang and Guolei Sun},
      year={2026},
      eprint={2605.19929},
      archivePrefix={arXiv},
      primaryClass={cs.CV},
      url={https://arxiv.org/abs/2605.19929}, 
}

@misc{zhou2026signaldegradationcomputationcollapse,
      title={From Signal Degradation to Computation Collapse: Uncovering the Two Failure Modes of LLM Quantization}, 
      author={Chenxi Zhou and Pengfei Cao and Jiang Li and Bohan Yu and Jinyu Ye and Jun Zhao and Kang Liu},
      year={2026},
      eprint={2604.19884},
      archivePrefix={arXiv},
      primaryClass={cs.CL},
      url={https://arxiv.org/abs/2604.19884}, 
}

@misc{zhu2025addressingrepresentationcollapsevector,
      title={Addressing Representation Collapse in Vector Quantized Models with One Linear Layer}, 
      author={Yongxin Zhu and Bocheng Li and Yifei Xin and Zhihua Xia and Linli Xu},
      year={2025},
      eprint={2411.02038},
      archivePrefix={arXiv},
      primaryClass={cs.LG},
      url={https://arxiv.org/abs/2411.02038}, 
}

@misc{zou2024secondshoursreviewingmultimodal,
      title={From Seconds to Hours: Reviewing MultiModal Large Language Models on Comprehensive Long Video Understanding}, 
      author={Heqing Zou and Tianze Luo and Guiyang Xie and Victor and Zhang and Fengmao Lv and Guangcong Wang and Junyang Chen and Zhuochen Wang and Hansheng Zhang and Huaijian Zhang},
      year={2024},
      eprint={2409.18938},
      archivePrefix={arXiv},
      primaryClass={cs.CV},
      url={https://arxiv.org/abs/2409.18938}, 
}

\end{document}